\documentclass[10pt,twocolumn,letterpaper]{article}
\usepackage[pagenumbers]{cvpr}
\usepackage{multirow}
\usepackage{array}
\usepackage{bm}          %
\usepackage{amsthm}      %
\usepackage{enumitem}    %
\usepackage{colortbl}    %
\usepackage{pifont}
\usepackage{placeins}    %
\usepackage{dblfloatfix}

\makeatletter\def\@flsetnum#1{}\makeatother
\newcommand{\cmark}{\ding{51}}

\definecolor{best}{RGB}{198,222,227}
\definecolor{second}{RGB}{230,240,242}
\theoremstyle{plain}

\definecolor{cvprblue}{rgb}{0.21,0.49,0.74}
\usepackage[pagebackref,breaklinks,colorlinks,allcolors=cvprblue]{hyperref}
\usepackage[capitalize]{cleveref}
\crefname{section}{Sec.}{Secs.}
\crefname{table}{Tab.}{Tabs.}
\crefname{figure}{Fig.}{Figs.}
\crefname{equation}{Eq.}{Eqs.}
\def\confName{CVPR}
\def\confYear{2027}
\graphicspath{{fig/}}
\AtBeginEnvironment{table}{}
\AtBeginEnvironment{table*}{}

\title{Any3R: Feedforward 3D Reconstruction from Any Camera Type}
\author{Jingong Chen\textsuperscript{1,2} \quad
Qingwen Zhang\textsuperscript{3$\dagger$} \quad
Youming Deng\textsuperscript{4$\dagger$} \quad
Chulwoo Pack\textsuperscript{1}\\
Olaf Wysocki\textsuperscript{5} \quad
Boris Jutzi\textsuperscript{6} \quad
Kwanghee Won\textsuperscript{1*} \quad
Kyle Gao\textsuperscript{2}\\[6pt]
{\small \textsuperscript{1}South Dakota State University \quad \textsuperscript{2}Aalto University \quad \textsuperscript{3}KTH Royal Institute of Technology}\\
{\small \textsuperscript{4}Cornell University \quad \textsuperscript{5}University of Cambridge \quad \textsuperscript{6}Karlsruhe Institute of Technology}\\[4pt]
{\small \textsuperscript{$\dagger$}Equal contribution \quad \textsuperscript{*}Corresponding author}}

\makeatletter
\def\abstract{\iftoggle{cvprpagenumbers}{}{\thispagestyle{empty}}\centerline{\large\bf Abstract}\vspace*{4pt}\noindent\it\ignorespaces}\makeatother
\begin{document}
\makeatletter
\twocolumn[%
  \@maketitle
  \begin{center}
  {\includegraphics[width=\textwidth]{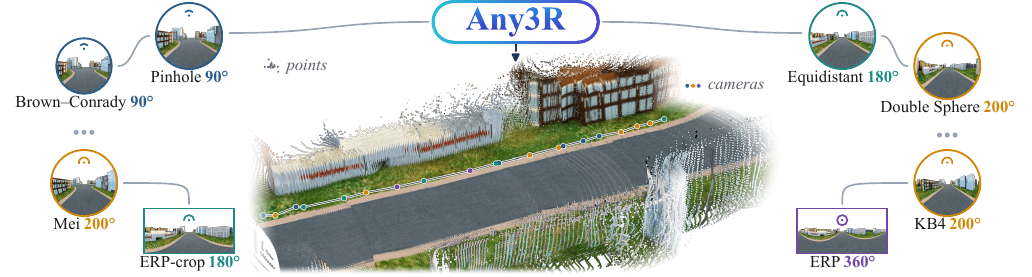}}
  \captionof{figure}{\textbf{Any3R} is a feed-forward model that reconstructs a scene and its camera poses from views of any mix of central cameras.}
  \label{fig::teaser}
  \end{center}
  \vspace{2pt}
]
\makeatother

\begin{abstract}
Geometry foundation models recover camera poses and dense geometry from a set of images with a feed-forward network, but they expect pinhole images.
That assumption is inside the backbone too: the positional encoding and the training loss both bind an image position to a direction in space, and both depend on the lens.
We condition both on the dense ray field, the unit direction through every pixel, which describes any central camera without a lens type, distortion coefficients or a bound on field of view.
\textit{Angular coordinates} give the rotary encoding the measured ray angle of each patch, evaluated per view, and \textit{solid-angle weighting} scales the point loss by the area each pixel covers on the sphere, so neither the encoding nor the supervision depends on the lens.
\textbf{Any3R} is therefore a feed-forward model that takes any central camera, and is robust to mixing many camera types per scene, with no retraining per lens. 
On a benchmark of thirteen central cameras across three datasets, where every method reconstructs with its own predicted camera poses, it is the most consistently accurate method. \textbf{Any3R} enables scalable 3D reconstruction from heterogeneous crowdsourced and archival imagery collections sourced across diverse camera types.
Project page: \url{https://jingong-chen.github.io/Any3R/}
\end{abstract}
\section{Introduction}
\label{sec::intro}
Recovering the 3D structure of a scene from a set of images is a fundamental problem in computer vision.
Geometry foundation models such as VGGT~\cite{wang2025vggt}, $\pi^3$~\cite{wang2025pi3}, MapAnything~\cite{keetha2025mapanything}, VGGT-$\Omega$~\cite{wang2026vggtomega} and Depth Anything 3~\cite{depthanything3} now solve it with a feed-forward network~\cite{wang2024dust3r,leroy2024mast3r,duisterhof2025mast3rsfm,cabon2025must3r,wang2025spann3r,wang2025cut3r,yang2025fast3r,zhang2025monst3r}.
A regular camera sees only a narrow part of a scene in each view, so covering a whole room or street takes many views captured densely and carefully.

\begin{figure*}[t]
\centering
\includegraphics[width=\textwidth]{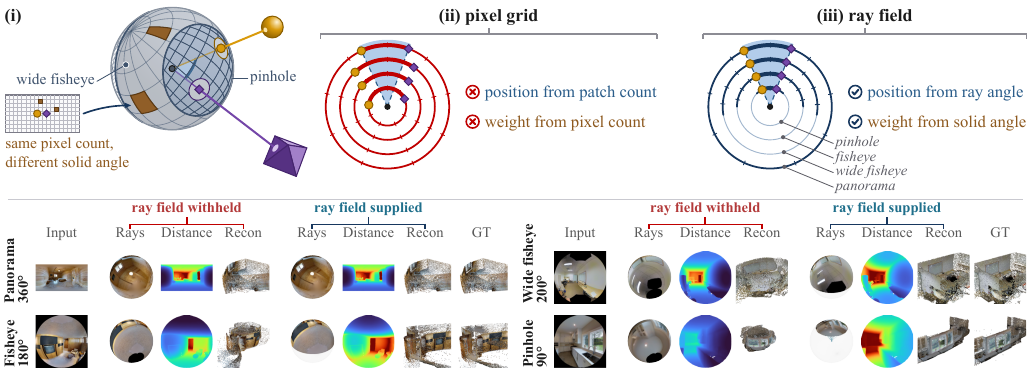}
\caption{\textbf{Motivation.} Read from the pixel grid (ii), a patch position stands for a different direction and a pixel for a different solid angle on every lens (i), and the ray field (iii) corrects both. In the bottom row the ray field is withheld, so the backbone reads every frame as a panorama and the reconstruction follows that reading, which is correct only at $360^\circ$.}
\label{fig::motiv}
\vspace{-1em}
\end{figure*}

The direct way to cover more of a scene per view is a wider lens, and wide-angle and panoramic cameras are already common in autonomous driving~\cite{yogamani2019woodscape,liao2022kitti360} and indoor scanning~\cite{chang2017matterport3d,armeni2017joint}.
Feed-forward models cannot use such images as they are, because they read every frame as a pinhole image.
Recent work adapts them to particular lenses, with a discrete token for the camera model, an adapter around a frozen backbone, a pairwise alignment stage, a conversion of the input to a cubemap, or retraining on equirectangular images~\cite{jung2026wid3r,duan2026fisheye3r,guruprasad2026cam3r,bozic2026pager,panovggt2026,realsee2026argus}, but each supports only a fixed list of camera models, so an unseen camera needs retraining or a new conversion step.
Mixing cameras of different types in one reconstruction, such as a pinhole view, a fisheye frame and a panorama of one scene as in \cref{fig::teaser}, is harder still.
Wid3R~\cite{jung2026wid3r} and Fisheye3R~\cite{duan2026fisheye3r} mix only the camera types they support, CAM3R~\cite{guruprasad2026cam3r} reconstructs two views at a time and aligns them afterwards, and the panoramic models accept panoramas alone.
MapAnything~\cite{keetha2025mapanything} and OmniVGGT~\cite{li2026omnivggt} take the intrinsics of each view as input and could mix cameras in principle, but a pinhole intrinsic matrix cannot describe a field of view of $180^\circ$ or more, the range of many fisheye lenses and every panorama.

Both limits trace back to the backbone's positional encoding, which none of these methods changes: it locates each patch by its pixel index~\cite{su2024roformer}. 
On a pinhole camera, the index tells the backbone roughly where the patch looks; on a fisheye, each pixel covers a wider angle, so the same index points somewhere else (\cref{fig::motiv}(ii)).
The training loss has a similar problem: it weights every pixel equally, but a fisheye pixel near the edge sees a different amount of the scene than one at the center (\cref{fig::motiv}(i)).

We present Any3R, which removes the pinhole assumption by giving the ray direction of every pixel to both the encoding and the loss (\cref{fig::motiv}(iii)).
In pinhole, fisheye and panoramic cameras, all rays pass through one point.
Such central cameras are fully described by the ray direction of each pixel, whatever the lens.
\textit{Angular coordinates} position a patch by the angle of its ray instead of its index, so one unit of position means the same angle on every camera. 
\textit{Solid-angle weighting} scales the loss by how much of the scene each pixel sees, so every part of the scene counts equally whatever the lens. 
Because each view carries its own ray field, Any3R reconstructs from any central camera without rectification, and views from different cameras can be processed together, in any number. 
On three public datasets imaged through thirteen camera models, where every method reconstructs with its own predicted camera poses, Any3R gives the most accurate reconstruction and camera poses on wide-angle and panoramic lenses and is on par with the strongest pinhole baseline at $90^\circ$. It also outperforms the methods built for wide-angle or panoramic lenses on every lens they take, and its pose estimates on real pinhole photographs are competitive with the strongest pinhole baselines. Its margin over the pinhole models is smallest at $90^\circ$ and much larger on the wider lenses.

Our contributions are threefold:
\begin{itemize}[leftmargin=*]
\itemsep2pt
\item A feed-forward model that reconstructs from views of any mix of central cameras, with no calibration search, no undistortion and no retraining per lens.

\item \textit{Angular coordinates} and \textit{solid-angle weighting}, two modules that condition the positional encoding and the training loss on the dense ray field, so that one set of weights serves every central camera.

\item State-of-the-art reconstruction and camera pose on three public datasets imaged through thirteen central camera models, where every method reconstructs with its own predicted camera poses, with the largest margins on wide-angle and panoramic lenses.
\end{itemize}

\section{Related Work}
\label{sec::related}
\begin{figure*}[t]
\centering
\includegraphics[width=\textwidth]{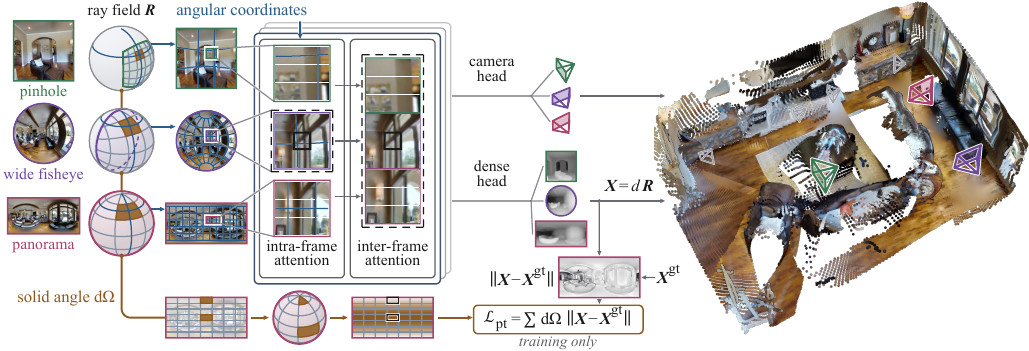}
\caption{\textbf{Overview.} The ray field $\bm{R}$ of each view gives its patches angular coordinates for the rotary encoding in intra-frame attention, and the heads predict the camera poses and a radial distance $d$ per pixel, decoded as $\bm{X}=d\bm{R}$. In training, the point loss $\mathcal{L}_{\mathrm{pt}}$ compares $\bm{X}$ with the ground truth $\bm{X}^{\mathrm{gt}}$, and the solid angle $\mathrm{d}\Omega$ of the same $\bm{R}$ weighs the error of each pixel, shown here on the panorama, where cells of equal pixel count cover unequal areas of the sphere.}
\label{fig::method}
\end{figure*}

\subsection{3D Geometry Foundation Models}

Classical structure-from-motion recovers geometry by optimising correspondences and poses jointly~\cite{schonberger2016colmap}, and per-scene neural representations~\cite{mildenhall2020nerf,kerbl20233dgs} keep that optimisation while changing what is optimised.
DUSt3R~\cite{wang2024dust3r} instead trained one network to regress a pointmap in a common frame, removing the optimisation loop and the need for known intrinsics, and MASt3R~\cite{leroy2024mast3r} added a matching head to the same backbone~\cite{weinzaepfel2022croco,weinzaepfel2023croco2,dosovitskiy2021vit,oquab2024dinov2}.

Extending two views to many has taken several forms: a global stage on top of the pairwise backbone~\cite{duisterhof2025mast3rsfm,cabon2025must3r}, a latent state that lets views be streamed~\cite{wang2025spann3r,wang2025cut3r}, and set-level attention that fuses hundreds at once~\cite{yang2025fast3r}.
VGGT~\cite{wang2025vggt} predicts poses, depth, pointmaps and tracks together, $\pi^3$~\cite{wang2025pi3} removes the dependence on a reference view, VGGT-$\Omega$~\cite{wang2026vggtomega} scales the recipe, and MonST3R~\cite{zhang2025monst3r} adapts it to moving scenes.
Depth Anything 3~\cite{depthanything3} reaches the same outputs from the monocular depth line~\cite{yang2024depthanything,yang2024depthanythingv2,yin2023metric3d,piccinelli2024unidepth,bochkovskii2025depthpro,ke2024marigold}, and the same shift has occurred in SLAM~\cite{teed2021droid} and in feed-forward radiance fields~\cite{hong2024lrm,zhang2024gslrm,charatan2024pixelsplat,chen2024mvsplat,ye2025noposplat,smart2024splatt3r}.
MapAnything~\cite{keetha2025mapanything} and OmniVGGT~\cite{li2026omnivggt} add optional conditioning on intrinsics, extrinsics or depth, which is the closest prior setting to ours.
All of them assume pinhole input: the focal length may vary, the projection model may not.

\subsection{3D Reconstruction for Different Camera Types}

A central camera is described by its ray field, parameterised compactly for rectilinear, fisheye and catadioptric optics~\cite{zhang2000flexible,kannala2006generic,geyer2000unifying,mei2007single,usenko2018double, scaramuzza2006toolbox,sturm2011camera}.
Learning pipelines instead rectify the image to a rectilinear grid, which is lossy at wide field of view and impossible past $180^\circ$, so work on panoramic and fisheye input changes the operator instead: spherical convolutions~\cite{tateno2018distortion,coors2018spherenet,cohen2018spherical,esteves2018learning, su2019kernel}, resampled charts~\cite{cheng2018cube,eder2020tangent}, monocular panoramic depth~\cite{zioulis2018omnidepth,wang2020bifuse,jiang2021unifuse,sun2021hohonet, rey2022360monodepth,li2022omnifusion,shen2022panoformer,yun2023egformer} and multi-view fisheye depth~\cite{won2019sweepnet,komatsu2020360depth,kumar2021svdistnet,yogamani2019woodscape, liao2022kitti360}.
Each fixes one projection, so a change of lens is a change of model.

Recent feed-forward methods attach the camera at the backbone's periphery, by retraining on one projection~\cite{panovggt2026,realsee2026argus}, a token~\cite{jung2026wid3r}, an adapter~\cite{duan2026fisheye3r}, pairwise alignment~\cite{guruprasad2026cam3r} or a cubemap~\cite{bozic2026pager}, or train on renderings through many lenses~\cite{li2026meow}, but keep the pinhole meaning of a rotary coordinate~\cite{su2024roformer} inside the attention, which is what a change of lens alters.
RayTun3R~\cite{sinitsyn2026raytun3r} reaches the same diagnosis but adapts per sequence at test time.
G-ray~\cite{zhang2026gray} parameterises the rotary phase by camera-local ray angles and evaluates it on pinhole and fisheye crops, while we mix eight central camera models from $90^\circ$ to $360^\circ$.
OmniPoint~\cite{ye2026omnipoint} encodes the ray field of the same three projection families from a single image, without relative pose or cross-view attention.
Ray and Pl\"ucker coordinates have also served as geometric position~\cite{sitzmann2021lfn,zhang2024raydiffusion}.
Reusing a pretrained rotary encoding adds a requirement that a freshly trained one lacks: wherever the pretrained coordinate already agrees with the ray angle, it must be reproduced exactly, or a valid prior is lost.

\section{Method}
\label{sec::Method}

\subsection{Preliminary}
\label{sec::setting}

The input is a set of $S$ views $\bm{V}=\{(\bm{I}_s,\mathcal{C}_s)\}_{s=1}^{S}$, where $\bm{I}_s\in\mathbb{R}^{H\times W\times3}$ is an image, $\mathcal{C}_s$ its calibration and each view may come from a different camera model, and a feed-forward network predicts the camera-to-world pose $\bm{T}_s\in SE(3)$ of every view and a point for every pixel.
A camera is central when all its rays pass through one point, so it is described completely by the map $\mathcal{R}:\mathcal{D}\to\mathbb{S}^2$ that sends each pixel $\bm{u}=(u,v)$ of the image domain $\mathcal{D}$ to the unit direction of its ray, expressed in the camera frame with $x$ to the right, $y$ down and $z$ along the optical axis.
We take this map, sampled at the pixel centres, as the camera,
\begin{equation}
\label{eq::raymap}
\mathcal{C}\equiv(\bm{R},\bm{M}),\quad
\bm{R}\in(\mathbb{S}^{2})^{H\times W},\;\;
\bm{M}\in\{0,1\}^{H\times W},
\end{equation}
where the ray field $\bm{R}$ holds the unit ray $\bm{R}_{\bm{u}}=\mathcal{R}(\bm{u})$ of every pixel and $\bm{M}_{\bm{u}}=1$ marks the pixels at which $\mathcal{R}$ is defined, so views of different cameras share one grid.
The network sees no lens type and no distortion coefficient, only $\bm{R}$, which is computed once from the calibration and, as a sample of a differentiable map, has derivatives.
The dense head predicts one distance $d_{\bm{u}}$ per pixel, read along the pixel's ray,
\begin{equation}
\label{eq::rdd}
\bm{X}_{\bm{u}}=d_{\bm{u}}\,\bm{R}_{\bm{u}},
\end{equation}
so one decoder serves every camera, including rays at $90^\circ$ or more from the optical axis.

Any3R is trained to read the ray field of each view (\cref{fig::method}).
Of the panoramic training samples, half are native panoramas and half are views synthesised by resampling a panorama through a randomly drawn camera, sampled uniformly over eight central models, pinhole, Brown--Conrady, equidistant, Kannala--Brandt, MEI, double sphere, equirectangular and cropped equirectangular, with field of view in $[30^\circ,360^\circ]$ and rectilinear models capped below the $180^\circ$ asymptote.
A fresh camera is drawn per view, so one sample routinely mixes a narrow rectilinear crop with a wide fisheye and a full sphere, and the view count $S$ is drawn from $\{2,\dots,8\}$.
The objective is the pretrained backbone's own, with camera, depth and point terms as in~\cite{wang2025vggt}, every term masked by $\bm{M}$, the point weight replaced by the solid-angle weight below and no auxiliary loss.

The backbone is a vision transformer with alternating intra- and inter-frame attention~\cite{wang2025vggt} over a grid of $H_p\times W_p$ patches, and we write $m=\max(H_p,W_p)$.
Two of its parts rely on the pixel grid, the rotary positional encoding and the per-pixel training loss, and Any3R gives $\bm{R}$ to both, as \cref{fig::method} shows.

\begin{table*}[!tb]\centering\footnotesize
\caption{\textbf{Comparison with the methods built for other cameras on the eight camera models used in training.} Acc$\downarrow$ (m) is the mean distance from a reconstructed point to its nearest ground-truth point, and AUC$\uparrow$ (\%) is the area under the curve of the pairwise pose error up to $15^\circ$ and $30^\circ$; every method fuses its views with its own predicted poses. Indoor averages Matterport3D and Stanford2D3DS, and outdoor is PanoCity. The cameras are pinhole (Pin.), Brown--Conrady (B--C), equidistant (Equi.), cropped equirectangular (ERP-C), MEI, KB4, double sphere (DS) and equirectangular (ERP), with their fields of view below them. OmniV and PanoV abbreviate OmniVGGT and PanoVGGT, Wid3R receives the true ray field, MapA-ray is MapAnything with the true ray field fed to its ray encoder, and -- marks cameras a method cannot take. In all tables, \colorbox{best}{\textbf{best}} and \colorbox{second}{second best} are shaded.}
\label{tab::camaware}
\setlength{\tabcolsep}{2.6pt}
\begin{tabular}{lcccccccccccccccc}
\toprule
 & \multicolumn{8}{c}{Indoor} & \multicolumn{8}{c}{Outdoor} \\
\cmidrule(lr){2-9}\cmidrule(lr){10-17}
Method & Pin. & B--C & Equi. & ERP-C & MEI & KB4 & DS & ERP & Pin. & B--C & Equi. & ERP-C & MEI & KB4 & DS & ERP \\
 & {\scriptsize $90^\circ$} & {\scriptsize $90^\circ$} & {\scriptsize $180^\circ$} & {\scriptsize $180^\circ$} & {\scriptsize $180^\circ$} & {\scriptsize $200^\circ$} & {\scriptsize $200^\circ$} & {\scriptsize $360^\circ$} & {\scriptsize $90^\circ$} & {\scriptsize $90^\circ$} & {\scriptsize $180^\circ$} & {\scriptsize $180^\circ$} & {\scriptsize $180^\circ$} & {\scriptsize $200^\circ$} & {\scriptsize $200^\circ$} & {\scriptsize $360^\circ$} \\
\midrule
\multicolumn{17}{l}{\textit{Acc$\downarrow$ (m)}} \\
OmniV~\cite{li2026omnivggt} & 0.250 & 0.201 & 0.326 & 0.318 & 0.329 & 0.355 & 0.350 & 0.406 & 2.875 & 2.676 & 3.313 & 2.804 & 3.703 & 3.941 & 3.670 & 1.479 \\
Wid3R~\cite{jung2026wid3r} & 0.144 & -- & \cellcolor{second}0.141 & -- & \cellcolor{second}0.146 & -- & -- & 0.101 & 2.278 & -- & \cellcolor{second}2.121 & -- & 2.252 & -- & -- & 2.330 \\
PanoV~\cite{panovggt2026} & 0.282 & 0.272 & 0.259 & 0.348 & 0.265 & \cellcolor{second}0.251 & 0.265 & \cellcolor{second}0.072 & 2.256 & 2.708 & 2.498 & 3.633 & 2.433 & \cellcolor{second}2.405 & \cellcolor{second}2.473 & \cellcolor{second}0.751 \\
Argus~\cite{realsee2026argus} & -- & -- & -- & -- & -- & -- & -- & 0.094 & -- & -- & -- & -- & -- & -- & -- & 1.517 \\
MapA-ray~\cite{keetha2025mapanything} & \cellcolor{second}0.092 & \cellcolor{second}0.146 & 0.277 & \cellcolor{second}0.254 & 0.239 & 0.274 & \cellcolor{second}0.258 & 0.405 & \cellcolor{second}1.484 & \cellcolor{second}1.558 & 2.368 & \cellcolor{second}2.686 & \cellcolor{second}2.029 & 2.523 & 2.579 & 1.894 \\
\addlinespace[2pt]
\textbf{Any3R} & \cellcolor{best}\textbf{0.076} & \cellcolor{best}\textbf{0.091} & \cellcolor{best}\textbf{0.058} & \cellcolor{best}\textbf{0.052} & \cellcolor{best}\textbf{0.059} & \cellcolor{best}\textbf{0.056} & \cellcolor{best}\textbf{0.060} & \cellcolor{best}\textbf{0.054} & \cellcolor{best}\textbf{1.160} & \cellcolor{best}\textbf{1.243} & \cellcolor{best}\textbf{0.866} & \cellcolor{best}\textbf{0.811} & \cellcolor{best}\textbf{0.854} & \cellcolor{best}\textbf{0.863} & \cellcolor{best}\textbf{0.901} & \cellcolor{best}\textbf{0.663} \\
\midrule
\multicolumn{17}{l}{\textit{AUC@$15^\circ\uparrow$ (\%)}} \\
OmniV~\cite{li2026omnivggt} & 6.3 & 1.8 & 1.9 & 12.1 & 0.6 & 0.9 & 2.9 & 0.7 & 1.9 & 0.7 & 3.7 & 8.5 & 3.2 & 3.1 & 5.4 & 4.2 \\
Wid3R~\cite{jung2026wid3r} & 46.4 & -- & \cellcolor{second}59.4 & -- & \cellcolor{second}59.5 & -- & -- & 85.3 & 8.1 & -- & 25.5 & -- & 22.9 & -- & -- & 0.0 \\
PanoV~\cite{panovggt2026} & 2.2 & 2.5 & 21.7 & \cellcolor{second}21.5 & 23.6 & \cellcolor{second}36.2 & \cellcolor{second}46.9 & \cellcolor{second}89.2 & 10.8 & 10.1 & \cellcolor{second}37.5 & \cellcolor{second}42.0 & \cellcolor{second}38.1 & \cellcolor{second}51.0 & \cellcolor{second}58.6 & \cellcolor{second}93.5 \\
Argus~\cite{realsee2026argus} & -- & -- & -- & -- & -- & -- & -- & 88.4 & -- & -- & -- & -- & -- & -- & -- & 81.9 \\
MapA-ray~\cite{keetha2025mapanything} & \cellcolor{second}53.6 & \cellcolor{second}31.2 & 10.4 & 6.5 & 15.5 & 12.4 & 12.0 & 4.1 & \cellcolor{second}56.3 & \cellcolor{second}45.0 & 17.5 & 10.0 & 27.8 & 15.1 & 20.0 & 2.1 \\
\addlinespace[2pt]
\textbf{Any3R} & \cellcolor{best}\textbf{74.8} & \cellcolor{best}\textbf{64.8} & \cellcolor{best}\textbf{90.6} & \cellcolor{best}\textbf{86.0} & \cellcolor{best}\textbf{90.5} & \cellcolor{best}\textbf{81.2} & \cellcolor{best}\textbf{80.5} & \cellcolor{best}\textbf{94.5} & \cellcolor{best}\textbf{61.5} & \cellcolor{best}\textbf{51.0} & \cellcolor{best}\textbf{85.0} & \cellcolor{best}\textbf{87.4} & \cellcolor{best}\textbf{85.2} & \cellcolor{best}\textbf{79.5} & \cellcolor{best}\textbf{81.0} & \cellcolor{best}\textbf{97.2} \\
\midrule
\multicolumn{17}{l}{\textit{AUC@$30^\circ\uparrow$ (\%)}} \\
OmniV~\cite{li2026omnivggt} & 13.9 & 5.4 & 6.6 & 26.2 & 5.4 & 4.8 & 7.0 & 4.1 & 9.6 & 4.8 & 9.0 & 19.3 & 9.3 & 10.2 & 13.2 & 6.1 \\
Wid3R~\cite{jung2026wid3r} & 67.0 & -- & \cellcolor{second}76.4 & -- & \cellcolor{second}76.4 & -- & -- & 89.0 & 22.8 & -- & 47.9 & -- & 42.6 & -- & -- & 0.0 \\
PanoV~\cite{panovggt2026} & 10.1 & 11.8 & 50.1 & \cellcolor{second}42.6 & 52.2 & \cellcolor{second}63.5 & \cellcolor{second}69.7 & \cellcolor{second}93.2 & 25.2 & 25.9 & \cellcolor{second}61.0 & \cellcolor{second}62.7 & \cellcolor{second}62.0 & \cellcolor{second}73.9 & \cellcolor{second}78.2 & \cellcolor{second}96.8 \\
Argus~\cite{realsee2026argus} & -- & -- & -- & -- & -- & -- & -- & 91.3 & -- & -- & -- & -- & -- & -- & -- & 90.5 \\
MapA-ray~\cite{keetha2025mapanything} & \cellcolor{second}72.8 & \cellcolor{second}55.5 & 25.5 & 19.7 & 34.9 & 30.2 & 32.8 & 14.1 & \cellcolor{second}74.7 & \cellcolor{second}61.6 & 33.3 & 23.1 & 49.4 & 31.9 & 39.4 & 6.3 \\
\addlinespace[2pt]
\textbf{Any3R} & \cellcolor{best}\textbf{85.0} & \cellcolor{best}\textbf{78.3} & \cellcolor{best}\textbf{93.9} & \cellcolor{best}\textbf{91.6} & \cellcolor{best}\textbf{94.6} & \cellcolor{best}\textbf{89.6} & \cellcolor{best}\textbf{88.5} & \cellcolor{best}\textbf{96.6} & \cellcolor{best}\textbf{80.7} & \cellcolor{best}\textbf{74.2} & \cellcolor{best}\textbf{92.5} & \cellcolor{best}\textbf{93.7} & \cellcolor{best}\textbf{92.6} & \cellcolor{best}\textbf{89.4} & \cellcolor{best}\textbf{90.5} & \cellcolor{best}\textbf{98.6} \\
\bottomrule
\end{tabular}
\end{table*}

\subsection{Angular Coordinates}
\label{sec::arope}

The backbone's rotary encoding~\cite{su2024roformer} locates a patch by its index (\cref{fig::motiv}(ii)), and Any3R locates it by the angles of its ray instead, the angular coordinates at the top left of \cref{fig::method}.
The patch in row $i\in\{0,\dots,H_p-1\}$ and column $j\in\{0,\dots,W_p-1\}$ receives the normalised indices
\begin{equation}
\label{eq::stockrope}
c^{\mathrm{lat}}_{ij}=\frac{2(i+0.5)}{m}-1,\qquad
c^{\mathrm{lon}}_{ij}=\frac{2(j+0.5)}{m}-1 ,
\end{equation}
and the query and key of that patch are rotated by the angles $\omega_k c^{\mathrm{lat}}_{ij}$ and $\omega_k c^{\mathrm{lon}}_{ij}$, each in its own channel pair, with frequencies $\omega_k=2\pi\cdot100^{-k/16}$ for $k=0,\dots,15$.
On a pinhole the index tells roughly where the patch looks, but on a wider lens the same index points elsewhere.
We take the coordinates from the ray field instead.
Let $\bar{\bm{r}}_{ij}=(\bar r_x,\bar r_y,\bar r_z)\in\mathbb{S}^2$ be the mean ray of the patch, obtained by averaging $\bm{R}$ over its pixels and renormalising, with elevation $\theta_{ij}=\arcsin(-\bar r_y)\in[-\tfrac\pi2,\tfrac\pi2]$ and azimuth $\phi_{ij}=\operatorname{atan2}(\bar r_x,\bar r_z)\in(-\pi,\pi]$.
The coordinates are
\begin{equation}
\label{eq::arope}
c^{\mathrm{lat}}_{ij}=\frac{H_p}{m}\!\left(1-\frac{2\theta_{ij}}{\pi}\right)-1,\qquad
c^{\mathrm{lon}}_{ij}=\frac{W_p}{m}\!\left(\frac{\phi_{ij}}{\pi}+1\right)-1 ,
\end{equation}
and every other component of the attention is unchanged.
One unit of coordinate then stands for the same angle on every camera, so the frequencies $\omega_k$, which all views share, mean the same angle in every view of a mixed batch.

The prefactors $H_p/m$ and $W_p/m$ are the unique choice under which \cref{eq::arope} reproduces \cref{eq::stockrope} on an equirectangular grid, the input on which patch index and ray angle agree, since substituting its patch-centre angles $\theta_i=\tfrac{\pi}{2}-\tfrac{\pi(i+0.5)}{H_p}$ and $\phi_j=\tfrac{2\pi(j+0.5)}{W_p}-\pi$ returns \cref{eq::stockrope} term by term, so the pretrained encoding is kept on panoramas.
\cref{eq::arope} is evaluated with $\bm{R}_s$ for each view $s$, so every view of a batch is encoded with its own lens rather than with that of the first view.
The released model applies \cref{eq::arope} to the eight lowest frequencies, $k\ge8$, and for the other eight divides by the half extents $\hat\theta=\tfrac{H_p}{H_p-1}\max_{ij}|\theta_{ij}|$ and $\hat\phi=\tfrac{W_p}{W_p-1}\max_{ij}|\phi_{ij}|$ of the view instead of by $\tfrac\pi2$ and $\pi$, so that these coordinates span each camera's own field of view, as \cref{eq::stockrope} does.
On an equirectangular grid $\hat\theta=\tfrac\pi2$ and $\hat\phi=\pi$, so exactness holds for the split as well, and because the encoder averages the rays of a patch rather than taking its centre, the coordinates of our $518\times1036$ panoramas differ from \cref{eq::stockrope} by at most $4.3\%$ of a patch step.
The backbone also adds $\alpha\,g(\bm{f}_{ij})$ to the token of patch $(i,j)$ before every attention block, with $g$ an MLP, $\alpha$ a learned scalar and $\bm{f}_{ij}=[-\sin\theta,\cos\theta,\sin\phi,\cos\phi]$ evaluated at the equirectangular angles of the patch grid.
We keep $g$ and $\alpha$ and evaluate $\bm{f}_{ij}$ at the angles of $\bar{\bm{r}}_{ij}$, so the absolute position of every token also follows its ray (\cref{fig::method}).

\subsection{Solid-Angle Weighting}
\label{sec::sa}

A pixel also covers a different part of the scene on every lens (\cref{fig::motiv}(i)), as the solid-angle band at the bottom of \cref{fig::method} shows on the panorama.
The backbone's point loss compares the predicted point $\bm{X}_{\bm{u}}\in\mathbb{R}^3$ of each pixel, in the camera frame of its view, with the ground truth $\bm{X}^{\mathrm{gt}}_{\bm{u}}$ and weights each pixel by its radial distance alone,
\begin{equation}
\label{eq::stockw}
\mathcal{L}_{\mathrm{pt}}=\frac{\sum_{\bm{u}}\bm{M}_{\bm{u}}\,w_{\bm{u}}\,
\bigl\lVert\lambda\bm{X}_{\bm{u}}-\bm{X}^{\mathrm{gt}}_{\bm{u}}\bigr\rVert_1}{\sum_{\bm{u}}\bm{M}_{\bm{u}}},\qquad
w_{\bm{u}}=\bigl\lVert\bm{X}^{\mathrm{gt}}_{\bm{u}}\bigr\rVert_2^{-1/2},
\end{equation}
with the sums over the pixels of all views and $\lambda$ the scale, one per sample, that best aligns the prediction with the ground truth under the same weights, since the backbone predicts geometry up to scale.
We also weight each pixel by the solid angle it covers, so the loss integrates over the sphere rather than over the pixel grid of each lens (\cref{fig::method}).
In view $s$ this is the area on $\mathbb{S}^2$ onto which $\mathcal{R}_s$ maps the pixel, the norm of the cross product of its partial derivatives, so it needs no camera model and no extra input,
\begin{equation}
\label{eq::sa}
\begin{gathered}
\mathrm{d}\Omega_{\bm{u}}=\Bigl\lVert\tfrac{\partial\mathcal{R}_s}{\partial u}\times
\tfrac{\partial\mathcal{R}_s}{\partial v}\Bigr\rVert_2 ,\qquad
\tilde{w}_{\bm{u}}=w_{\bm{u}}\cdot\frac{\mathrm{d}\Omega_{\bm{u}}}
{\bigl\langle\mathrm{d}\Omega\bigr\rangle_{\bm{M}}},\\[3pt]
\bigl\langle\mathrm{d}\Omega\bigr\rangle_{\bm{M}}=
\frac{\sum_{\bm{u}}\bm{M}_{\bm{u}}\,\mathrm{d}\Omega_{\bm{u}}}{\sum_{\bm{u}}\bm{M}_{\bm{u}}} ,
\end{gathered}
\end{equation}
and normalising by $\langle\mathrm{d}\Omega\rangle_{\bm{M}}$, the mean over the valid pixels of the same view, puts $\tilde w$ and $w$ on one scale, so the magnitude of \cref{eq::stockw} and with it the learning-rate schedule are unchanged and only the distribution of supervision moves.
Derivatives are finite differences of $\bm{R}$, central inside the image and one-sided at its border, the weight carries no gradient, and before normalising we clamp $\mathrm{d}\Omega$ to the $1$st and $99$th percentiles of the valid pixels in the batch.

\section{Experiments}
\label{sec::exp}
\begin{figure*}[!tb]\centering
\includegraphics[width=\textwidth]{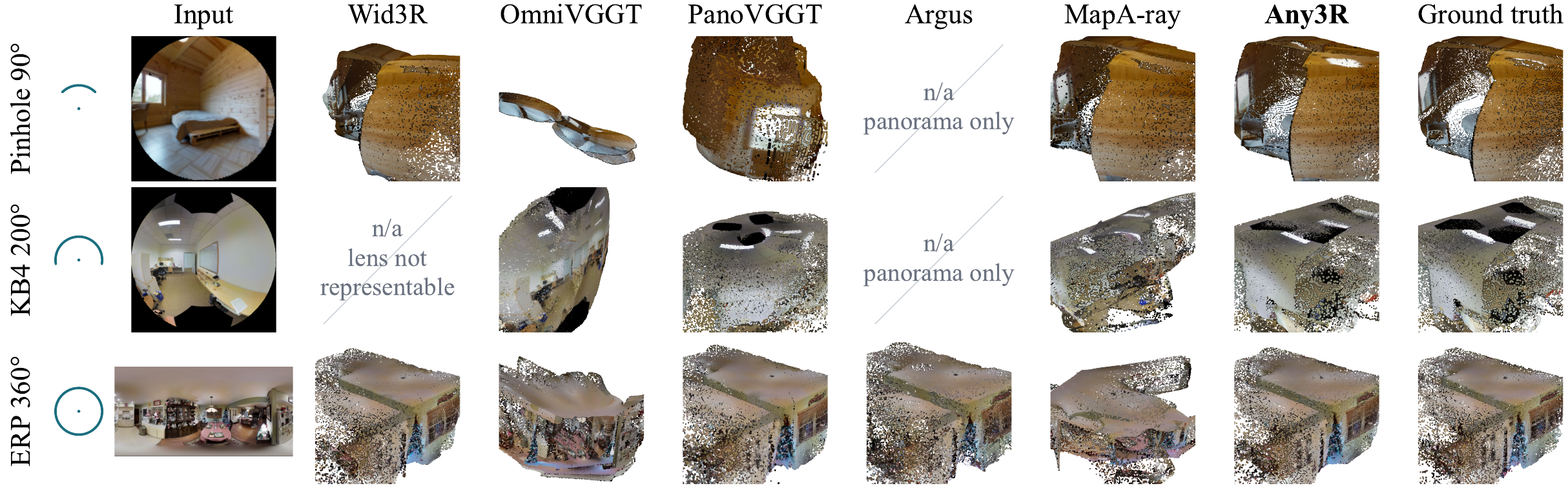}
\caption{\textbf{Qualitative comparison with the camera-aware methods.} One camera per row, every method fused with its own predicted poses and rendered from one viewpoint per row. Cells marked \emph{n/a} are structural, since Argus accepts only full panoramas and Wid3R's camera classes cannot represent the $200^\circ$ KB4 lens. MapA-ray is MapAnything given the true ray field, and PaGeR, which predicts no pose, is not shown.}
\label{fig::rivalsqual}
\end{figure*}

\subsection{Setup}
\label{sec::setup}

We train Any3R for $40$ epochs on $32$ GH200 GPUs over about $27$ hours, on Matterport3D~\cite{chang2017matterport3d}, Stanford2D3DS~\cite{armeni2017joint}, Structured3D~\cite{zhang2022structured3d}, PanoCity and DL3DV~\cite{ling2024dl3dv} videos, holding out all test scenes.
We evaluate on two indoor datasets, Matterport3D and Stanford2D3DS, and one outdoor dataset, PanoCity, with $24$ held-out scenes per dataset and $3$ views per scene.
Every scene is rendered through $13$ central camera models, the eight used in training and five held out, from the same panoramas and view directions, so two results differ by the lens alone, and \cref{tab::camaware,tab::gfm} report the eight.

Every method fuses its views into one point cloud with its own predicted camera poses, and we measure both the cloud and the poses.
Accuracy (Acc) is the mean distance from a reconstructed point to its nearest ground-truth point after the similarity alignment and ICP refinement of the PanoVGGT evaluation~\cite{panovggt2026}, the point-cloud accuracy used across feed-forward reconstruction~\cite{wang2025cut3r,wang2025pi3}.
Pose accuracy is the area under the curve (AUC) of the pairwise pose error, the larger of the relative rotation error and the translation-direction error, up to $15^\circ$ and $30^\circ$, as in~\cite{wang2025vggt,wang2025pi3}.
Any3R receives the ray map, MapAnything a pinhole intrinsic matrix on the pinhole cameras, Wid3R the true ray field and the other methods images alone, and MapA-ray is MapAnything given the true ray field on every camera.

\subsection{Comparison with the State of the Art}
\label{sec::cmp}

\paragraph{Camera-aware methods.} \cref{tab::camaware} compares Any3R with the methods built for other cameras on the same scenes and views, and each of them covers part of the table.
\begin{table}[!tb]\centering\footnotesize
\caption{\textbf{Zero-shot reconstruction and camera pose on real fisheye lenses.} ScanNet++ is its DSLR fisheye on its validation split and FIORD the $200^\circ$ lens of its rig; neither is in Any3R's training data. Measures and protocol as in \cref{tab::camaware}; Any3R, MapA-ray and Wid3R receive the calibrated ray field, and -- marks a lens a method cannot take. $^\dagger$Trained on ScanNet++. Any3R$^\ast$ is not ranked.}
\label{tab::zeroshot}
\setlength{\tabcolsep}{3pt}
\begin{tabular}{lcccccc}
\toprule
 & \multicolumn{3}{c}{ScanNet++} & \multicolumn{3}{c}{FIORD} \\
\cmidrule(lr){2-4}\cmidrule(lr){5-7}
Method & Acc$\downarrow$ & @15$\uparrow$ & @30$\uparrow$ & Acc$\downarrow$ & @15$\uparrow$ & @30$\uparrow$ \\
\midrule
OmniV~\cite{li2026omnivggt} & 0.097 & 18.6 & 41.5 & \cellcolor{second}0.363 & 18.6 & 40.3 \\
Wid3R$^\dagger$~\cite{jung2026wid3r} & \cellcolor{best}\textbf{0.039} & \cellcolor{best}\textbf{85.3} & \cellcolor{best}\textbf{92.6} & -- & -- & -- \\
PanoV~\cite{panovggt2026} & 0.277 & 1.8 & 9.7 & 0.511 & \cellcolor{second}29.4 & \cellcolor{second}51.2 \\
MapA-ray~\cite{keetha2025mapanything} & 0.093 & 18.5 & 43.6 & 0.396 & 19.4 & 36.5 \\
\midrule
$\pi^3$~\cite{wang2025pi3} & 0.086 & 20.6 & 46.6 & 0.378 & 15.7 & 36.7 \\
MapA~\cite{keetha2025mapanything} & 0.099 & 19.3 & 40.2 & 0.399 & 17.7 & 37.2 \\
DA3~\cite{depthanything3} & 0.118 & 17.2 & 41.5 & 0.380 & 9.1 & 25.8 \\
VGGT~\cite{wang2025vggt} & 0.118 & 18.0 & 41.7 & 0.442 & 15.6 & 35.3 \\
V-$\Omega$~\cite{wang2026vggtomega} & 0.112 & 16.7 & 42.9 & 0.451 & 19.9 & 38.9 \\
\midrule
Any3R$^\ast$ & 0.273 & 7.5 & 22.9 & 0.480 & 28.3 & 47.5 \\
\textbf{Any3R} & \cellcolor{second}0.055 & \cellcolor{second}62.7 & \cellcolor{second}79.7 & \cellcolor{best}\textbf{0.274} & \cellcolor{best}\textbf{63.8} & \cellcolor{best}\textbf{78.4} \\
\bottomrule
\end{tabular}
\end{table}

\begin{table*}[!tb]\centering\footnotesize
\caption{\textbf{Comparison with geometry foundation models on the eight camera models used in training.} Measures, domains and cameras are those of \cref{tab::camaware}. MapA and V-$\Omega$ abbreviate MapAnything and VGGT-$\Omega$, and Any3R$^\ast$ is Any3R without its ray map, which assumes an equirectangular grid, and is not ranked.}
\label{tab::gfm}
\setlength{\tabcolsep}{2.6pt}
\begin{tabular}{lcccccccccccccccc}
\toprule
 & \multicolumn{8}{c}{Indoor} & \multicolumn{8}{c}{Outdoor} \\
\cmidrule(lr){2-9}\cmidrule(lr){10-17}
Method & Pin. & B--C & Equi. & ERP-C & MEI & KB4 & DS & ERP & Pin. & B--C & Equi. & ERP-C & MEI & KB4 & DS & ERP \\
 & {\scriptsize $90^\circ$} & {\scriptsize $90^\circ$} & {\scriptsize $180^\circ$} & {\scriptsize $180^\circ$} & {\scriptsize $180^\circ$} & {\scriptsize $200^\circ$} & {\scriptsize $200^\circ$} & {\scriptsize $360^\circ$} & {\scriptsize $90^\circ$} & {\scriptsize $90^\circ$} & {\scriptsize $180^\circ$} & {\scriptsize $180^\circ$} & {\scriptsize $180^\circ$} & {\scriptsize $200^\circ$} & {\scriptsize $200^\circ$} & {\scriptsize $360^\circ$} \\
\midrule
\multicolumn{17}{l}{\textit{Acc$\downarrow$ (m)}} \\
$\pi^3$~\cite{wang2025pi3} & \cellcolor{second}0.084 & \cellcolor{second}0.119 & \cellcolor{second}0.171 & 0.287 & \cellcolor{second}0.164 & \cellcolor{second}0.189 & \cellcolor{second}0.183 & 0.414 & \cellcolor{second}1.479 & 1.711 & 2.093 & \cellcolor{second}2.426 & \cellcolor{second}2.036 & \cellcolor{second}2.394 & \cellcolor{second}2.271 & \cellcolor{second}1.628 \\
MapA~\cite{keetha2025mapanything} & 0.092 & 0.149 & 0.273 & 0.269 & 0.267 & 0.283 & 0.294 & \cellcolor{second}0.386 & 1.485 & 1.633 & \cellcolor{second}2.066 & 2.861 & 2.236 & 3.297 & 2.789 & 1.833 \\
DA3~\cite{depthanything3} & 0.313 & 0.294 & 0.361 & 0.288 & 0.363 & 0.396 & 0.379 & 0.402 & 4.360 & 5.689 & 6.656 & 2.717 & 6.844 & 6.463 & 6.477 & 3.149 \\
VGGT~\cite{wang2025vggt} & 0.270 & 0.282 & 0.361 & 0.274 & 0.379 & 0.356 & 0.374 & 0.398 & 4.153 & 3.940 & 6.296 & 3.078 & 6.886 & 6.623 & 6.430 & 1.902 \\
V-$\Omega$~\cite{wang2026vggtomega} & 0.149 & 0.161 & 0.252 & \cellcolor{second}0.265 & 0.250 & 0.281 & 0.302 & 0.403 & 1.711 & \cellcolor{second}1.572 & 2.162 & 2.455 & 2.490 & 2.680 & 2.544 & 1.876 \\
\addlinespace[2pt]
Any3R$^\ast$ & 0.282 & 0.283 & 0.274 & 0.278 & 0.278 & 0.279 & 0.298 & 0.054 & 2.079 & 2.525 & 2.370 & 3.324 & 2.463 & 2.420 & 2.552 & 0.663 \\
\textbf{Any3R} & \cellcolor{best}\textbf{0.076} & \cellcolor{best}\textbf{0.091} & \cellcolor{best}\textbf{0.058} & \cellcolor{best}\textbf{0.052} & \cellcolor{best}\textbf{0.059} & \cellcolor{best}\textbf{0.056} & \cellcolor{best}\textbf{0.060} & \cellcolor{best}\textbf{0.054} & \cellcolor{best}\textbf{1.160} & \cellcolor{best}\textbf{1.243} & \cellcolor{best}\textbf{0.866} & \cellcolor{best}\textbf{0.811} & \cellcolor{best}\textbf{0.854} & \cellcolor{best}\textbf{0.863} & \cellcolor{best}\textbf{0.901} & \cellcolor{best}\textbf{0.663} \\
\midrule
\multicolumn{17}{l}{\textit{AUC@$15^\circ\uparrow$ (\%)}} \\
$\pi^3$~\cite{wang2025pi3} & \cellcolor{best}\textbf{76.9} & \cellcolor{second}53.5 & \cellcolor{second}42.1 & 18.0 & \cellcolor{second}43.1 & \cellcolor{second}35.2 & \cellcolor{second}34.6 & 3.5 & 44.6 & 34.0 & 18.8 & \cellcolor{second}15.9 & \cellcolor{second}20.0 & 14.4 & \cellcolor{second}16.0 & 2.3 \\
MapA~\cite{keetha2025mapanything} & 53.6 & 28.5 & 13.2 & 9.0 & 10.8 & 8.8 & 8.4 & 6.4 & \cellcolor{second}56.5 & 20.6 & \cellcolor{second}19.6 & 10.6 & 17.1 & \cellcolor{second}16.9 & 13.8 & 0.1 \\
DA3~\cite{depthanything3} & 7.8 & 2.7 & 0.9 & \cellcolor{second}22.1 & 1.2 & 0.4 & 0.7 & \cellcolor{second}7.0 & 4.8 & 3.0 & 6.4 & 7.7 & 4.9 & 3.4 & 4.6 & \cellcolor{second}5.5 \\
VGGT~\cite{wang2025vggt} & 23.3 & 9.9 & 5.2 & 14.2 & 5.3 & 5.0 & 2.9 & 3.3 & 15.1 & 2.6 & 3.7 & 9.5 & 6.5 & 5.5 & 7.3 & 0.9 \\
V-$\Omega$~\cite{wang2026vggtomega} & 63.5 & 50.4 & 31.3 & 15.0 & 29.6 & 24.0 & 19.5 & 6.9 & 45.5 & \cellcolor{second}44.4 & 15.8 & 8.1 & 15.6 & 13.8 & 8.8 & 3.9 \\
\addlinespace[2pt]
Any3R$^\ast$ & 24.0 & 43.8 & 48.3 & 41.7 & 48.2 & 55.0 & 58.5 & 94.5 & 20.1 & 36.0 & 45.0 & 29.9 & 47.3 & 48.8 & 50.1 & 97.2 \\
\textbf{Any3R} & \cellcolor{second}74.8 & \cellcolor{best}\textbf{64.8} & \cellcolor{best}\textbf{90.6} & \cellcolor{best}\textbf{86.0} & \cellcolor{best}\textbf{90.5} & \cellcolor{best}\textbf{81.2} & \cellcolor{best}\textbf{80.5} & \cellcolor{best}\textbf{94.5} & \cellcolor{best}\textbf{61.5} & \cellcolor{best}\textbf{51.0} & \cellcolor{best}\textbf{85.0} & \cellcolor{best}\textbf{87.4} & \cellcolor{best}\textbf{85.2} & \cellcolor{best}\textbf{79.5} & \cellcolor{best}\textbf{81.0} & \cellcolor{best}\textbf{97.2} \\
\midrule
\multicolumn{17}{l}{\textit{AUC@$30^\circ\uparrow$ (\%)}} \\
$\pi^3$~\cite{wang2025pi3} & \cellcolor{second}82.7 & 66.0 & \cellcolor{second}61.9 & 37.1 & \cellcolor{second}62.7 & \cellcolor{second}56.4 & \cellcolor{second}58.9 & 14.1 & 61.7 & 47.1 & 36.1 & \cellcolor{second}28.3 & 37.3 & 30.0 & \cellcolor{second}35.0 & 4.8 \\
MapA~\cite{keetha2025mapanything} & 72.8 & 52.3 & 31.3 & 23.2 & 29.6 & 27.6 & 24.8 & 16.2 & \cellcolor{second}74.8 & 36.1 & \cellcolor{second}41.9 & 25.0 & \cellcolor{second}39.7 & \cellcolor{second}34.2 & 30.1 & 3.1 \\
DA3~\cite{depthanything3} & 17.8 & 10.3 & 5.7 & \cellcolor{second}46.2 & 7.2 & 5.5 & 5.1 & 16.8 & 13.6 & 11.5 & 14.8 & 22.1 & 10.2 & 9.8 & 10.6 & \cellcolor{second}11.3 \\
VGGT~\cite{wang2025vggt} & 50.0 & 25.6 & 18.8 & 32.2 & 16.6 & 16.4 & 13.9 & 10.6 & 32.8 & 11.0 & 8.8 & 24.3 & 13.9 & 13.2 & 17.3 & 3.5 \\
V-$\Omega$~\cite{wang2026vggtomega} & 76.9 & \cellcolor{second}69.7 & 53.8 & 34.6 & 52.3 & 47.1 & 42.0 & \cellcolor{second}17.8 & 55.3 & \cellcolor{second}54.7 & 33.8 & 23.6 & 34.8 & 31.4 & 24.4 & 8.5 \\
\addlinespace[2pt]
Any3R$^\ast$ & 47.5 & 66.2 & 71.3 & 65.8 & 72.0 & 75.9 & 77.5 & 96.6 & 38.9 & 58.3 & 67.7 & 53.7 & 69.6 & 72.3 & 72.2 & 98.6 \\
\textbf{Any3R} & \cellcolor{best}\textbf{85.0} & \cellcolor{best}\textbf{78.3} & \cellcolor{best}\textbf{93.9} & \cellcolor{best}\textbf{91.6} & \cellcolor{best}\textbf{94.6} & \cellcolor{best}\textbf{89.6} & \cellcolor{best}\textbf{88.5} & \cellcolor{best}\textbf{96.6} & \cellcolor{best}\textbf{80.7} & \cellcolor{best}\textbf{74.2} & \cellcolor{best}\textbf{92.5} & \cellcolor{best}\textbf{93.7} & \cellcolor{best}\textbf{92.6} & \cellcolor{best}\textbf{89.4} & \cellcolor{best}\textbf{90.5} & \cellcolor{best}\textbf{98.6} \\
\bottomrule
\end{tabular}
\end{table*}

Wid3R represents each camera by one of a few classes, which hold only the pinhole, equidistant, MEI and equirectangular cameras exactly and none of the $200^\circ$ lenses, and it receives the ray field only after the transformer, where it can correct the decoded points but neither the attention nor the poses.
OmniVGGT conditions on a pinhole field of view below $180^\circ$, which cannot describe these lenses, so it runs on images alone and reads them through VGGT's index encoding.
PanoVGGT and Argus are built for panoramas, and off the panorama PanoVGGT reads every frame as one while Argus cannot run.
MapAnything accepts a dense ray map, and MapA-ray gives it the true one on every camera, but the map is added to the image features after a pinhole encoder has read the frame on its pixel grid, and its accuracy stays close to that of MapAnything without it in \cref{tab::gfm}.
On every camera and dataset that each of these methods can take, Any3R is more accurate and better posed in all three measures, including the equirectangular frame of the panoramic models, where our coordinates coincide with the stock encoding, so the other lenses were added without taking anything from it.
In the typical scenes of \cref{fig::rivalsqual}, whose Acc for Any3R lies between the quartiles of its cell, PanoVGGT bends a pinhole frame into a bowl and MapA-ray still folds the equirectangular room despite the true ray field, while Any3R returns the room on every lens.
Beyond these comparisons, Any3R is one general model, and the same weights fill every column of \cref{tab::camaware}, where each method built for one family of lenses covers a subset.

\begin{figure*}[!tb]\centering
\includegraphics[width=\textwidth]{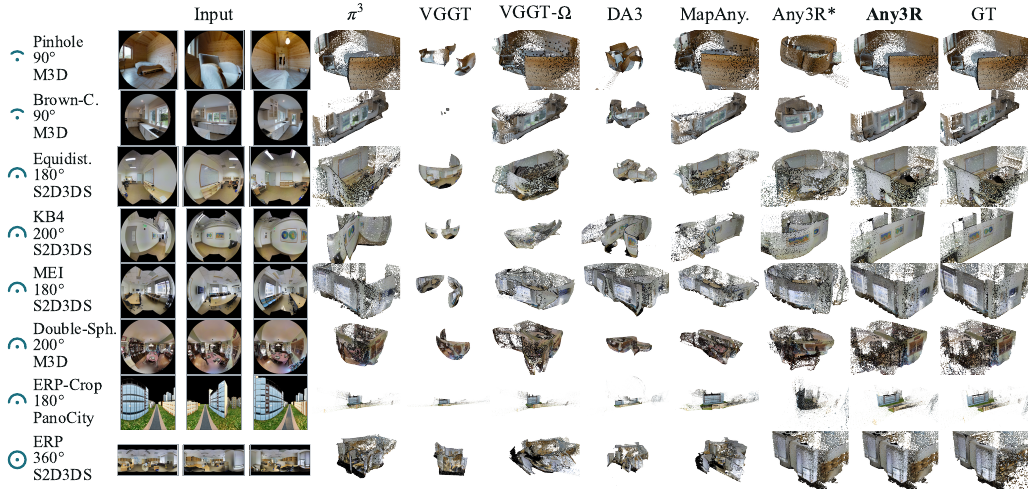}
\caption{\textbf{Eight central camera models, seven methods, one scene per row.} Every method fuses the same three native views with its own predicted poses and is placed in the ground-truth frame by one similarity transform, and the highest $30\%$ of each indoor cloud is removed so the room is seen from above. Column \textit{MapAny.} is MapAnything and Any3R$^\ast$ is Any3R without its ray map.}
\label{fig::cams}
\end{figure*}

\paragraph{Geometry foundation models.} \cref{tab::gfm} compares Any3R with $\pi^3$, MapAnything, DA3, VGGT and VGGT-$\Omega$, which are built for pinhole images, so the $90^\circ$ columns are their home ground and the place where a model that serves every lens would pay for its generality.
Any3R does not pay there, and it is more accurate and better posed than every one of them on every camera, indoors and outdoors, except for the AUC at $15^\circ$ on the indoor pinhole, where $\pi^3$ leads because it is ahead on Stanford2D3DS.
Away from that ground their pose collapses, with an AUC at $30^\circ$ below a fifth at $360^\circ$, and indoors each of them is least accurate at $360^\circ$, whereas Any3R is more accurate on every wider lens than at $90^\circ$.
A pinhole model can infer where a patch looks only from its index, through the tangent mapping of the images it was trained on, so it reads a wider lens as a pinhole and places the rays that look sideways or backwards inside its frustum, whereas a wider lens only gives our angular coordinates more of the room to place.
In the typical scenes of \cref{fig::cams}, from $180^\circ$ on they place the parts of a frame that look sideways or backwards in front of the camera, flattening the room or folding it into a shell, and VGGT and DA3 often lose the relative pose as well.
Any3R$^\ast$, our model without its ray map, assumes an equirectangular grid and reads every frame as a panorama, and its error rises sharply on every lens but the equirectangular one, where that assumption holds and the two rows coincide, so the gain comes from the ray map; in \cref{fig::cams} it bends a pinhole frame into a bowl.

\subsection{Zero-Shot Real Lenses}
\label{sec::zeroshot}
The cameras above are rendered from panoramas, so \cref{tab::zeroshot} also tests real lenses on two datasets that are not in our training data, under the same protocol.
\begin{figure}[!tb]\centering
\includegraphics[width=\columnwidth]{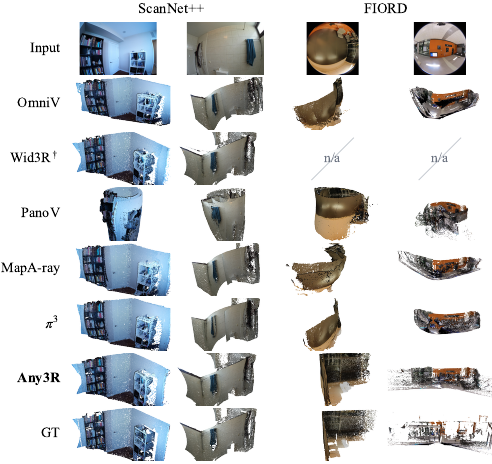}
\caption{\textbf{Zero-shot reconstruction on real fisheye lenses.} Per dataset, the two samples from different scenes on which Any3R leads the zero-shot methods shown most (Wid3R$^\dagger$ is trained on ScanNet++). Poses and alignment as in \cref{tab::zeroshot}; the highest $30\%$ of each cloud is removed so the room is seen from above.}
\label{fig::zeroshot}
\end{figure}
ScanNet++~\cite{yeshwanth2023scannetpp} records each scene with a DSLR fisheye of about $146^\circ$, and we take one view triplet from each scene of its validation split, with ground truth from the laser-scanned mesh.
FIORD~\cite{gunes2025fiord} records with a $200^\circ$ fisheye whose image reaches beyond $90^\circ$ from its axis, and we take $24$ view triplets from two of its scenes, with ground truth from a terrestrial laser scan.
Any3R, MapA-ray and Wid3R receive the ray field of the calibrated lens.
On FIORD, Any3R is the most accurate and the best posed of the methods compared, with an AUC at $30^\circ$ about half again that of the next one, and Wid3R cannot represent the lens.
On ScanNet++, Any3R is second only to Wid3R and ahead of every other method in all three measures.
Without its ray map, Any3R$^\ast$ falls far behind on both datasets, so on real lenses as well the gain comes from the ray map.
In \cref{fig::zeroshot}, the other zero-shot methods bend the FIORD rooms into curved shells, while Any3R stays close to the ground truth.

\subsection{Ablation}
\label{sec::abl}

\begin{table}[tb]\centering\footnotesize
\setlength{\tabcolsep}{1.6pt}
\caption{\textbf{Module ablation on full sequences} under the PanoVGGT protocol (Structured3D, with no multi-view overlap, is not reported). AUC at $15^\circ$ and $\overline{\mathrm{R}}_\text{med}$, the per-sequence median rotation error averaged over sequences; the row marked $^{*}$ has the warm-start architecture and is not directly comparable.}
\label{tab::abl}
\begin{tabular}{ccc *{6}{c}}
\toprule
\multicolumn{3}{c}{Component} & \multicolumn{2}{c}{M3D} & \multicolumn{2}{c}{S2D3DS} & \multicolumn{2}{c}{PanoCity} \\
\cmidrule(lr){1-3}\cmidrule(lr){4-5}\cmidrule(lr){6-7}\cmidrule(lr){8-9}
Ang. & \begin{tabular}[b]{@{}c@{}}Per-\\view\end{tabular} & \begin{tabular}[b]{@{}c@{}}Solid-\\ang.\end{tabular} & AUC$\uparrow$ & $\overline{\mathrm{R}}_\text{med}\downarrow$ & AUC$\uparrow$ & $\overline{\mathrm{R}}_\text{med}\downarrow$ & AUC$\uparrow$ & $\overline{\mathrm{R}}_\text{med}\downarrow$ \\
\midrule
-- & -- & \cmark$^{*}$ & 0.619 & 4.61 & 0.636 & 4.18 & 0.854 & 0.77 \\
\midrule
\cmark & -- & -- & 0.364 & 9.26 & 0.381 & 8.94 & 0.653 & 1.88 \\
\cmark & -- & \cmark & 0.380 & 8.25 & 0.395 & 8.50 & \cellcolor{best}\textbf{0.686} & \cellcolor{second}1.72 \\
\cmark & \cmark & -- & \cellcolor{second}0.488 & \cellcolor{best}\textbf{6.36} & \cellcolor{second}0.527 & \cellcolor{best}\textbf{6.44} & 0.633 & \cellcolor{best}\textbf{1.41} \\
\cmark & \cmark & \cmark & \cellcolor{best}\textbf{0.492} & \cellcolor{second}7.13 & \cellcolor{best}\textbf{0.528} & \cellcolor{second}6.90 & \cellcolor{second}0.669 & \cellcolor{best}\textbf{1.41} \\
\bottomrule
\end{tabular}\end{table}

The angular coordinates, the absolute pathway and radial decoding read the same ray field and cannot be switched off one at a time, so Any3R$^\ast$ in \cref{tab::gfm}, without the ray map, is their joint ablation, and \cref{tab::abl} isolates the two remaining choices on full panoramic sequences under the PanoVGGT protocol.
Per-view evaluation concerns what the angular coordinates read in a sample whose views come from different cameras.
With it, the coordinates of each view are computed from that view's own ray field, and without it, from the ray field of the first view, which all views of the sample then share.
Every test view of \cref{tab::abl} is a panorama, so the two inputs give identical coordinates there, and the difference between the rows was learned in training, where the shared field assigns every view but the first the angles of a lens it was not taken with and the network learns to relate views through wrong angles.
Reading each view's own field gives the largest pose gain on the indoor datasets, and on PanoCity it lowers the median rotation error but also the AUC.
On its own, solid-angle weighting gives a smaller gain by counting each direction of the scene once instead of each pixel.
The pose gains cost some depth accuracy, most of it with per-view evaluation, which raises the mean $\log_{10}$ depth error on Matterport3D from $0.033$ to $0.045$.
We accept this trade for the larger gain in pose.

\FloatBarrier   %
\section{Conclusion}
\label{sec::conclusion}
Geometry foundation models typically assume pinhole cameras, an assumption embedded in their positional encoding and training objectives. Any3R conditions on dense ray fields through angular coordinates and solid-angle weighting, adapting each camera to the backbone to enable a unified model across central camera types. This feed-forward approach reconstructs scenes from arbitrary combinations of central cameras without calibration search, image undistortion, or lens-specific retraining. Evaluated on thirteen central camera models across three datasets, Any3R achieves state-of-the-art reconstruction accuracy using its own predicted camera poses. Any3R enables scalable 3D reconstruction from heterogeneous image collections, including crowdsourced and archival imagery captured by smartphones, fisheye surveillance cameras, and panoramic imaging systems.

\FloatBarrier   %
\section*{Acknowledgments}
This work was supported by the U.S.\ National Science Foundation under Grant No.~2316400.
This work was also supported by the Wallenberg AI, Autonomous Systems and Software Program (WASP) funded by the Knut and Alice Wallenberg Foundation.
The computations and data handling were enabled by the Arrhenius HPC resource provided by National Academic Infrastructure for Supercomputing in Sweden. The authors also acknowledge the South Dakota State University High Performance Computing (SDSU HPC) team for providing access to and support for the university's high-performance computing resources.

{\small\bibliographystyle{ieeenat_fullname}\bibliography{references}}

\maketitlesupplementary \appendix
\makeatletter\def\@flsetnum#1{}\makeatother
\setcounter{dbltopnumber}{2}
\setcounter{dblbotnumber}{0}
\makeatletter\setlength{\@dblfptop}{0pt}\setlength{\@fptop}{0pt}\makeatother

\section{Evaluation Details}
\label{sec::evaldetails}
Each of the $24$ scenes of a dataset is a set of three views drawn without replacement, whose view directions are fixed by a pre-registered rule on the $90^\circ$ pinhole and reused for every camera.
The thirteen camera models are the eight used in training and five held out from it, FOV/ATAN at $150^\circ$, stereographic at $180^\circ$, equisolid and EUCM at $200^\circ$ and cylindrical at $360^\circ$.
Structured3D is evaluated in the same way on all $16$ of its test scenes and reported only in this supplement, since the views of a scene share no content and relative pose cannot be recovered there.

Each method's per-view points are placed with its own predicted poses and fused, and $40{,}000$ pixels are sampled per scene.
The evaluation function of PanoVGGT then fits one similarity transform on the pixel correspondences between the predicted and the ground-truth points, refines it by point-to-point ICP with a maximum correspondence distance of $0.1$\,m indoors and $1.0$\,m on PanoCity, and reports Acc, the mean distance from a reconstructed point to its nearest ground-truth point, and a cell's value is the mean over its scenes.
The AUC follows the authors' pose evaluation, taking for every pair of views the larger of the relative rotation error and the translation-direction error, in $1^\circ$ bins, and dropping pairs whose ground-truth translation is below $0.01$.
Baselines with two input options, $\pi^3$, VGGT and VGGT-$\Omega$ at their native or their official resolution and MapAnything with or without the pinhole intrinsics, are shown with the option of lower median Acc in each cell, and \cref{tab::p13arms} lists both.
MapAnything accepts a dense ray map, but its released preprocessing fits a pinhole intrinsic matrix to any ray map before cropping, so MapA-ray takes the images through that preprocessing and then receives, for every processed pixel, the true ray of the camera, which enters its ray encoder unchanged.

The zero-shot samples of \cref{tab::zeroshot} use the calibrated Kannala--Brandt model of each lens, inverted per pixel, with the images resized to the size of the rendered cameras of similar field of view.
On ScanNet++ the views are the DSLR images of every scene of the validation split, the full set with ground truth since the test splits publish no laser scan, posed by its COLMAP reconstruction, which lies in the frame of the laser scan, and the ground truth is the range along each ray to the scanned mesh.
On FIORD the views are those of the second lens of the rig, posed by its COLMAP reconstruction brought into the frame of the laser scan by a similarity transform, and the ground truth is the laser cloud z-buffered into each view; the scenes are meetingroom and parakennus, and upstairs is left out because its reconstruction does not register to its scan.
In each scene the second view is the first one after an anchor that differs from it by at least $15^\circ$ or $0.3$\,m and sees at least $30\%$ of its ground truth, and the third the first such view before the anchor that also overlaps the second, a rule fixed before any method was run.
Wid3R receives the lens through its Fisheye624 class, which contains this model, and the rays it consumes are checked per pixel against the calibrated ones.
Where this supplement counts a method as best in a cell, a pre-registered one-sided per-scene sign test also separates it from the runner-up at $p<0.05$.

\section{Further Results on the Eight Camera Models}

\providecommand{\ppWidCells}{}\renewcommand{\ppWidCells}{12}
\providecommand{\ppWidOursBest}{}\renewcommand{\ppWidOursBest}{11}
\providecommand{\ppWidOursTied}{}\renewcommand{\ppWidOursTied}{1}
\providecommand{\ppWidRayBeatsPred}{}\renewcommand{\ppWidRayBeatsPred}{10}
\providecommand{\ppWidOfficialChosen}{}\renewcommand{\ppWidOfficialChosen}{14}
\providecommand{\ppWidSthreeCells}{}\renewcommand{\ppWidSthreeCells}{4}
\providecommand{\ppWidSthreeOursBest}{}\renewcommand{\ppWidSthreeOursBest}{2}
\providecommand{\ppWidSthreeOursTied}{}\renewcommand{\ppWidSthreeOursTied}{2}
\providecommand{\ppWidSthreeRayBeatsPred}{}\renewcommand{\ppWidSthreeRayBeatsPred}{4}
\providecommand{\ppWidSthreeOfficialChosen}{}\renewcommand{\ppWidSthreeOfficialChosen}{5}
\providecommand{\pinOurs}{}\renewcommand{\pinOurs}{0.0690}
\providecommand{\nocalPin}{}\renewcommand{\nocalPin}{0.2928}
\providecommand{\pinPi}{}\renewcommand{\pinPi}{0.1093}
\providecommand{\erpOurs}{}\renewcommand{\erpOurs}{0.0447}
\providecommand{\erpPi}{}\renewcommand{\erpPi}{0.4190}
\providecommand{\ppScenes}{}\renewcommand{\ppScenes}{24}
\providecommand{\ppDatasets}{}\renewcommand{\ppDatasets}{3}
\providecommand{\ppMainCells}{}\renewcommand{\ppMainCells}{24}
\providecommand{\ppMainOursBest}{}\renewcommand{\ppMainOursBest}{18}
\providecommand{\ppMainOursTied}{}\renewcommand{\ppMainOursTied}{5}
\providecommand{\ppMainOursBehind}{}\renewcommand{\ppMainOursBehind}{1}
\providecommand{\ppMixCells}{}\renewcommand{\ppMixCells}{6}
\providecommand{\ppMixOursBest}{}\renewcommand{\ppMixOursBest}{3}
\providecommand{\ppMixOursTied}{}\renewcommand{\ppMixOursTied}{2}
\providecommand{\ppMixOursBehind}{}\renewcommand{\ppMixOursBehind}{1}

\providecommand{\ppSthreeScenes}{}\renewcommand{\ppSthreeScenes}{16}
\providecommand{\ppSthreeCells}{}\renewcommand{\ppSthreeCells}{8}
\providecommand{\ppSthreeOursBest}{}\renewcommand{\ppSthreeOursBest}{6}
\providecommand{\ppSthreeOursTied}{}\renewcommand{\ppSthreeOursTied}{1}
\providecommand{\ppSthreeOursBehind}{}\renewcommand{\ppSthreeOursBehind}{1}

\begin{table*}[!tb]\centering\footnotesize
\setlength{\tabcolsep}{1.4pt}
\caption{\textbf{Reconstruction accuracy on the Any3R benchmark.} Acc$\downarrow$ (m) for each camera and dataset, every method with its own predicted poses. In all tables \colorbox{best}{\textbf{best}} and \colorbox{second}{second best} are shaded, and MapA, V-$\Omega$, PanoV, M3D, S2D3DS and S3D abbreviate MapAnything, VGGT-$\Omega$, PanoVGGT, Matterport3D, Stanford2D3DS and Structured3D.}
\label{tab::main}
\begin{tabular}{lcccccccccccccc}
\toprule
 & \multicolumn{7}{c}{Matterport3D} & \multicolumn{7}{c}{Stanford2D3DS} \\
\cmidrule(lr){2-8}\cmidrule(lr){9-15}
Camera (FoV) & Any3R & $\pi^3${\footnotesize\cite{wang2025pi3}} & MapA{\footnotesize\cite{keetha2025mapanything}} & DA3{\footnotesize\cite{depthanything3}} & V-$\Omega${\footnotesize\cite{wang2026vggtomega}} & VGGT{\footnotesize\cite{wang2025vggt}} & PanoV{\footnotesize\cite{panovggt2026}} & Any3R & $\pi^3$ & MapA & DA3 & V-$\Omega$ & VGGT & PanoV \\
\midrule
Pinhole (90$^\circ$) & \cellcolor{best}\textbf{0.069} & 0.109 & \cellcolor{second}0.089 & 0.300 & 0.161 & 0.288 & 0.281 & \cellcolor{second}0.083 & \cellcolor{best}\textbf{0.059} & 0.095 & 0.325 & 0.138 & 0.252 & 0.284 \\
Brown--C. (90$^\circ$) & \cellcolor{best}\textbf{0.096} & \cellcolor{second}0.117 & 0.150 & 0.284 & 0.139 & 0.320 & 0.261 & \cellcolor{best}\textbf{0.085} & \cellcolor{second}0.120 & 0.149 & 0.303 & 0.184 & 0.243 & 0.284 \\
Equidist. (180$^\circ$) & \cellcolor{best}\textbf{0.056} & \cellcolor{second}0.166 & 0.246 & 0.360 & 0.215 & 0.343 & 0.248 & \cellcolor{best}\textbf{0.061} & \cellcolor{second}0.177 & 0.299 & 0.363 & 0.288 & 0.380 & 0.270 \\
ERP-Cr. (180$^\circ$) & \cellcolor{best}\textbf{0.051} & 0.265 & \cellcolor{second}0.242 & 0.267 & 0.245 & 0.261 & 0.331 & \cellcolor{best}\textbf{0.053} & 0.310 & 0.296 & 0.309 & \cellcolor{second}0.286 & 0.287 & 0.364 \\
MEI (180$^\circ$) & \cellcolor{best}\textbf{0.054} & \cellcolor{second}0.163 & 0.245 & 0.361 & 0.212 & 0.371 & 0.250 & \cellcolor{best}\textbf{0.064} & \cellcolor{second}0.165 & 0.289 & 0.365 & 0.288 & 0.387 & 0.281 \\
KB4 (200$^\circ$) & \cellcolor{best}\textbf{0.050} & \cellcolor{second}0.175 & 0.246 & 0.388 & 0.243 & 0.344 & 0.240 & \cellcolor{best}\textbf{0.061} & \cellcolor{second}0.203 & 0.320 & 0.404 & 0.320 & 0.368 & 0.261 \\
Double-Sph. (200$^\circ$) & \cellcolor{best}\textbf{0.055} & \cellcolor{second}0.165 & 0.259 & 0.368 & 0.259 & 0.358 & 0.249 & \cellcolor{best}\textbf{0.065} & \cellcolor{second}0.201 & 0.328 & 0.389 & 0.344 & 0.390 & 0.282 \\
ERP (360$^\circ$) & \cellcolor{best}\textbf{0.045} & 0.419 & 0.342 & 0.390 & 0.386 & 0.401 & \cellcolor{second}0.077 & \cellcolor{best}\textbf{0.064} & 0.408 & 0.430 & 0.415 & 0.421 & 0.395 & \cellcolor{second}0.066 \\
\midrule
 & \multicolumn{7}{c}{Structured3D} & \multicolumn{7}{c}{PanoCity} \\
\cmidrule(lr){2-8}\cmidrule(lr){9-15}
Camera (FoV) & Any3R & $\pi^3$ & MapA & DA3 & V-$\Omega$ & VGGT & PanoV & Any3R & $\pi^3$ & MapA & DA3 & V-$\Omega$ & VGGT & PanoV \\
\midrule
Pinhole (90$^\circ$) & 0.757 & 0.819 & 0.774 & \cellcolor{best}\textbf{0.665} & \cellcolor{second}0.729 & 0.781 & 0.812 & \cellcolor{best}\textbf{1.160} & \cellcolor{second}1.479 & 1.485 & 4.360 & 1.711 & 4.153 & 2.256 \\
Brown--C. (90$^\circ$) & 0.706 & \cellcolor{second}0.647 & 0.671 & 0.716 & 0.845 & \cellcolor{best}\textbf{0.645} & 0.771 & \cellcolor{best}\textbf{1.243} & 1.711 & 1.633 & 5.689 & \cellcolor{second}1.572 & 3.940 & 2.708 \\
Equidist. (180$^\circ$) & \cellcolor{best}\textbf{0.443} & \cellcolor{second}0.531 & 0.773 & 0.539 & 0.693 & 0.623 & 0.542 & \cellcolor{best}\textbf{0.866} & 2.093 & \cellcolor{second}2.066 & 6.656 & 2.162 & 6.296 & 2.498 \\
ERP-Cr. (180$^\circ$) & \cellcolor{best}\textbf{0.428} & 0.611 & 0.555 & 0.610 & 0.622 & \cellcolor{second}0.513 & 0.601 & \cellcolor{best}\textbf{0.811} & \cellcolor{second}2.426 & 2.861 & 2.717 & 2.455 & 3.078 & 3.633 \\
MEI (180$^\circ$) & \cellcolor{best}\textbf{0.436} & \cellcolor{second}0.523 & 0.527 & 0.549 & 0.572 & 0.637 & 0.533 & \cellcolor{best}\textbf{0.854} & \cellcolor{second}2.036 & 2.236 & 6.844 & 2.490 & 6.886 & 2.433 \\
KB4 (200$^\circ$) & \cellcolor{best}\textbf{0.294} & \cellcolor{second}0.508 & 0.519 & 0.563 & 0.653 & 0.635 & 0.568 & \cellcolor{best}\textbf{0.863} & \cellcolor{second}2.394 & 3.297 & 6.463 & 2.680 & 6.623 & 2.405 \\
Double-Sph. (200$^\circ$) & \cellcolor{best}\textbf{0.340} & 0.543 & \cellcolor{second}0.502 & 0.624 & 0.602 & 0.597 & 0.596 & \cellcolor{best}\textbf{0.901} & \cellcolor{second}2.271 & 2.789 & 6.477 & 2.544 & 6.430 & 2.473 \\
ERP (360$^\circ$) & \cellcolor{best}\textbf{0.128} & 0.445 & 0.440 & 0.553 & 0.496 & 0.442 & \cellcolor{second}0.144 & \cellcolor{best}\textbf{0.663} & 1.628 & 1.833 & 3.149 & 1.876 & 1.902 & \cellcolor{second}0.751 \\
\bottomrule
\end{tabular}\end{table*}

\label{sec::camsplitpool} \cref{tab::main} gives the Acc of \cref{tab::gfm} on each dataset separately, together with PanoVGGT and with Structured3D as a fourth dataset.
\cref{tab::mainmed} gives the median over scenes on which each verdict of \cref{tab::main} is decided.
\cref{tab::fov} averages \cref{tab::main} over the narrow and over the wide cameras.
On Matterport3D, Stanford2D3DS and PanoCity, Any3R is more accurate on the wide cameras than on the narrow ones, while no perspective-native baseline improves towards the wide end.
On Structured3D, whose three views share no content, every method is more accurate on the wide cameras, and Any3R gains the most.
Structured3D shows what remains when no view can help another.
Its views share no content, so no method can relate them and the comparison rests on how each method reads a single view; on the pinhole cell this favours the pinhole models, and DA3, which comes from monocular depth estimation, has the lowest mean error.
From $180^\circ$ on Any3R leads every Structured3D cell of \cref{tab::main}, so its advantage on wide lenses does not depend on relating views but lies already in how a single view is read.
\cref{tab::mainpose} gives the relative rotation error of each arm shown in \cref{tab::main}, so every Acc can be read against the poses it was fused with.

\cref{tab::mix} feeds three different cameras into one forward pass, view $k$ of each scene through the $k$-th camera listed.
With Brown--Conrady, double sphere and the cropped equirectangular frame, Any3R is best on all three datasets.
With pinhole, KB4 and the full equirectangular frame, PanoVGGT has the lowest median on all three, Any3R ties it on Matterport3D and Stanford2D3DS, and Any3R is behind on PanoCity.
\cref{tab::mixarms} lists both inputs of the baselines that have two.

\begin{table*}[!tb]\centering\footnotesize
\setlength{\tabcolsep}{1.4pt}
\caption{\textbf{Median accuracy.} The median over scenes of the per-scene Acc$\downarrow$ (m), the statistic the verdicts of \cref{tab::main} are decided on.}
\label{tab::mainmed}
\begin{tabular}{lcccccccccccccc}
\toprule
 & \multicolumn{7}{c}{Matterport3D} & \multicolumn{7}{c}{Stanford2D3DS} \\
\cmidrule(lr){2-8}\cmidrule(lr){9-15}
Camera (FoV) & Any3R & $\pi^3${\footnotesize\cite{wang2025pi3}} & MapA{\footnotesize\cite{keetha2025mapanything}} & DA3{\footnotesize\cite{depthanything3}} & V-$\Omega${\footnotesize\cite{wang2026vggtomega}} & VGGT{\footnotesize\cite{wang2025vggt}} & PanoV{\footnotesize\cite{panovggt2026}} & Any3R & $\pi^3$ & MapA & DA3 & V-$\Omega$ & VGGT & PanoV \\
\midrule
Pinhole (90$^\circ$) & \cellcolor{second}0.052 & \cellcolor{best}\textbf{0.041} & 0.069 & 0.261 & 0.101 & 0.252 & 0.266 & \cellcolor{second}0.053 & \cellcolor{best}\textbf{0.041} & 0.062 & 0.312 & 0.087 & 0.239 & 0.259 \\
Brown--C. (90$^\circ$) & \cellcolor{best}\textbf{0.063} & \cellcolor{second}0.072 & 0.151 & 0.259 & 0.126 & 0.274 & 0.253 & \cellcolor{second}0.069 & \cellcolor{best}\textbf{0.068} & 0.121 & 0.297 & 0.122 & 0.247 & 0.287 \\
Equidist. (180$^\circ$) & \cellcolor{best}\textbf{0.051} & \cellcolor{second}0.139 & 0.265 & 0.328 & 0.195 & 0.313 & 0.230 & \cellcolor{best}\textbf{0.050} & \cellcolor{second}0.141 & 0.296 & 0.357 & 0.257 & 0.383 & 0.243 \\
ERP-Cr. (180$^\circ$) & \cellcolor{best}\textbf{0.046} & 0.243 & \cellcolor{second}0.212 & 0.267 & 0.239 & 0.243 & 0.337 & \cellcolor{best}\textbf{0.041} & 0.268 & 0.268 & 0.290 & \cellcolor{second}0.245 & 0.251 & 0.344 \\
MEI (180$^\circ$) & \cellcolor{best}\textbf{0.048} & \cellcolor{second}0.139 & 0.249 & 0.302 & 0.201 & 0.332 & 0.236 & \cellcolor{best}\textbf{0.051} & \cellcolor{second}0.130 & 0.268 & 0.343 & 0.258 & 0.376 & 0.260 \\
KB4 (200$^\circ$) & \cellcolor{best}\textbf{0.049} & \cellcolor{second}0.146 & 0.239 & 0.337 & 0.231 & 0.339 & 0.231 & \cellcolor{best}\textbf{0.052} & \cellcolor{second}0.150 & 0.287 & 0.397 & 0.317 & 0.366 & 0.241 \\
Double-Sph. (200$^\circ$) & \cellcolor{best}\textbf{0.051} & \cellcolor{second}0.144 & 0.266 & 0.300 & 0.256 & 0.331 & 0.241 & \cellcolor{best}\textbf{0.055} & \cellcolor{second}0.146 & 0.323 & 0.373 & 0.334 & 0.386 & 0.255 \\
ERP (360$^\circ$) & \cellcolor{best}\textbf{0.045} & 0.374 & 0.347 & 0.388 & 0.351 & 0.348 & \cellcolor{second}0.065 & \cellcolor{best}\textbf{0.042} & 0.382 & 0.428 & 0.404 & 0.392 & 0.388 & \cellcolor{second}0.051 \\
\midrule
 & \multicolumn{7}{c}{Structured3D} & \multicolumn{7}{c}{PanoCity} \\
\cmidrule(lr){2-8}\cmidrule(lr){9-15}
Camera (FoV) & Any3R & $\pi^3$ & MapA & DA3 & V-$\Omega$ & VGGT & PanoV & Any3R & $\pi^3$ & MapA & DA3 & V-$\Omega$ & VGGT & PanoV \\
\midrule
Pinhole (90$^\circ$) & 0.481 & 0.476 & \cellcolor{second}0.410 & 0.434 & \cellcolor{best}\textbf{0.380} & 0.485 & 0.575 & \cellcolor{second}1.085 & \cellcolor{best}\textbf{1.053} & 1.161 & 3.935 & 1.344 & 3.724 & 1.890 \\
Brown--C. (90$^\circ$) & \cellcolor{second}0.373 & \cellcolor{best}\textbf{0.341} & 0.469 & 0.484 & 0.484 & 0.469 & 0.442 & \cellcolor{best}\textbf{1.158} & \cellcolor{second}1.319 & 1.477 & 5.368 & 1.321 & 3.782 & 2.299 \\
Equidist. (180$^\circ$) & \cellcolor{best}\textbf{0.344} & \cellcolor{second}0.355 & 0.427 & 0.434 & 0.484 & 0.497 & 0.398 & \cellcolor{best}\textbf{0.831} & \cellcolor{second}2.000 & 2.023 & 6.635 & 2.184 & 5.861 & 2.290 \\
ERP-Cr. (180$^\circ$) & \cellcolor{best}\textbf{0.326} & 0.546 & 0.480 & \cellcolor{second}0.418 & 0.511 & 0.443 & 0.477 & \cellcolor{best}\textbf{0.765} & \cellcolor{second}2.509 & 3.031 & 3.062 & 2.694 & 3.072 & 3.485 \\
MEI (180$^\circ$) & \cellcolor{best}\textbf{0.299} & \cellcolor{second}0.326 & 0.417 & 0.452 & 0.523 & 0.498 & 0.344 & \cellcolor{best}\textbf{0.815} & \cellcolor{second}1.966 & 2.128 & 6.556 & 2.277 & 5.992 & 2.274 \\
KB4 (200$^\circ$) & \cellcolor{best}\textbf{0.254} & \cellcolor{second}0.345 & 0.417 & 0.502 & 0.464 & 0.492 & 0.362 & \cellcolor{best}\textbf{0.862} & \cellcolor{second}2.226 & 2.459 & 6.480 & 2.686 & 6.388 & 2.333 \\
Double-Sph. (200$^\circ$) & \cellcolor{best}\textbf{0.200} & \cellcolor{second}0.398 & 0.414 & 0.554 & 0.442 & 0.451 & 0.444 & \cellcolor{best}\textbf{0.891} & \cellcolor{second}2.141 & 2.390 & 6.529 & 2.387 & 6.047 & 2.359 \\
ERP (360$^\circ$) & \cellcolor{best}\textbf{0.048} & 0.381 & 0.322 & 0.467 & 0.432 & 0.383 & \cellcolor{second}0.066 & \cellcolor{best}\textbf{0.587} & 0.853 & 0.905 & 2.861 & 1.073 & 0.978 & \cellcolor{second}0.737 \\
\bottomrule
\end{tabular}\end{table*}

\begin{table*}[!tb]\centering\footnotesize
\setlength{\tabcolsep}{1.6pt}
\caption{\textbf{Accuracy split by field of view.} Acc$\downarrow$ (m), mean of the cell means of the 2 narrow ($90^\circ$) and the 6 wide ($\ge180^\circ$) cameras.}
\label{tab::fov}
\begin{tabular}{lcccccccc}
\toprule
Method & \multicolumn{2}{c}{Matterport3D} & \multicolumn{2}{c}{Stanford2D3DS} & \multicolumn{2}{c}{Structured3D} & \multicolumn{2}{c}{PanoCity} \\
\cmidrule(lr){2-3}\cmidrule(lr){4-5}\cmidrule(lr){6-7}\cmidrule(lr){8-9}
 & $90^\circ$ & $\ge180^\circ$ & $90^\circ$ & $\ge180^\circ$ & $90^\circ$ & $\ge180^\circ$ & $90^\circ$ & $\ge180^\circ$ \\
\midrule
Any3R (ours) & \cellcolor{best}\textbf{0.0826} & \cellcolor{best}\textbf{0.0517} & \cellcolor{best}\textbf{0.0838} & \cellcolor{best}\textbf{0.0614} & 0.7317 & \cellcolor{best}\textbf{0.3449} & \cellcolor{best}\textbf{1.2019} & \cellcolor{best}\textbf{0.8265} \\
$\pi^3$~\cite{wang2025pi3} & \cellcolor{second}0.1132 & \cellcolor{second}0.2253 & \cellcolor{second}0.0896 & \cellcolor{second}0.2439 & 0.7331 & 0.5266 & 1.5954 & \cellcolor{second}2.1415 \\
MapAnything~\cite{keetha2025mapanything} & 0.1196 & 0.2634 & 0.1221 & 0.3269 & 0.7226 & 0.5527 & \cellcolor{second}1.5591 & 2.5135 \\
DA3~\cite{depthanything3} & 0.2922 & 0.3557 & 0.3139 & 0.3738 & \cellcolor{best}\textbf{0.6907} & 0.5728 & 5.0245 & 5.3842 \\
VGGT-$\Omega$~\cite{wang2026vggtomega} & 0.1501 & 0.2601 & 0.1607 & 0.3243 & 0.7866 & 0.6062 & 1.6416 & 2.3678 \\
VGGT~\cite{wang2025vggt} & 0.3042 & 0.3464 & 0.2472 & 0.3678 & \cellcolor{second}0.7129 & 0.5744 & 4.0461 & 5.2024 \\
PanoVGGT~\cite{panovggt2026} & 0.2709 & 0.2326 & 0.2838 & 0.2540 & 0.7917 & \cellcolor{second}0.4973 & 2.4818 & 2.3655 \\
\bottomrule
\end{tabular}\end{table*}

\begin{table*}[!tb]\centering\footnotesize
\setlength{\tabcolsep}{1.6pt}
\caption{\textbf{Mixed cameras in one forward pass.} Acc$\downarrow$ (m) when view $k$ of each scene comes from the $k$-th camera listed, each view at its own size on one zero-padded canvas.}
\label{tab::mix}
\begin{tabular}{llccccccc}
\toprule
Dataset & Cameras & Any3R & $\pi^3${\footnotesize\cite{wang2025pi3}} & MapA{\footnotesize\cite{keetha2025mapanything}} & DA3{\footnotesize\cite{depthanything3}} & V-$\Omega${\footnotesize\cite{wang2026vggtomega}} & VGGT{\footnotesize\cite{wang2025vggt}} & PanoV{\footnotesize\cite{panovggt2026}} \\
\midrule
Matterport3D & Pinhole + KB4 + ERP & \cellcolor{best}\textbf{0.108} & 0.343 & 0.444 & 0.357 & 0.377 & 0.357 & \cellcolor{second}0.114 \\
Matterport3D & Brown--C. + Double-Sph. + ERP-Cr. & \cellcolor{best}\textbf{0.057} & 0.278 & \cellcolor{second}0.254 & 0.325 & 0.268 & 0.355 & 0.290 \\
Stanford2D3DS & Pinhole + KB4 + ERP & \cellcolor{best}\textbf{0.116} & 0.359 & 0.374 & 0.399 & 0.405 & 0.433 & \cellcolor{second}0.133 \\
Stanford2D3DS & Brown--C. + Double-Sph. + ERP-Cr. & \cellcolor{best}\textbf{0.064} & \cellcolor{second}0.244 & 0.306 & 0.358 & 0.341 & 0.395 & 0.329 \\
PanoCity & Pinhole + KB4 + ERP & \cellcolor{second}1.282 & 1.821 & 1.438 & 3.242 & 1.791 & 2.538 & \cellcolor{best}\textbf{1.085} \\
PanoCity & Brown--C. + Double-Sph. + ERP-Cr. & \cellcolor{best}\textbf{1.215} & 2.568 & \cellcolor{second}2.329 & 4.944 & 2.543 & 3.842 & 3.006 \\
\bottomrule
\end{tabular}\end{table*}

\begin{table*}[!tb]\centering\footnotesize
\setlength{\tabcolsep}{1.4pt}
\caption{\textbf{Both inputs of the two-input baselines on mixed cameras.} Median per-scene Acc$\downarrow$ (m); \cref{tab::mix} uses the \textbf{bold} one.}
\label{tab::mixarms}
\begin{tabular}{llcccccccccc}
\toprule
 & & \multicolumn{2}{c}{$\pi^3$} & \multicolumn{2}{c}{VGGT} & \multicolumn{2}{c}{V-$\Omega$} & \multicolumn{2}{c}{DA3} & \multicolumn{2}{c}{MapA} \\
\cmidrule(lr){3-4}\cmidrule(lr){5-6}\cmidrule(lr){7-8}\cmidrule(lr){9-10}\cmidrule(lr){11-12}
Dataset & Cameras & nat. & off. & nat. & off. & nat. & off. & fixed & canvas & fixed & canvas \\
\midrule
Matterport3D & Pinhole + KB4 + ERP & 0.345 & \textbf{0.328} & \textbf{0.314} & 0.324 & 0.354 & \textbf{0.348} & \textbf{0.315} & 0.351 & 0.344 & \textbf{0.315} \\
Matterport3D & Brown--C. + Double-Sph. + ERP-Cr. & \textbf{0.249} & 0.256 & \textbf{0.313} & 0.362 & \textbf{0.244} & 0.251 & 0.335 & \textbf{0.251} & \textbf{0.233} & 0.260 \\
Stanford2D3DS & Pinhole + KB4 + ERP & 0.378 & \textbf{0.357} & 0.447 & \textbf{0.424} & \textbf{0.401} & 0.433 & \textbf{0.378} & 0.410 & 0.422 & \textbf{0.379} \\
Stanford2D3DS & Brown--C. + Double-Sph. + ERP-Cr. & \textbf{0.242} & 0.243 & \textbf{0.383} & 0.399 & \textbf{0.300} & 0.331 & 0.359 & \textbf{0.346} & 0.325 & \textbf{0.301} \\
PanoCity & Pinhole + KB4 + ERP & 2.513 & \textbf{1.238} & 4.019 & \textbf{1.844} & 1.277 & \textbf{1.109} & \textbf{2.831} & 3.128 & \textbf{1.049} & 1.801 \\
PanoCity & Brown--C. + Double-Sph. + ERP-Cr. & 3.004 & \textbf{2.830} & 3.890 & \textbf{3.329} & \textbf{2.616} & 2.716 & 5.053 & \textbf{5.014} & 2.852 & \textbf{2.304} \\
\bottomrule
\end{tabular}\end{table*}

\begin{table*}[!tb]\centering\footnotesize
\setlength{\tabcolsep}{1.4pt}
\caption{\textbf{Relative pose of the methods in \cref{tab::main}.} $\overline{\mathrm{R}}_\text{med}\downarrow$ (degrees), each scene's median rotation error over its view pairs averaged over the scenes, with view pairs of ground-truth translation below $0.01$ dropped as in the PanoVGGT code; this table is not ranked.}
\label{tab::mainpose}
\begin{tabular}{lcccccccccccccc}
\toprule
 & \multicolumn{7}{c}{Matterport3D} & \multicolumn{7}{c}{Stanford2D3DS} \\
\cmidrule(lr){2-8}\cmidrule(lr){9-15}
Camera (FoV) & Any3R & $\pi^3${\footnotesize\cite{wang2025pi3}} & MapA{\footnotesize\cite{keetha2025mapanything}} & DA3{\footnotesize\cite{depthanything3}} & V-$\Omega${\footnotesize\cite{wang2026vggtomega}} & VGGT{\footnotesize\cite{wang2025vggt}} & PanoV{\footnotesize\cite{panovggt2026}} & Any3R & $\pi^3$ & MapA & DA3 & V-$\Omega$ & VGGT & PanoV \\
\midrule
Pinhole (90$^\circ$) & 3.91 & 9.84 & 13.55 & 56.33 & 14.94 & 23.82 & 35.08 & 11.00 & 7.30 & 23.29 & 55.28 & 12.52 & 23.80 & 38.99 \\
Brown--C. (90$^\circ$) & 11.68 & 14.92 & 21.13 & 68.61 & 14.31 & 41.34 & 26.91 & 11.08 & 26.68 & 38.14 & 64.58 & 21.61 & 51.50 & 33.58 \\
Equidist. (180$^\circ$) & 8.61 & 12.20 & 33.63 & 55.00 & 18.92 & 47.20 & 13.11 & 4.82 & 18.93 & 43.03 & 67.73 & 34.08 & 49.77 & 18.56 \\
ERP-Cr. (180$^\circ$) & 8.40 & 24.13 & 36.10 & 22.10 & 26.86 & 30.50 & 16.66 & 5.23 & 33.80 & 52.67 & 21.25 & 35.16 & 41.16 & 22.83 \\
MEI (180$^\circ$) & 1.47 & 11.62 & 33.56 & 59.84 & 19.45 & 47.58 & 11.31 & 4.92 & 18.75 & 43.39 & 61.79 & 35.53 & 52.14 & 22.03 \\
KB4 (200$^\circ$) & 3.10 & 14.96 & 32.54 & 49.45 & 20.85 & 47.60 & 9.21 & 6.13 & 25.88 & 41.89 & 50.54 & 36.78 & 47.09 & 15.95 \\
Double-Sph. (200$^\circ$) & 9.99 & 12.77 & 37.66 & 55.73 & 22.79 & 46.71 & 8.44 & 6.11 & 17.77 & 44.65 & 62.40 & 36.46 & 47.91 & 10.83 \\
ERP (360$^\circ$) & 1.12 & 50.34 & 51.76 & 56.19 & 42.46 & 47.96 & 1.90 & 4.88 & 52.03 & 59.17 & 44.77 & 43.22 & 54.41 & 4.90 \\
\midrule
 & \multicolumn{7}{c}{Structured3D} & \multicolumn{7}{c}{PanoCity} \\
\cmidrule(lr){2-8}\cmidrule(lr){9-15}
Camera (FoV) & Any3R & $\pi^3$ & MapA & DA3 & V-$\Omega$ & VGGT & PanoV & Any3R & $\pi^3$ & MapA & DA3 & V-$\Omega$ & VGGT & PanoV \\
\midrule
Pinhole (90$^\circ$) & 70.02 & 64.23 & 49.02 & 92.82 & 45.28 & 45.78 & 60.91 & 6.89 & 39.65 & 13.84 & 51.53 & 33.01 & 39.13 & 35.32 \\
Brown--C. (90$^\circ$) & 71.91 & 59.20 & 52.76 & 98.18 & 45.03 & 68.74 & 43.66 & 8.55 & 53.79 & 49.55 & 61.55 & 45.35 & 73.91 & 39.09 \\
Equidist. (180$^\circ$) & 29.54 & 39.78 & 41.73 & 95.54 & 38.66 & 101.77 & 16.84 & 2.69 & 34.73 & 27.59 & 38.12 & 33.50 & 60.06 & 12.06 \\
ERP-Cr. (180$^\circ$) & 36.67 & 76.71 & 54.07 & 47.66 & 62.22 & 45.78 & 49.17 & 2.61 & 50.30 & 39.31 & 35.63 & 44.94 & 43.18 & 8.17 \\
MEI (180$^\circ$) & 28.13 & 43.08 & 46.78 & 86.50 & 37.23 & 100.93 & 18.17 & 2.75 & 32.62 & 26.79 & 40.40 & 32.61 & 48.48 & 11.83 \\
KB4 (200$^\circ$) & 12.72 & 37.52 & 49.66 & 96.91 & 40.27 & 115.44 & 30.47 & 3.49 & 37.06 & 41.86 & 38.78 & 39.31 & 55.71 & 7.45 \\
Double-Sph. (200$^\circ$) & 13.36 & 38.38 & 45.52 & 91.02 & 41.41 & 92.37 & 32.09 & 3.35 & 31.10 & 48.14 & 45.99 & 39.76 & 44.66 & 6.20 \\
ERP (360$^\circ$) & 1.26 & 60.96 & 77.63 & 63.73 & 74.99 & 45.36 & 1.87 & 0.51 & 83.70 & 85.15 & 55.74 & 79.83 & 79.65 & 1.37 \\
\bottomrule
\end{tabular}\end{table*}

\begin{figure*}[!tb]\centering
\includegraphics[width=\textwidth]{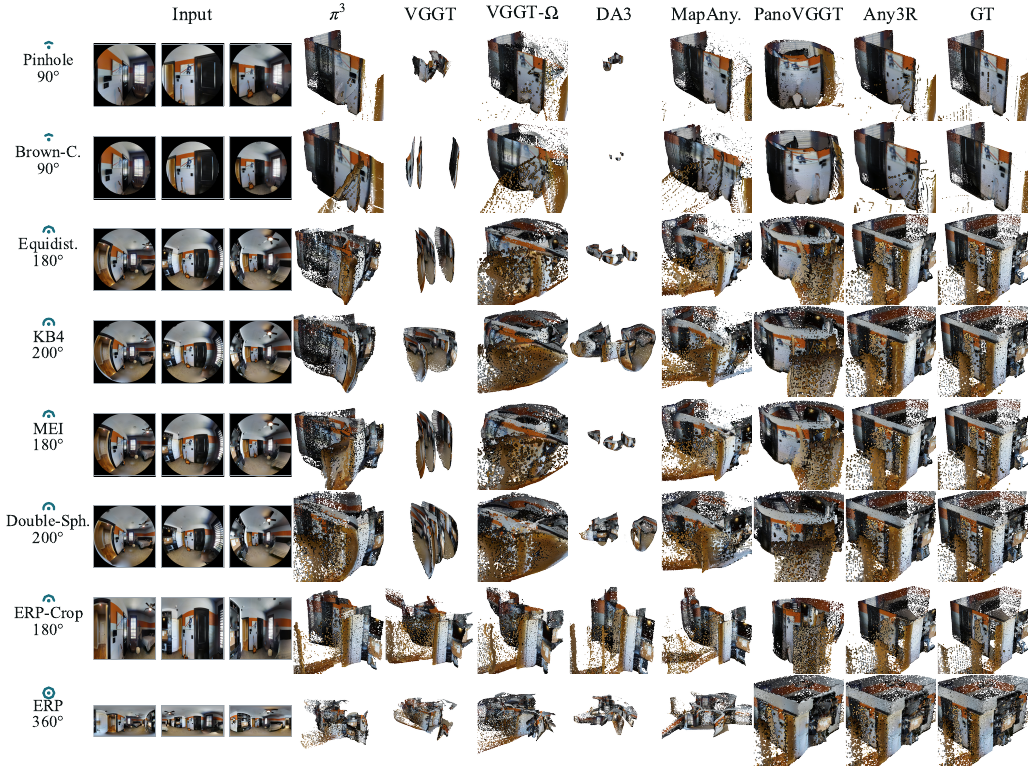}
\caption{One Matterport3D scene, all eight central camera models, all seven methods. Identical protocol to \cref{fig::cams}. The scene is fixed, so the vertical axis is field of view alone.}
\label{fig::same}
\end{figure*}

\label{sec::app_samescene}

Fig.~\ref{fig::cams} in the main text draws a different scene in each row, which shows the range of content the benchmark covers but means that rows cannot be compared with one another: a difference between two rows mixes a change of camera with a change of scene.
Fig.~\ref{fig::same} holds the scene fixed and varies only the camera, so the vertical axis isolates field of view.
The two figures are produced by the same code path and the same alignment; only the scene assignment differs.

Down each column, the perspective-native methods stay plausible while the lens is narrow and flatten or fold the room once it is not, and VGGT and DA3 lose the relative pose on most rows, whereas Any3R keeps the room on every lens and the panoramic baseline keeps it only on the full equirectangular frame.
This is the qualitative form of the split reported in \cref{tab::fov}, which is computed over all $32$ cells of \cref{tab::main} rather than the single scene shown here.

\begin{table*}[!tb]\renewenvironment{table}[1][]{}{}
\begin{minipage}[t]{0.485\textwidth}
\begin{table}[tb]\centering\footnotesize
\setlength{\tabcolsep}{2.6pt}
\caption{\textbf{Comparison with the wide-field-of-view specialist Wid3R~\cite{jung2026wid3r}.} Acc$\downarrow$ (m) on the four cameras its camera classes represent, with its ray field supplied (ray fld.) or predicted (pred.).}
\label{tab::wid3r}
\begin{tabular}{llcccc}
\toprule
Camera & Method & M3D & S2D3DS & S3D & PanoCity \\
\midrule
\multirow{3}{*}{Pinhole $90^\circ$} & Any3R & \cellcolor{best}\textbf{0.069} & \cellcolor{best}\textbf{0.083} & \cellcolor{best}\textbf{0.757} & \cellcolor{best}\textbf{1.160} \\
 & Wid3R ray fld. & \cellcolor{second}0.150 & \cellcolor{second}0.139 & \cellcolor{second}0.770 & 2.278 \\
 & Wid3R pred. & 0.199 & 0.183 & 0.818 & \cellcolor{second}2.182 \\
\midrule
\multirow{3}{*}{Equidistant $180^\circ$} & Any3R & \cellcolor{best}\textbf{0.056} & \cellcolor{best}\textbf{0.061} & \cellcolor{best}\textbf{0.443} & \cellcolor{best}\textbf{0.866} \\
 & Wid3R ray fld. & \cellcolor{second}0.111 & \cellcolor{second}0.172 & \cellcolor{second}0.568 & 2.121 \\
 & Wid3R pred. & 0.124 & 0.184 & 0.608 & \cellcolor{second}2.100 \\
\midrule
\multirow{3}{*}{MEI $180^\circ$} & Any3R & \cellcolor{best}\textbf{0.054} & \cellcolor{best}\textbf{0.064} & \cellcolor{best}\textbf{0.436} & \cellcolor{best}\textbf{0.854} \\
 & Wid3R ray fld. & \cellcolor{second}0.126 & \cellcolor{second}0.166 & \cellcolor{second}0.597 & \cellcolor{second}2.252 \\
 & Wid3R pred. & 0.138 & 0.188 & 0.607 & 2.274 \\
\midrule
\multirow{3}{*}{ERP $360^\circ$} & Any3R & \cellcolor{best}\textbf{0.045} & \cellcolor{best}\textbf{0.064} & \cellcolor{best}\textbf{0.128} & \cellcolor{best}\textbf{0.663} \\
 & Wid3R ray fld. & \cellcolor{second}0.069 & \cellcolor{second}0.133 & \cellcolor{second}0.444 & \cellcolor{second}2.330 \\
 & Wid3R pred. & 0.070 & 0.134 & 0.448 & 3.878 \\
\bottomrule
\end{tabular}\end{table}
\vspace{6pt}
\providecommand{\panoErpArgusWins}{}\renewcommand{\panoErpArgusWins}{9/9}
\providecommand{\panoErpArgusRatio}{}\renewcommand{\panoErpArgusRatio}{0.48}
\providecommand{\panoErpPanoWins}{}\renewcommand{\panoErpPanoWins}{8/9}
\providecommand{\panoErpPanoRatio}{}\renewcommand{\panoErpPanoRatio}{0.88}
\providecommand{\panoErpWidWins}{}\renewcommand{\panoErpWidWins}{9/9}
\providecommand{\panoErpWidRatio}{}\renewcommand{\panoErpWidRatio}{0.48}

\begin{table}[tb]\centering\footnotesize
\setlength{\tabcolsep}{3pt}
\caption{\textbf{Comparison with the panoramic-only methods.} Acc$\downarrow$ (m) on the equirectangular frames, Argus at its own $560{\times}280$ input without the top and bottom $15\%$; PaGeR predicts no pose and is not shown, and -- marks Argus on S3D, whose views share one orientation, so its rotation convention cannot be checked.}
\label{tab::pano13}
\begin{tabular}{lcccc}
\toprule
Dataset & Any3R & Argus~\cite{realsee2026argus} & PanoVGGT~\cite{panovggt2026} & Wid3R~\cite{jung2026wid3r} \\
\midrule
M3D & \cellcolor{best}\textbf{0.045} & 0.094 & 0.077 & \cellcolor{second}0.069 \\
S2D3DS & \cellcolor{best}\textbf{0.064} & 0.094 & \cellcolor{second}0.066 & 0.133 \\
S3D & \cellcolor{best}\textbf{0.128} & -- & \cellcolor{second}0.144 & 0.444 \\
PanoCity & \cellcolor{best}\textbf{0.663} & 1.517 & \cellcolor{second}0.751 & 2.330 \\
\bottomrule
\end{tabular}\end{table}

\end{minipage}\hfill
\begin{minipage}[t]{0.485\textwidth}
\providecommand{\ppOmniCells}{}\renewcommand{\ppOmniCells}{39}
\providecommand{\ppOmniOursBest}{}\renewcommand{\ppOmniOursBest}{37}
\providecommand{\ppOmniOursTied}{}\renewcommand{\ppOmniOursTied}{2}
\providecommand{\ppOmniMedianRatio}{}\renewcommand{\ppOmniMedianRatio}{0.18}
\providecommand{\ppOmniOfficialChosen}{}\renewcommand{\ppOmniOfficialChosen}{16}
\providecommand{\ppOmniSthreeCells}{}\renewcommand{\ppOmniSthreeCells}{13}
\providecommand{\ppOmniSthreeOursBest}{}\renewcommand{\ppOmniSthreeOursBest}{6}
\providecommand{\ppOmniSthreeOursTied}{}\renewcommand{\ppOmniSthreeOursTied}{7}

\begin{table}[tb]\centering\footnotesize
\setlength{\tabcolsep}{1.2pt}
\caption{\textbf{Comparison with OmniVGGT~\cite{li2026omnivggt}, a backbone conditioned on intrinsics.} Acc$\downarrow$ (m) with images alone as input; Omni.\ is OmniVGGT.}
\label{tab::omni}
\begin{tabular*}{\columnwidth}{@{\extracolsep{\fill}}lcccccccc}
\toprule
 & \multicolumn{2}{c}{M3D} & \multicolumn{2}{c}{S2D3DS} & \multicolumn{2}{c}{S3D} & \multicolumn{2}{c}{PanoCity} \\
\cmidrule(lr){2-3}\cmidrule(lr){4-5}\cmidrule(lr){6-7}\cmidrule(lr){8-9}
Camera & {\scriptsize Any3R} & {\scriptsize Omni.} & {\scriptsize Any3R} & {\scriptsize Omni.} & {\scriptsize Any3R} & {\scriptsize Omni.} & {\scriptsize Any3R} & {\scriptsize Omni.} \\
\midrule
Pinhole $90^\circ$ & \cellcolor{best}\textbf{0.069} & 0.263 & \cellcolor{best}\textbf{0.083} & 0.237 & \cellcolor{best}\textbf{0.757} & 0.764 & \cellcolor{best}\textbf{1.160} & 2.875 \\
Brown--C. $90^\circ$ & \cellcolor{best}\textbf{0.096} & 0.225 & \cellcolor{best}\textbf{0.085} & 0.178 & 0.706 & \cellcolor{best}\textbf{0.657} & \cellcolor{best}\textbf{1.243} & 2.676 \\
FOV/ATAN $150^\circ$ & \cellcolor{best}\textbf{0.051} & 0.311 & \cellcolor{best}\textbf{0.053} & 0.288 & \cellcolor{best}\textbf{0.522} & 0.723 & \cellcolor{best}\textbf{0.929} & 2.898 \\
Equidist. $180^\circ$ & \cellcolor{best}\textbf{0.056} & 0.308 & \cellcolor{best}\textbf{0.061} & 0.343 & \cellcolor{best}\textbf{0.443} & 0.530 & \cellcolor{best}\textbf{0.866} & 3.313 \\
Stereogr. $180^\circ$ & \cellcolor{best}\textbf{0.060} & 0.329 & \cellcolor{best}\textbf{0.069} & 0.340 & \cellcolor{best}\textbf{0.451} & 0.519 & \cellcolor{best}\textbf{0.917} & 3.181 \\
ERP-Cr. $180^\circ$ & \cellcolor{best}\textbf{0.051} & 0.302 & \cellcolor{best}\textbf{0.053} & 0.333 & \cellcolor{best}\textbf{0.428} & 0.594 & \cellcolor{best}\textbf{0.811} & 2.804 \\
MEI $180^\circ$ & \cellcolor{best}\textbf{0.054} & 0.311 & \cellcolor{best}\textbf{0.064} & 0.347 & \cellcolor{best}\textbf{0.436} & 0.516 & \cellcolor{best}\textbf{0.854} & 3.703 \\
KB4 $200^\circ$ & \cellcolor{best}\textbf{0.050} & 0.338 & \cellcolor{best}\textbf{0.061} & 0.371 & \cellcolor{best}\textbf{0.294} & 0.447 & \cellcolor{best}\textbf{0.863} & 3.941 \\
Equisolid $200^\circ$ & \cellcolor{best}\textbf{0.056} & 0.329 & \cellcolor{best}\textbf{0.055} & 0.352 & \cellcolor{best}\textbf{0.310} & 0.506 & \cellcolor{best}\textbf{0.893} & 3.425 \\
Double-Sph. $200^\circ$ & \cellcolor{best}\textbf{0.055} & 0.339 & \cellcolor{best}\textbf{0.065} & 0.362 & \cellcolor{best}\textbf{0.340} & 0.531 & \cellcolor{best}\textbf{0.901} & 3.670 \\
EUCM $200^\circ$ & \cellcolor{best}\textbf{0.054} & 0.335 & \cellcolor{best}\textbf{0.067} & 0.357 & \cellcolor{best}\textbf{0.345} & 0.495 & \cellcolor{best}\textbf{0.896} & 3.855 \\
ERP $360^\circ$ & \cellcolor{best}\textbf{0.045} & 0.409 & \cellcolor{best}\textbf{0.064} & 0.403 & \cellcolor{best}\textbf{0.128} & 0.449 & \cellcolor{best}\textbf{0.663} & 1.479 \\
Cylind. $360^\circ$ & \cellcolor{best}\textbf{0.082} & 0.373 & \cellcolor{best}\textbf{0.098} & 0.411 & \cellcolor{best}\textbf{0.275} & 0.430 & \cellcolor{best}\textbf{1.298} & 3.021 \\
\bottomrule
\end{tabular*}\end{table}

\end{minipage}
\par\vspace{8pt}{\renewenvironment{table*}[1][]{}{}\begin{table*}[!tb]\centering\footnotesize
\caption{\textbf{MapAnything given the true ray field.} Acc$\downarrow$ (m) of Any3R, MapAnything as in \cref{tab::main} (MapA: images alone, or a pinhole intrinsic matrix on the pinhole camera, the better per camera) and MapAnything with the true ray field fed to its ray encoder (MapA-ray), since its released preprocessing fits a pinhole intrinsic matrix to any ray map.}
\label{tab::maparay}
\setlength{\tabcolsep}{3pt}
\begin{tabular}{lcccccccccccc}
\toprule
 & \multicolumn{3}{c}{M3D} & \multicolumn{3}{c}{S2D3DS} & \multicolumn{3}{c}{S3D} & \multicolumn{3}{c}{PanoCity} \\
\cmidrule(lr){2-4}\cmidrule(lr){5-7}\cmidrule(lr){8-10}\cmidrule(lr){11-13}
Camera & {\scriptsize Any3R} & {\scriptsize MapA} & {\scriptsize MapA-ray} & {\scriptsize Any3R} & {\scriptsize MapA} & {\scriptsize MapA-ray} & {\scriptsize Any3R} & {\scriptsize MapA} & {\scriptsize MapA-ray} & {\scriptsize Any3R} & {\scriptsize MapA} & {\scriptsize MapA-ray} \\
\midrule
Pinhole $90^\circ$ & \cellcolor{best}\textbf{0.069} & \cellcolor{second}0.089 & \cellcolor{second}0.089 & \cellcolor{best}\textbf{0.083} & \cellcolor{second}0.095 & \cellcolor{second}0.095 & \cellcolor{second}0.757 & 0.774 & \cellcolor{best}\textbf{0.675} & \cellcolor{best}\textbf{1.160} & 1.485 & \cellcolor{second}1.484 \\
Brown--C. $90^\circ$ & \cellcolor{best}\textbf{0.096} & 0.150 & \cellcolor{second}0.142 & \cellcolor{best}\textbf{0.085} & \cellcolor{second}0.149 & \cellcolor{second}0.149 & \cellcolor{second}0.706 & \cellcolor{best}\textbf{0.671} & \cellcolor{second}0.706 & \cellcolor{best}\textbf{1.243} & 1.633 & \cellcolor{second}1.558 \\
FOV/ATAN $150^\circ$ & \cellcolor{best}\textbf{0.051} & \cellcolor{second}0.229 & 0.262 & \cellcolor{best}\textbf{0.053} & \cellcolor{second}0.243 & 0.270 & \cellcolor{best}\textbf{0.522} & \cellcolor{second}0.533 & 0.548 & \cellcolor{best}\textbf{0.929} & \cellcolor{second}1.930 & 2.201 \\
Equidist. $180^\circ$ & \cellcolor{best}\textbf{0.056} & \cellcolor{second}0.246 & 0.271 & \cellcolor{best}\textbf{0.061} & 0.299 & \cellcolor{second}0.282 & \cellcolor{best}\textbf{0.443} & 0.773 & \cellcolor{second}0.496 & \cellcolor{best}\textbf{0.866} & \cellcolor{second}2.066 & 2.368 \\
Stereogr. $180^\circ$ & \cellcolor{best}\textbf{0.060} & \cellcolor{second}0.269 & 0.281 & \cellcolor{best}\textbf{0.069} & 0.295 & \cellcolor{second}0.258 & \cellcolor{best}\textbf{0.451} & 0.549 & \cellcolor{second}0.509 & \cellcolor{best}\textbf{0.917} & 2.383 & \cellcolor{second}2.346 \\
ERP-Cr. $180^\circ$ & \cellcolor{best}\textbf{0.051} & 0.242 & \cellcolor{second}0.241 & \cellcolor{best}\textbf{0.053} & 0.296 & \cellcolor{second}0.268 & \cellcolor{best}\textbf{0.428} & 0.555 & \cellcolor{second}0.522 & \cellcolor{best}\textbf{0.811} & 2.861 & \cellcolor{second}2.686 \\
MEI $180^\circ$ & \cellcolor{best}\textbf{0.054} & 0.245 & \cellcolor{second}0.226 & \cellcolor{best}\textbf{0.064} & 0.289 & \cellcolor{second}0.251 & \cellcolor{second}0.436 & 0.527 & \cellcolor{best}\textbf{0.425} & \cellcolor{best}\textbf{0.854} & 2.236 & \cellcolor{second}2.029 \\
KB4 $200^\circ$ & \cellcolor{best}\textbf{0.050} & \cellcolor{second}0.246 & 0.256 & \cellcolor{best}\textbf{0.061} & 0.320 & \cellcolor{second}0.292 & \cellcolor{best}\textbf{0.294} & 0.519 & \cellcolor{second}0.500 & \cellcolor{best}\textbf{0.863} & 3.297 & \cellcolor{second}2.523 \\
Equisolid $200^\circ$ & \cellcolor{best}\textbf{0.056} & \cellcolor{second}0.226 & 0.255 & \cellcolor{best}\textbf{0.055} & 0.291 & \cellcolor{second}0.281 & \cellcolor{best}\textbf{0.310} & 0.624 & \cellcolor{second}0.495 & \cellcolor{best}\textbf{0.893} & \cellcolor{second}2.268 & 2.525 \\
Double-Sph. $200^\circ$ & \cellcolor{best}\textbf{0.055} & 0.259 & \cellcolor{second}0.237 & \cellcolor{best}\textbf{0.065} & 0.328 & \cellcolor{second}0.278 & \cellcolor{best}\textbf{0.340} & \cellcolor{second}0.502 & 0.572 & \cellcolor{best}\textbf{0.901} & 2.789 & \cellcolor{second}2.579 \\
EUCM $200^\circ$ & \cellcolor{best}\textbf{0.054} & 0.261 & \cellcolor{second}0.259 & \cellcolor{best}\textbf{0.067} & 0.331 & \cellcolor{second}0.279 & \cellcolor{best}\textbf{0.345} & 0.544 & \cellcolor{second}0.507 & \cellcolor{best}\textbf{0.896} & 2.690 & \cellcolor{second}2.640 \\
ERP $360^\circ$ & \cellcolor{best}\textbf{0.045} & \cellcolor{second}0.342 & 0.350 & \cellcolor{best}\textbf{0.064} & \cellcolor{second}0.430 & 0.460 & \cellcolor{best}\textbf{0.128} & 0.440 & \cellcolor{second}0.404 & \cellcolor{best}\textbf{0.663} & \cellcolor{second}1.833 & 1.894 \\
Cylind. $360^\circ$ & \cellcolor{best}\textbf{0.082} & 0.312 & \cellcolor{second}0.294 & \cellcolor{best}\textbf{0.098} & \cellcolor{second}0.367 & 0.410 & \cellcolor{best}\textbf{0.275} & \cellcolor{second}0.421 & 0.446 & \cellcolor{best}\textbf{1.298} & \cellcolor{second}2.680 & 3.001 \\
\bottomrule
\end{tabular}
\end{table*}
}
\end{table*}
\section{Camera-Aware Methods per Camera}
\label{sec::wid3rcells}
\cref{tab::wid3r}, \cref{tab::pano13} and \cref{tab::omni} give the methods of \cref{tab::camaware} per camera and dataset, the panoramic methods on the equirectangular frame and OmniVGGT on all thirteen cameras with images alone, and \cref{tab::maparay} gives MapAnything with the true ray field beside MapAnything as run in \cref{tab::main}.
On Structured3D, whose views share no content, MapA-ray is more accurate than Any3R on the pinhole and MEI cameras.

\cref{tab::wid3r} keeps four of the thirteen cameras, pinhole, equidistant, MEI and equirectangular, whose mapping into Wid3R's camera classes is exact, and omits the other nine.
Wid3R's Fisheye624 class and its EUCM class both pin $z>0$, so neither represents a frame reaching past $90^\circ$ from the axis, which rules out the four $200^\circ$ lenses, and the radial polynomial of its Brown--Conrady class is not invertible outside the inscribed circle.
For the field-of-view, stereographic, cropped equirectangular and cylindrical models we have no exact mapping into its classes.
We report the cells it can express and name the ones it cannot rather than scoring it on a frame its parameterisation truncates or approximates.
Every kept cell's parameter mapping is checked against our own ray field before use, per pixel and not per lens label, with both the median and the $99$th percentile of the angular difference required to pass.

Wid3R trains at $518{\times}336$ while our frames are native-resolution, so we run it at both sizes, and \cref{tab::wid3r} shows the one with the lower median Acc in each cell, the training size in $19$ of the $32$ pairs of cell and ray-field mode.

\section{Per-Cell Accuracy on All Thirteen Cameras} 
\begin{figure*}[!tbp]\centering
\includegraphics[width=0.95\textwidth]{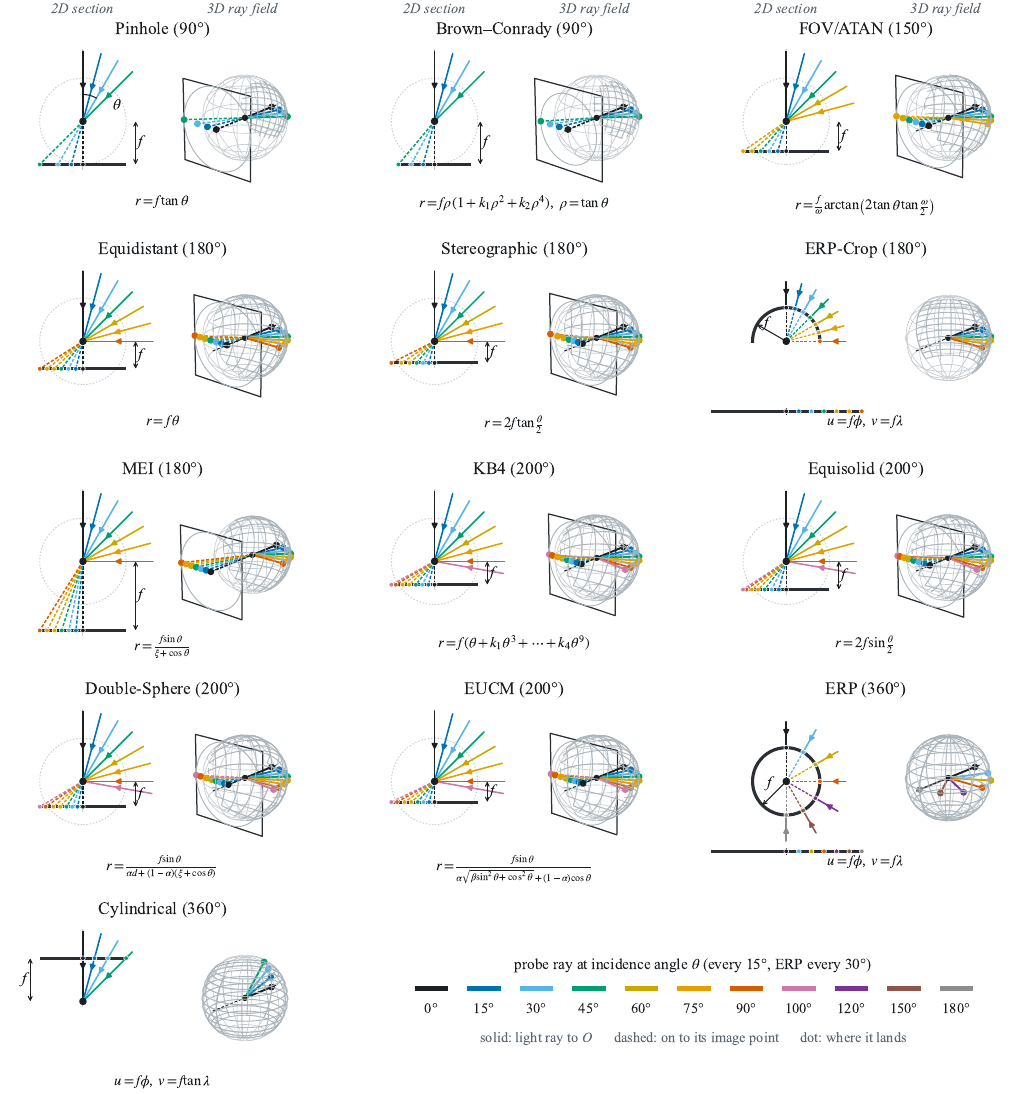}
\caption{\textbf{Projection of the thirteen camera models of \cref{tab::p13full}.} Left of each pair, a section through the camera centre $O$ shows probe rays spaced $15^\circ$ apart in incidence angle $\theta$, and $30^\circ$ apart for ERP, one colour per angle. Each travels from the scene to $O$ (solid) and on to its image point (dashed) on the image plane, a focal length $f$ behind $O$, so the spacing of the dots is the model's $r(\theta)$. ERP and ERP-Crop are seen from above, their image being the sphere of radius $f$ unrolled, and Cylindrical from the side, its image the cylinder wall a distance $f$ in front. Right of each pair, the same probes in 3D, from their image points through $O$ to the unit sphere, with the image's rows and columns drawn on the sphere in grey. Here $\phi$ and $\lambda$ are azimuth and elevation, $u$ and $v$ image coordinates, $\rho=\tan\theta$ and $d=\sqrt{\sin^2\theta+(\xi+\cos\theta)^2}$. Every camera is built as in the benchmark, with the declared field of view reaching half the image height, $k_1=-0.18$ and $k_2=0.02$ for Brown--Conrady together with the tangential terms $p_1=8\times10^{-4}$ and $p_2=-6\times10^{-4}$, $\omega=0.9$ for FOV/ATAN, $\xi=1.6$ for MEI, $(k_1,k_2,k_3,k_4)=(0.04,-0.01,0.002,-0.0003)$ for KB4, $\xi=0.25$ and $\alpha=0.55$ for Double-Sphere, $\alpha=0.55$ and $\beta=1$ for EUCM, and a vertical field of view of $90^\circ$ for ERP-Crop and Cylindrical.}
\label{fig::cammodels}
\end{figure*}

The five cameras marked $\dagger$ were added because the original eight leave two axes confounded.
Equisolid and stereographic pair with Kannala--Brandt and equidistant at the same field of view, which isolates the radial mapping from the angular extent and gives \cref{tab::p13ctrl} its pairs.
Cylindrical covers $360^\circ$ horizontally like equirectangular but maps elevation through a tangent, which separates panoramic extent from the equirectangular grid itself.
\cref{fig::cammodels} draws how each of the thirteen models projects, as a section through the camera centre and as the ray field it gives on the sphere.

\providecommand{\pairedWidCams}{}\renewcommand{\pairedWidCams}{4}
\providecommand{\ppAllCells}{}\renewcommand{\ppAllCells}{39}
\providecommand{\ppAllOursBest}{}\renewcommand{\ppAllOursBest}{32}
\providecommand{\ppAllOursTied}{}\renewcommand{\ppAllOursTied}{6}
\providecommand{\ppAllOursBehind}{}\renewcommand{\ppAllOursBehind}{1}
\providecommand{\ppAllSthreeCells}{}\renewcommand{\ppAllSthreeCells}{13}
\providecommand{\ppAllSthreeOursBest}{}\renewcommand{\ppAllSthreeOursBest}{9}
\providecommand{\ppAllSthreeOursTied}{}\renewcommand{\ppAllSthreeOursTied}{3}
\providecommand{\ppAllSthreeOursBehind}{}\renewcommand{\ppAllSthreeOursBehind}{1}

\begin{table*}[!tb]\centering\footnotesize
\setlength{\tabcolsep}{1.4pt}
\caption{\textbf{Accuracy by field of view on all thirteen cameras.} Acc$\downarrow$ (m) averaged over the cameras of each field of view.}
\label{tab::p13fov}
\begin{tabular}{lcccccccccccccc}
\toprule
 & \multicolumn{7}{c}{Matterport3D} & \multicolumn{7}{c}{Stanford2D3DS} \\
\cmidrule(lr){2-8}\cmidrule(lr){9-15}
FoV & Any3R & DA3{\footnotesize\cite{depthanything3}} & VGGT{\footnotesize\cite{wang2025vggt}} & $\pi^3${\footnotesize\cite{wang2025pi3}} & V-$\Omega${\footnotesize\cite{wang2026vggtomega}} & MapA{\footnotesize\cite{keetha2025mapanything}} & PanoV{\footnotesize\cite{panovggt2026}} & Any3R & DA3 & VGGT & $\pi^3$ & V-$\Omega$ & MapA & PanoV \\
\midrule
90$^\circ$ & \cellcolor{best}\textbf{0.083} & 0.292 & 0.304 & \cellcolor{second}0.113 & 0.150 & 0.120 & 0.271 & \cellcolor{best}\textbf{0.084} & 0.314 & 0.247 & \cellcolor{second}0.090 & 0.161 & 0.122 & 0.284 \\
150$^\circ$ & \cellcolor{best}\textbf{0.051} & 0.330 & 0.331 & \cellcolor{second}0.139 & 0.176 & 0.229 & 0.259 & \cellcolor{best}\textbf{0.053} & 0.309 & 0.344 & \cellcolor{second}0.133 & 0.261 & 0.243 & 0.281 \\
180$^\circ$ & \cellcolor{best}\textbf{0.055} & 0.334 & 0.329 & \cellcolor{second}0.184 & 0.229 & 0.250 & 0.273 & \cellcolor{best}\textbf{0.062} & 0.355 & 0.363 & \cellcolor{second}0.207 & 0.297 & 0.295 & 0.302 \\
200$^\circ$ & \cellcolor{best}\textbf{0.054} & 0.376 & 0.361 & \cellcolor{second}0.177 & 0.261 & 0.248 & 0.242 & \cellcolor{best}\textbf{0.062} & 0.381 & 0.378 & \cellcolor{second}0.203 & 0.328 & 0.317 & 0.265 \\
360$^\circ$ & \cellcolor{best}\textbf{0.063} & 0.366 & 0.383 & 0.380 & 0.364 & 0.327 & \cellcolor{second}0.246 & \cellcolor{best}\textbf{0.081} & 0.390 & 0.390 & 0.393 & 0.408 & 0.399 & \cellcolor{second}0.275 \\
\midrule
 & \multicolumn{7}{c}{Structured3D} & \multicolumn{7}{c}{PanoCity} \\
\cmidrule(lr){2-8}\cmidrule(lr){9-15}
FoV & Any3R & DA3 & VGGT & $\pi^3$ & V-$\Omega$ & MapA & PanoV & Any3R & DA3 & VGGT & $\pi^3$ & V-$\Omega$ & MapA & PanoV \\
\midrule
90$^\circ$ & 0.732 & \cellcolor{best}\textbf{0.691} & \cellcolor{second}0.713 & 0.733 & 0.787 & 0.723 & 0.792 & \cellcolor{best}\textbf{1.202} & 5.024 & 4.046 & 1.595 & 1.642 & \cellcolor{second}1.559 & 2.482 \\
150$^\circ$ & \cellcolor{best}\textbf{0.522} & 0.614 & 0.623 & 0.683 & 0.678 & \cellcolor{second}0.533 & 0.730 & \cellcolor{best}\textbf{0.929} & 5.700 & 6.161 & \cellcolor{second}1.856 & 1.944 & 1.930 & 2.368 \\
180$^\circ$ & \cellcolor{best}\textbf{0.440} & 0.578 & 0.602 & \cellcolor{second}0.562 & 0.631 & 0.601 & 0.573 & \cellcolor{best}\textbf{0.862} & 5.654 & 5.486 & \cellcolor{second}2.136 & 2.433 & 2.387 & 2.766 \\
200$^\circ$ & \cellcolor{best}\textbf{0.322} & 0.573 & 0.596 & \cellcolor{second}0.516 & 0.621 & 0.547 & 0.566 & \cellcolor{best}\textbf{0.888} & 6.366 & 6.607 & \cellcolor{second}2.368 & 2.725 & 2.761 & 2.425 \\
360$^\circ$ & \cellcolor{best}\textbf{0.201} & 0.510 & 0.456 & 0.457 & 0.486 & 0.431 & \cellcolor{second}0.340 & \cellcolor{best}\textbf{0.980} & 3.654 & 2.780 & \cellcolor{second}2.207 & 2.490 & 2.256 & 2.484 \\
\bottomrule
\end{tabular}\end{table*}

\cref{tab::p13fov} averages all thirteen cameras by field of view.
On every dataset our error is lower at every wider lens than at $90^\circ$, and the best pinhole model fares worse against us at each of them than at $90^\circ$.

Reading \cref{tab::p13full} by column shows the effect of the lens on one method, and reading it by row shows the ordering of the methods under one lens.
\cref{tab::p13arms} lists both inputs of the baselines that have two, on all thirteen cameras.
On the three datasets of the paired pool Any3R is best in $32$ of the $39$ cells and tied in six, five of them at $90^\circ$ and the sixth the cylindrical PanoCity cell, and it is behind $\pi^3$ on the Stanford2D3DS pinhole cell.
On the $13$ Structured3D cells it is best in nine and tied in three, the Brown--Conrady, FOV/ATAN and Equisolid cells, and it is behind VGGT-$\Omega$ and MapAnything on the pinhole cell.

The cylindrical cell is the one place the format matters rather than the field of view.
Our Acc there is higher than on the equirectangular cell in all four datasets, and on PanoCity it is the only wide cell where Any3R is tied rather than best.
Equirectangular images are $83.6\%$ of the training mixture, so a panoramic format that is not equirectangular is where that mixture shows.

\cref{tab::p13ctrl} holds the field of view fixed and changes only how angle maps to radius.
Any3R is best in $15$ of its $16$ cells and tied with $\pi^3$ in the remaining one, the Structured3D Equisolid cell.
Each value also carries the method's own pose error, and that error differs between the two members of a pair for every method, ours included, so under predicted poses the table ranks the methods under each mapping but does not isolate the effect of the mapping itself.

\begin{table*}[!tb]\centering\footnotesize
\setlength{\tabcolsep}{1.4pt}
\caption{\textbf{Accuracy on all $52$ cells.} Acc$\downarrow$ (m) per camera, grouped by field of view; the five cameras marked $\dagger$ each isolate one axis the original eight leave confounded.}
\label{tab::p13full}
\begin{tabular}{lcccccccccccccc}
\toprule
 & \multicolumn{7}{c}{Matterport3D} & \multicolumn{7}{c}{Stanford2D3DS} \\
\cmidrule(lr){2-8}\cmidrule(lr){9-15}
Camera (FoV) & Any3R & DA3{\footnotesize\cite{depthanything3}} & VGGT{\footnotesize\cite{wang2025vggt}} & $\pi^3${\footnotesize\cite{wang2025pi3}} & V-$\Omega${\footnotesize\cite{wang2026vggtomega}} & MapA{\footnotesize\cite{keetha2025mapanything}} & PanoV{\footnotesize\cite{panovggt2026}} & Any3R & DA3 & VGGT & $\pi^3$ & V-$\Omega$ & MapA & PanoV \\
\midrule
Pinhole (90$^\circ$) & \cellcolor{best}\textbf{0.069} & 0.300 & 0.288 & 0.109 & 0.161 & \cellcolor{second}0.089 & 0.281 & \cellcolor{second}0.083 & 0.325 & 0.252 & \cellcolor{best}\textbf{0.059} & 0.138 & 0.095 & 0.284 \\
Brown--C. (90$^\circ$) & \cellcolor{best}\textbf{0.096} & 0.284 & 0.320 & \cellcolor{second}0.117 & 0.139 & 0.150 & 0.261 & \cellcolor{best}\textbf{0.085} & 0.303 & 0.243 & \cellcolor{second}0.120 & 0.184 & 0.149 & 0.284 \\
\midrule
FOV/ATAN$^\dagger$ (150$^\circ$) & \cellcolor{best}\textbf{0.051} & 0.330 & 0.331 & \cellcolor{second}0.139 & 0.176 & 0.229 & 0.259 & \cellcolor{best}\textbf{0.053} & 0.309 & 0.344 & \cellcolor{second}0.133 & 0.261 & 0.243 & 0.281 \\
\midrule
Equidist. (180$^\circ$) & \cellcolor{best}\textbf{0.056} & 0.360 & 0.343 & \cellcolor{second}0.166 & 0.215 & 0.246 & 0.248 & \cellcolor{best}\textbf{0.061} & 0.363 & 0.380 & \cellcolor{second}0.177 & 0.288 & 0.299 & 0.270 \\
Stereogr.$^\dagger$ (180$^\circ$) & \cellcolor{best}\textbf{0.060} & 0.349 & 0.340 & \cellcolor{second}0.144 & 0.246 & 0.269 & 0.262 & \cellcolor{best}\textbf{0.069} & 0.384 & 0.399 & \cellcolor{second}0.177 & 0.328 & 0.295 & 0.293 \\
ERP-Cr. (180$^\circ$) & \cellcolor{best}\textbf{0.051} & 0.267 & 0.261 & 0.265 & 0.245 & \cellcolor{second}0.242 & 0.331 & \cellcolor{best}\textbf{0.053} & 0.309 & 0.287 & 0.310 & \cellcolor{second}0.286 & 0.296 & 0.364 \\
MEI (180$^\circ$) & \cellcolor{best}\textbf{0.054} & 0.361 & 0.371 & \cellcolor{second}0.163 & 0.212 & 0.245 & 0.250 & \cellcolor{best}\textbf{0.064} & 0.365 & 0.387 & \cellcolor{second}0.165 & 0.288 & 0.289 & 0.281 \\
\midrule
KB4 (200$^\circ$) & \cellcolor{best}\textbf{0.050} & 0.388 & 0.344 & \cellcolor{second}0.175 & 0.243 & 0.246 & 0.240 & \cellcolor{best}\textbf{0.061} & 0.404 & 0.368 & \cellcolor{second}0.203 & 0.320 & 0.320 & 0.261 \\
Equisolid$^\dagger$ (200$^\circ$) & \cellcolor{best}\textbf{0.056} & 0.382 & 0.371 & \cellcolor{second}0.201 & 0.254 & 0.226 & 0.233 & \cellcolor{best}\textbf{0.055} & 0.364 & 0.362 & \cellcolor{second}0.213 & 0.305 & 0.291 & 0.250 \\
Double-Sph. (200$^\circ$) & \cellcolor{best}\textbf{0.055} & 0.368 & 0.358 & \cellcolor{second}0.165 & 0.259 & 0.259 & 0.249 & \cellcolor{best}\textbf{0.065} & 0.389 & 0.390 & \cellcolor{second}0.201 & 0.344 & 0.328 & 0.282 \\
EUCM$^\dagger$ (200$^\circ$) & \cellcolor{best}\textbf{0.054} & 0.364 & 0.370 & \cellcolor{second}0.168 & 0.286 & 0.261 & 0.244 & \cellcolor{best}\textbf{0.067} & 0.369 & 0.392 & \cellcolor{second}0.196 & 0.344 & 0.331 & 0.267 \\
\midrule
ERP (360$^\circ$) & \cellcolor{best}\textbf{0.045} & 0.390 & 0.401 & 0.419 & 0.386 & 0.342 & \cellcolor{second}0.077 & \cellcolor{best}\textbf{0.064} & 0.415 & 0.395 & 0.408 & 0.421 & 0.430 & \cellcolor{second}0.066 \\
Cylind.$^\dagger$ (360$^\circ$) & \cellcolor{best}\textbf{0.082} & 0.343 & 0.364 & 0.342 & 0.342 & \cellcolor{second}0.312 & 0.415 & \cellcolor{best}\textbf{0.098} & \cellcolor{second}0.366 & 0.385 & 0.377 & 0.396 & 0.367 & 0.484 \\
\midrule
 & \multicolumn{7}{c}{Structured3D} & \multicolumn{7}{c}{PanoCity} \\
\cmidrule(lr){2-8}\cmidrule(lr){9-15}
Camera (FoV) & Any3R & DA3 & VGGT & $\pi^3$ & V-$\Omega$ & MapA & PanoV & Any3R & DA3 & VGGT & $\pi^3$ & V-$\Omega$ & MapA & PanoV \\
\midrule
Pinhole (90$^\circ$) & 0.757 & \cellcolor{best}\textbf{0.665} & 0.781 & 0.819 & \cellcolor{second}0.729 & 0.774 & 0.812 & \cellcolor{best}\textbf{1.160} & 4.360 & 4.153 & \cellcolor{second}1.479 & 1.711 & 1.485 & 2.256 \\
Brown--C. (90$^\circ$) & 0.706 & 0.716 & \cellcolor{best}\textbf{0.645} & \cellcolor{second}0.647 & 0.845 & 0.671 & 0.771 & \cellcolor{best}\textbf{1.243} & 5.689 & 3.940 & 1.711 & \cellcolor{second}1.572 & 1.633 & 2.708 \\
\midrule
FOV/ATAN$^\dagger$ (150$^\circ$) & \cellcolor{best}\textbf{0.522} & 0.614 & 0.623 & 0.683 & 0.678 & \cellcolor{second}0.533 & 0.730 & \cellcolor{best}\textbf{0.929} & 5.700 & 6.161 & \cellcolor{second}1.856 & 1.944 & 1.930 & 2.368 \\
\midrule
Equidist. (180$^\circ$) & \cellcolor{best}\textbf{0.443} & 0.539 & 0.623 & \cellcolor{second}0.531 & 0.693 & 0.773 & 0.542 & \cellcolor{best}\textbf{0.866} & 6.656 & 6.296 & 2.093 & 2.162 & \cellcolor{second}2.066 & 2.498 \\
Stereogr.$^\dagger$ (180$^\circ$) & \cellcolor{best}\textbf{0.451} & 0.613 & 0.636 & 0.582 & 0.636 & \cellcolor{second}0.549 & 0.615 & \cellcolor{best}\textbf{0.917} & 6.398 & 5.684 & \cellcolor{second}1.991 & 2.624 & 2.383 & 2.501 \\
ERP-Cr. (180$^\circ$) & \cellcolor{best}\textbf{0.428} & 0.610 & \cellcolor{second}0.513 & 0.611 & 0.622 & 0.555 & 0.601 & \cellcolor{best}\textbf{0.811} & 2.717 & 3.078 & \cellcolor{second}2.426 & 2.455 & 2.861 & 3.633 \\
MEI (180$^\circ$) & \cellcolor{best}\textbf{0.436} & 0.549 & 0.637 & \cellcolor{second}0.523 & 0.572 & 0.527 & 0.533 & \cellcolor{best}\textbf{0.854} & 6.844 & 6.886 & \cellcolor{second}2.036 & 2.490 & 2.236 & 2.433 \\
\midrule
KB4 (200$^\circ$) & \cellcolor{best}\textbf{0.294} & 0.563 & 0.635 & \cellcolor{second}0.508 & 0.653 & 0.519 & 0.568 & \cellcolor{best}\textbf{0.863} & 6.463 & 6.623 & \cellcolor{second}2.394 & 2.680 & 3.297 & 2.405 \\
Equisolid$^\dagger$ (200$^\circ$) & \cellcolor{best}\textbf{0.310} & 0.519 & 0.541 & \cellcolor{second}0.467 & 0.643 & 0.624 & 0.512 & \cellcolor{best}\textbf{0.893} & 6.336 & 6.280 & 2.425 & 3.065 & \cellcolor{second}2.268 & 2.316 \\
Double-Sph. (200$^\circ$) & \cellcolor{best}\textbf{0.340} & 0.624 & 0.597 & 0.543 & 0.602 & \cellcolor{second}0.502 & 0.596 & \cellcolor{best}\textbf{0.901} & 6.477 & 6.430 & \cellcolor{second}2.271 & 2.544 & 2.789 & 2.473 \\
EUCM$^\dagger$ (200$^\circ$) & \cellcolor{best}\textbf{0.345} & 0.587 & 0.612 & 0.545 & 0.587 & \cellcolor{second}0.544 & 0.586 & \cellcolor{best}\textbf{0.896} & 6.189 & 7.097 & \cellcolor{second}2.381 & 2.613 & 2.690 & 2.504 \\
\midrule
ERP (360$^\circ$) & \cellcolor{best}\textbf{0.128} & 0.553 & 0.442 & 0.445 & 0.496 & 0.440 & \cellcolor{second}0.144 & \cellcolor{best}\textbf{0.663} & 3.149 & 1.902 & 1.628 & 1.876 & 1.833 & \cellcolor{second}0.751 \\
Cylind.$^\dagger$ (360$^\circ$) & \cellcolor{best}\textbf{0.275} & 0.467 & 0.470 & 0.468 & 0.476 & \cellcolor{second}0.421 & 0.535 & \cellcolor{best}\textbf{1.298} & 4.160 & 3.657 & 2.785 & 3.104 & \cellcolor{second}2.680 & 4.216 \\
\bottomrule
\end{tabular}\end{table*}

\begin{table*}[!tb]\centering\footnotesize
\footnotesize
\setlength{\tabcolsep}{1.4pt}
\caption{\textbf{Both inputs of the two-input baselines.} Median per-scene Acc$\downarrow$ (m) for the native and the official input of $\pi^3$, VGGT and VGGT-$\Omega$ and for MapAnything without and with pinhole intrinsics; every other table uses the \textbf{bold} one.}
\label{tab::p13arms}
\begin{tabular}{lcccccccccccccccc}
\toprule
 & \multicolumn{8}{c}{Matterport3D} & \multicolumn{8}{c}{Stanford2D3DS} \\
\cmidrule(lr){2-9}\cmidrule(lr){10-17}
 & \multicolumn{2}{c}{$\pi^3$} & \multicolumn{2}{c}{VGGT} & \multicolumn{2}{c}{V-$\Omega$} & \multicolumn{2}{c}{MapA} & \multicolumn{2}{c}{$\pi^3$} & \multicolumn{2}{c}{VGGT} & \multicolumn{2}{c}{V-$\Omega$} & \multicolumn{2}{c}{MapA} \\
Camera & nat. & off. & nat. & off. & nat. & off. & $-$K & +K & nat. & off. & nat. & off. & nat. & off. & $-$K & +K \\
\midrule
Pinhole (90$^\circ$) & 0.050 & \textbf{0.041} & 0.279 & \textbf{0.252} & 0.128 & \textbf{0.101} & 0.125 & \textbf{0.069} & 0.045 & \textbf{0.041} & 0.291 & \textbf{0.239} & 0.118 & \textbf{0.087} & 0.119 & \textbf{0.062} \\
Brown--C. (90$^\circ$) & 0.080 & \textbf{0.072} & 0.295 & \textbf{0.274} & 0.175 & \textbf{0.126} & \textbf{0.151} & -- & 0.071 & \textbf{0.068} & 0.338 & \textbf{0.247} & 0.177 & \textbf{0.122} & \textbf{0.121} & -- \\
FOV/ATAN (150$^\circ$) & \textbf{0.119} & 0.127 & 0.286 & \textbf{0.279} & \textbf{0.155} & 0.158 & \textbf{0.221} & -- & 0.115 & \textbf{0.108} & 0.352 & \textbf{0.333} & \textbf{0.206} & 0.211 & \textbf{0.236} & -- \\
Equidist. (180$^\circ$) & 0.161 & \textbf{0.139} & 0.347 & \textbf{0.313} & 0.211 & \textbf{0.195} & \textbf{0.265} & -- & 0.141 & \textbf{0.141} & 0.415 & \textbf{0.383} & \textbf{0.257} & 0.265 & \textbf{0.296} & -- \\
Stereogr. (180$^\circ$) & 0.121 & \textbf{0.118} & \textbf{0.332} & 0.333 & 0.249 & \textbf{0.245} & \textbf{0.272} & -- & \textbf{0.131} & 0.152 & 0.406 & \textbf{0.397} & \textbf{0.295} & 0.304 & \textbf{0.296} & -- \\
ERP-Cr. (180$^\circ$) & \textbf{0.243} & 0.249 & \textbf{0.243} & 0.245 & \textbf{0.239} & 0.250 & \textbf{0.212} & -- & 0.282 & \textbf{0.268} & 0.265 & \textbf{0.251} & \textbf{0.245} & 0.274 & \textbf{0.268} & -- \\
MEI (180$^\circ$) & 0.150 & \textbf{0.139} & 0.353 & \textbf{0.332} & \textbf{0.201} & 0.216 & \textbf{0.249} & -- & \textbf{0.130} & 0.134 & \textbf{0.376} & 0.389 & 0.263 & \textbf{0.258} & \textbf{0.268} & -- \\
KB4 (200$^\circ$) & 0.180 & \textbf{0.146} & 0.351 & \textbf{0.339} & \textbf{0.231} & 0.240 & \textbf{0.239} & -- & \textbf{0.150} & 0.151 & 0.389 & \textbf{0.366} & \textbf{0.317} & 0.323 & \textbf{0.287} & -- \\
Equisolid (200$^\circ$) & 0.218 & \textbf{0.187} & \textbf{0.335} & 0.336 & \textbf{0.231} & 0.262 & \textbf{0.221} & -- & 0.178 & \textbf{0.163} & 0.370 & \textbf{0.353} & \textbf{0.294} & 0.327 & \textbf{0.272} & -- \\
Double-Sph. (200$^\circ$) & 0.151 & \textbf{0.144} & 0.353 & \textbf{0.331} & \textbf{0.256} & 0.268 & \textbf{0.266} & -- & \textbf{0.146} & 0.153 & \textbf{0.386} & 0.397 & \textbf{0.334} & 0.335 & \textbf{0.323} & -- \\
EUCM (200$^\circ$) & 0.159 & \textbf{0.148} & 0.346 & \textbf{0.335} & 0.265 & \textbf{0.265} & \textbf{0.271} & -- & \textbf{0.149} & 0.156 & \textbf{0.381} & 0.389 & 0.325 & \textbf{0.314} & \textbf{0.309} & -- \\
ERP (360$^\circ$) & 0.387 & \textbf{0.374} & \textbf{0.348} & 0.352 & \textbf{0.351} & 0.376 & \textbf{0.347} & -- & 0.405 & \textbf{0.382} & \textbf{0.388} & 0.413 & \textbf{0.392} & 0.401 & \textbf{0.428} & -- \\
Cylind. (360$^\circ$) & 0.342 & \textbf{0.274} & \textbf{0.302} & 0.337 & 0.361 & \textbf{0.347} & \textbf{0.310} & -- & \textbf{0.356} & 0.370 & \textbf{0.374} & 0.400 & \textbf{0.360} & 0.370 & \textbf{0.334} & -- \\
\midrule
 & \multicolumn{8}{c}{Structured3D} & \multicolumn{8}{c}{PanoCity} \\
\cmidrule(lr){2-9}\cmidrule(lr){10-17}
 & \multicolumn{2}{c}{$\pi^3$} & \multicolumn{2}{c}{VGGT} & \multicolumn{2}{c}{V-$\Omega$} & \multicolumn{2}{c}{MapA} & \multicolumn{2}{c}{$\pi^3$} & \multicolumn{2}{c}{VGGT} & \multicolumn{2}{c}{V-$\Omega$} & \multicolumn{2}{c}{MapA} \\
Camera & nat. & off. & nat. & off. & nat. & off. & $-$K & +K & nat. & off. & nat. & off. & nat. & off. & $-$K & +K \\
\midrule
Pinhole (90$^\circ$) & \textbf{0.476} & 0.595 & \textbf{0.485} & 0.575 & \textbf{0.380} & 0.531 & \textbf{0.410} & 0.522 & \textbf{1.053} & 1.078 & 4.222 & \textbf{3.724} & 1.553 & \textbf{1.344} & 1.532 & \textbf{1.161} \\
Brown--C. (90$^\circ$) & 0.419 & \textbf{0.341} & \textbf{0.469} & 0.532 & \textbf{0.484} & 0.515 & \textbf{0.469} & -- & 1.339 & \textbf{1.319} & \textbf{3.782} & 3.976 & 2.054 & \textbf{1.321} & \textbf{1.477} & -- \\
FOV/ATAN (150$^\circ$) & \textbf{0.496} & 0.516 & 0.562 & \textbf{0.468} & \textbf{0.489} & 0.496 & \textbf{0.403} & -- & 1.795 & \textbf{1.720} & 6.213 & \textbf{5.701} & \textbf{1.852} & 1.927 & \textbf{1.904} & -- \\
Equidist. (180$^\circ$) & \textbf{0.355} & 0.358 & 0.592 & \textbf{0.497} & \textbf{0.484} & 0.568 & \textbf{0.427} & -- & 2.171 & \textbf{2.000} & 6.639 & \textbf{5.861} & \textbf{2.184} & 2.244 & \textbf{2.023} & -- \\
Stereogr. (180$^\circ$) & \textbf{0.456} & 0.521 & \textbf{0.451} & 0.490 & 0.487 & \textbf{0.479} & \textbf{0.432} & -- & 1.893 & \textbf{1.765} & 5.925 & \textbf{5.615} & 2.386 & \textbf{2.190} & \textbf{2.300} & -- \\
ERP-Cr. (180$^\circ$) & \textbf{0.546} & 0.581 & 0.538 & \textbf{0.443} & 0.522 & \textbf{0.511} & \textbf{0.480} & -- & \textbf{2.509} & 2.641 & \textbf{3.072} & 3.250 & 2.773 & \textbf{2.694} & \textbf{3.031} & -- \\
MEI (180$^\circ$) & \textbf{0.326} & 0.386 & 0.584 & \textbf{0.498} & \textbf{0.523} & 0.561 & \textbf{0.417} & -- & 2.143 & \textbf{1.966} & \textbf{5.992} & 6.418 & 2.295 & \textbf{2.277} & \textbf{2.128} & -- \\
KB4 (200$^\circ$) & \textbf{0.345} & 0.356 & \textbf{0.492} & 0.511 & \textbf{0.464} & 0.469 & \textbf{0.417} & -- & 2.781 & \textbf{2.226} & 7.146 & \textbf{6.388} & \textbf{2.686} & 2.708 & \textbf{2.459} & -- \\
Equisolid (200$^\circ$) & \textbf{0.326} & 0.334 & 0.463 & \textbf{0.459} & \textbf{0.433} & 0.524 & \textbf{0.412} & -- & \textbf{2.229} & 3.054 & 6.697 & \textbf{6.098} & \textbf{2.984} & 3.244 & \textbf{2.179} & -- \\
Double-Sph. (200$^\circ$) & \textbf{0.398} & 0.409 & 0.463 & \textbf{0.451} & 0.449 & \textbf{0.442} & \textbf{0.414} & -- & 2.519 & \textbf{2.141} & \textbf{6.047} & 6.218 & \textbf{2.387} & 2.531 & \textbf{2.390} & -- \\
EUCM (200$^\circ$) & 0.447 & \textbf{0.402} & 0.589 & \textbf{0.469} & 0.509 & \textbf{0.417} & \textbf{0.395} & -- & 2.543 & \textbf{2.138} & 7.112 & \textbf{6.369} & \textbf{2.450} & 2.492 & \textbf{2.556} & -- \\
ERP (360$^\circ$) & \textbf{0.381} & 0.504 & \textbf{0.383} & 0.511 & \textbf{0.432} & 0.453 & \textbf{0.322} & -- & 1.097 & \textbf{0.853} & 1.046 & \textbf{0.978} & \textbf{1.073} & 1.105 & \textbf{0.905} & -- \\
Cylind. (360$^\circ$) & \textbf{0.414} & 0.449 & \textbf{0.369} & 0.419 & \textbf{0.423} & 0.449 & \textbf{0.372} & -- & 3.132 & \textbf{2.011} & 2.812 & \textbf{2.537} & 2.063 & \textbf{1.884} & \textbf{1.838} & -- \\
\bottomrule
\end{tabular}\end{table*}

\begin{table*}[!tb]\centering\footnotesize
\setlength{\tabcolsep}{1.6pt}
\caption{\textbf{Two radial mappings at the same field of view.} Acc$\downarrow$ (m) of two camera pairs covering the same solid angle (KB4 and Equisolid at $200^\circ$, Equidistant and Stereographic at $180^\circ$), with the $\overline{\mathrm{R}}_\text{med}$ (degrees) of the same arms below, so a change of Acc can be read against its change of pose.}
\label{tab::p13ctrl}
\begin{tabular}{lcccccccccccccccc}
\toprule
 & \multicolumn{4}{c}{Matterport3D} & \multicolumn{4}{c}{Stanford2D3DS} & \multicolumn{4}{c}{Structured3D} & \multicolumn{4}{c}{PanoCity} \\
\cmidrule(lr){2-5}\cmidrule(lr){6-9}\cmidrule(lr){10-13}\cmidrule(lr){14-17}
Method & KB4 & Eqs. & Eqd. & Str. & KB4 & Eqs. & Eqd. & Str. & KB4 & Eqs. & Eqd. & Str. & KB4 & Eqs. & Eqd. & Str. \\
\midrule
\multicolumn{17}{l}{\emph{Acc$\downarrow$ (m)}} \\
Any3R & \cellcolor{best}\textbf{0.050} & \cellcolor{best}\textbf{0.056} & \cellcolor{best}\textbf{0.056} & \cellcolor{best}\textbf{0.060} & \cellcolor{best}\textbf{0.061} & \cellcolor{best}\textbf{0.055} & \cellcolor{best}\textbf{0.061} & \cellcolor{best}\textbf{0.069} & \cellcolor{best}\textbf{0.294} & \cellcolor{best}\textbf{0.310} & \cellcolor{best}\textbf{0.443} & \cellcolor{best}\textbf{0.451} & \cellcolor{best}\textbf{0.863} & \cellcolor{best}\textbf{0.893} & \cellcolor{best}\textbf{0.866} & \cellcolor{best}\textbf{0.917} \\
DA3~\cite{depthanything3} & 0.388 & 0.382 & 0.360 & 0.349 & 0.404 & 0.364 & 0.363 & 0.384 & 0.563 & 0.519 & 0.539 & 0.613 & 6.463 & 6.336 & 6.656 & 6.398 \\
VGGT~\cite{wang2025vggt} & 0.344 & 0.371 & 0.343 & 0.340 & 0.368 & 0.362 & 0.380 & 0.399 & 0.635 & 0.541 & 0.623 & 0.636 & 6.623 & 6.280 & 6.296 & 5.684 \\
$\pi^3$~\cite{wang2025pi3} & \cellcolor{second}0.175 & \cellcolor{second}0.201 & \cellcolor{second}0.166 & \cellcolor{second}0.144 & \cellcolor{second}0.203 & \cellcolor{second}0.213 & \cellcolor{second}0.177 & \cellcolor{second}0.177 & \cellcolor{second}0.508 & \cellcolor{second}0.467 & \cellcolor{second}0.531 & 0.582 & \cellcolor{second}2.394 & 2.425 & 2.093 & \cellcolor{second}1.991 \\
VGGT-$\Omega$~\cite{wang2026vggtomega} & 0.243 & 0.254 & 0.215 & 0.246 & 0.320 & 0.305 & 0.288 & 0.328 & 0.653 & 0.643 & 0.693 & 0.636 & 2.680 & 3.065 & 2.162 & 2.624 \\
MapAnything~\cite{keetha2025mapanything} & 0.246 & 0.226 & 0.246 & 0.269 & 0.320 & 0.291 & 0.299 & 0.295 & 0.519 & 0.624 & 0.773 & \cellcolor{second}0.549 & 3.297 & \cellcolor{second}2.268 & \cellcolor{second}2.066 & 2.383 \\
PanoVGGT~\cite{panovggt2026} & 0.240 & 0.233 & 0.248 & 0.262 & 0.261 & 0.250 & 0.270 & 0.293 & 0.568 & 0.512 & 0.542 & 0.615 & 2.405 & 2.316 & 2.498 & 2.501 \\
\midrule
\multicolumn{17}{l}{\emph{$\overline{\mathrm{R}}_\text{med}\downarrow$ ($^\circ$) of the same arm}} \\
Any3R & 3.10 & 7.36 & 8.61 & 9.03 & 6.13 & 2.73 & 4.82 & 5.26 & 12.72 & 12.75 & 29.54 & 23.21 & 3.49 & 3.74 & 2.69 & 3.47 \\
DA3 & 49.45 & 58.80 & 55.00 & 59.44 & 50.54 & 42.27 & 67.73 & 48.43 & 96.91 & 86.65 & 95.54 & 88.26 & 38.78 & 39.03 & 38.12 & 41.40 \\
VGGT & 47.60 & 60.52 & 47.20 & 36.31 & 47.09 & 52.73 & 49.77 & 40.01 & 115.44 & 91.15 & 101.77 & 81.11 & 55.71 & 47.36 & 60.06 & 52.56 \\
$\pi^3$ & 14.96 & 14.59 & 12.20 & 10.89 & 25.88 & 29.76 & 18.93 & 16.22 & 37.52 & 42.28 & 39.78 & 43.62 & 37.06 & 80.55 & 34.73 & 26.32 \\
VGGT-$\Omega$ & 20.85 & 20.71 & 18.92 & 22.08 & 36.78 & 35.62 & 34.08 & 35.88 & 40.27 & 39.13 & 38.66 & 34.43 & 39.31 & 51.54 & 33.50 & 34.36 \\
MapAnything & 32.54 & 34.02 & 33.63 & 33.86 & 41.89 & 44.74 & 43.03 & 43.40 & 49.66 & 59.14 & 41.73 & 44.40 & 41.86 & 34.06 & 27.59 & 43.24 \\
PanoVGGT & 9.21 & 13.41 & 13.11 & 8.89 & 15.95 & 20.03 & 18.56 & 19.27 & 30.47 & 19.23 & 16.84 & 26.58 & 7.45 & 10.20 & 12.06 & 10.03 \\
\bottomrule
\end{tabular}\end{table*}

\section{Per-Point Error}
\label{sec::supperr} 
\begin{figure*}[!tb]\centering
\includegraphics[width=\textwidth]{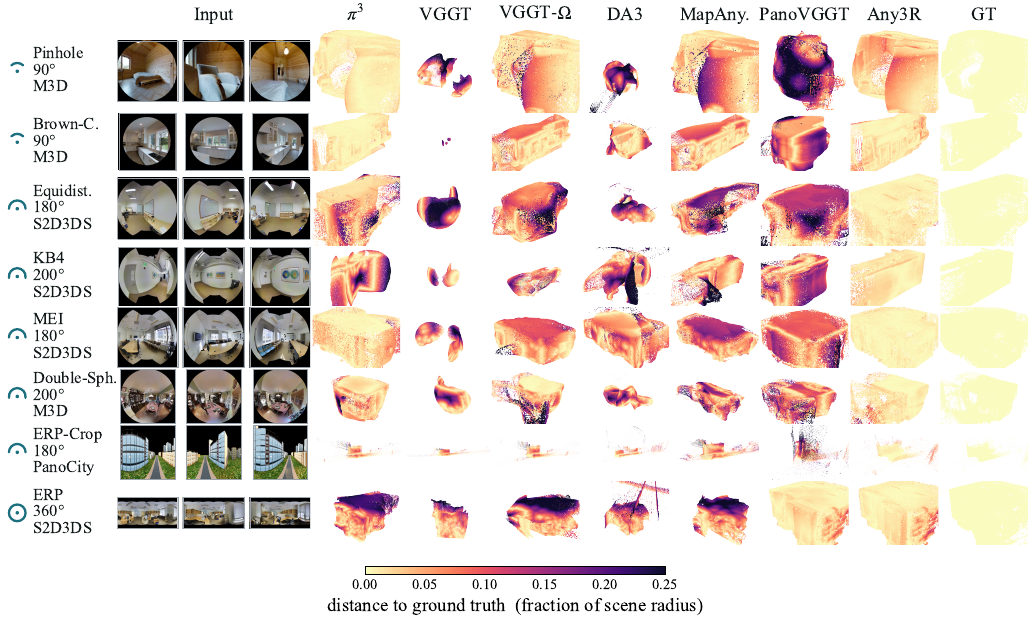}
\caption{\textbf{Per-point error on the eight camera models.} Same cells and same viewpoints as \cref{fig::cams}, with colour giving each point's distance to the ground truth as a fraction of the scene radius. The colour scale is shared across all cells.}
\label{fig::supperr}
\end{figure*}

\cref{fig::cams} colours each point with the scene's own texture, which shows what a reconstruction looks like but not how wrong it is.
\cref{fig::supperr} redraws the same cells with every point coloured by its distance to the ground truth, normalised by the scene radius, which is the quantity Acc averages.
Two things are visible here that the averages in \cref{tab::main} cannot show.
The first is that the error of the perspective-native baselines on the wide lenses covers whole walls rather than a few points, because each fused cloud carries its method's own pose error and a misplaced view moves every point it contributes, while our error stays low across the room on the wide lenses.
The second is that the failures are structured rather than noisy.
On the wide cells the baselines return smooth surfaces that are confidently wrong, bent into a shell or folded inwards, rather than a scattered cloud.

\begin{table*}[!tb]\renewenvironment{table}[1][]{}{}
\begin{minipage}[t]{0.485\textwidth}
\begin{table}[tb]\centering\footnotesize
\setlength{\tabcolsep}{1.6pt}
\caption{\textbf{One checkpoint under three ray fields.} Acc$\downarrow$ (m) averaged over the eight cameras of \cref{tab::main}; Ang.\ err.\ is the mean angle between the field used and the true one, measured on the earlier per-camera pool.}
\label{tab::calib}
\begin{tabular}{lccccc}
\toprule
Ray field & Ang.\ err. & M3D & S2D3DS & S3D & PanoCity \\
\midrule
True & $0.0^\circ$ & \cellcolor{best}\textbf{0.059} & \cellcolor{best}\textbf{0.067} & \cellcolor{best}\textbf{0.442} & \cellcolor{best}\textbf{0.920} \\
Predicted & $7.9^\circ$ & \cellcolor{second}0.192 & 0.278 & 0.621 & 3.378 \\
ERP fallback & $35.8^\circ$ & 0.243 & \cellcolor{second}0.263 & \cellcolor{second}0.579 & \cellcolor{second}2.299 \\
\bottomrule
\end{tabular}\end{table}
\end{minipage}\hfill
\begin{minipage}[t]{0.485\textwidth}
\providecommand{\poseOursAuc}{}\renewcommand{\poseOursAuc}{0.7211}
\providecommand{\poseOursRmed}{}\renewcommand{\poseOursRmed}{3.00$^\circ$}
\providecommand{\poseOursCol}{}\renewcommand{\poseOursCol}{2}
\providecommand{\poseBestName}{}\renewcommand{\poseBestName}{DA3}
\providecommand{\poseBestAuc}{}\renewcommand{\poseBestAuc}{0.9022}
\providecommand{\poseScenes}{}\renewcommand{\poseScenes}{60}
\providecommand{\posePanoAuc}{}\renewcommand{\posePanoAuc}{0.0000}
\providecommand{\posePanoCol}{}\renewcommand{\posePanoCol}{58/60}

\begin{table}[tb]\centering\footnotesize
\setlength{\tabcolsep}{2.4pt}
\caption{\textbf{Camera pose on DL3DV,} 60 scenes of real perspective photographs at 3 views each. AUC of the pairwise pose error at $5^\circ$ and $30^\circ$, and $\overline{\mathrm{R}}_\text{med}$, the per-scene median rotation error averaged over scenes.}
\label{tab::pose}
\begin{tabular}{lccc}
\toprule
Method & AUC@$5^\circ\uparrow$ & AUC@$30^\circ\uparrow$ & $\overline{\mathrm{R}}_\text{med}\downarrow$ \\
\midrule
DA3~\cite{depthanything3} & \cellcolor{best}\textbf{0.9022} & \cellcolor{best}\textbf{0.9543} & 3.32$^\circ$ \\
MapAnything~\cite{keetha2025mapanything} & \cellcolor{second}0.7567 & 0.9196 & 8.69$^\circ$ \\
$\pi^3$~\cite{wang2025pi3} & 0.7400 & 0.9206 & 8.82$^\circ$ \\
\textbf{Any3R} & 0.7211 & \cellcolor{second}0.9313 & \cellcolor{best}\textbf{3.00$^\circ$} \\
VGGT~\cite{wang2025vggt} & 0.7011 & 0.9300 & \cellcolor{second}3.06$^\circ$ \\
VGGT-$\Omega$~\cite{wang2026vggtomega} & 0.6011 & 0.9135 & 3.34$^\circ$ \\
PanoVGGT~\cite{panovggt2026} & 0.0000 & 0.0515 & 51.18$^\circ$ \\
\bottomrule
\end{tabular}\end{table}

\end{minipage}
\end{table*}
\section{The Ray Map Withdrawn}
\label{sec::suppcalib} The last row is the part worth dwelling on.
On equirectangular input the fallback and the true ray field coincide up to floating-point rounding, so the two columns are not merely close, they are the same computation, and their Acc agrees to three decimals on all four datasets.
Everything the module does elsewhere therefore costs nothing on the input the backbone was pretrained for.
Read together with the rest of the figure, the pair says that the gain is not a general improvement to the backbone but a correction that activates exactly where the pretrained assumption stops holding, and is inert where it still holds.

\begin{figure*}[!tb]\centering
\includegraphics[width=\textwidth]{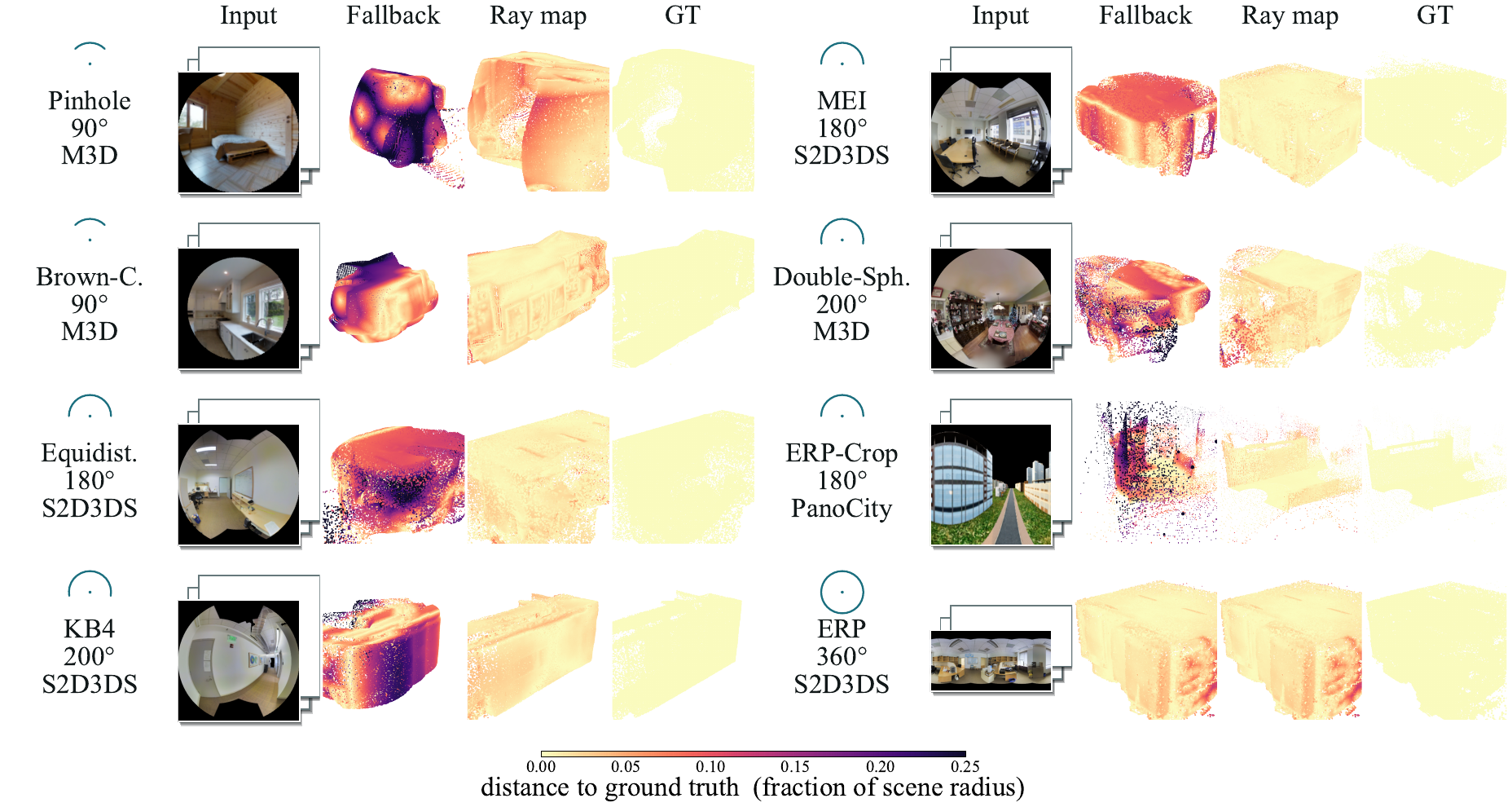}
\caption{\textbf{One checkpoint with and without the ray map.} \textit{Ray map} is the true ray field and \textit{Fallback} is the equirectangular one the backbone assumes when none is supplied. Colour is distance to the ground truth as a fraction of the scene radius, on the scale of \cref{fig::supperr}. Every camera degrades when the ray map is withdrawn except the equirectangular one in the last row, where the fallback coincides with the true field and nothing changes.}
\label{fig::suppcalib}
\end{figure*}

\cref{tab::calib} gives the Acc of one checkpoint under three ray maps, and without the ray map its error grows on every dataset.
A small CNN trained leave-one-dataset-out to predict each view's camera type and field of view gives a far more accurate map than the fallback, yet recovers part of the gap only on Matterport3D and is worse than the fallback elsewhere, so the backbone needs an accurate map rather than an approximate one.
The decoder places each point along its ray, so a direction error moves a point in proportion to its distance, which is largest on PanoCity, whose outdoor scenes the predictor never saw, and the predictor also replaces the exact map of the equirectangular frames, where the fallback needed no correction.
\cref{fig::suppcalib} shows two of them on the same frames.
With no calibration the backbone falls back to assuming an equirectangular grid, so a $90^\circ$ pinhole image is treated as a full sphere, and the geometry is wrong over most of the frame.
The bottom row is the internal control: on equirectangular input the fallback is the correct ray map, and the two columns are indistinguishable.
That row is the exactness constraint of \cref{sec::arope} made visible.

\section{The Camera Models and How Their Images Are Made}
\label{sec::suppcams}

\begin{figure*}[!tb]\centering
\includegraphics[width=\textwidth]{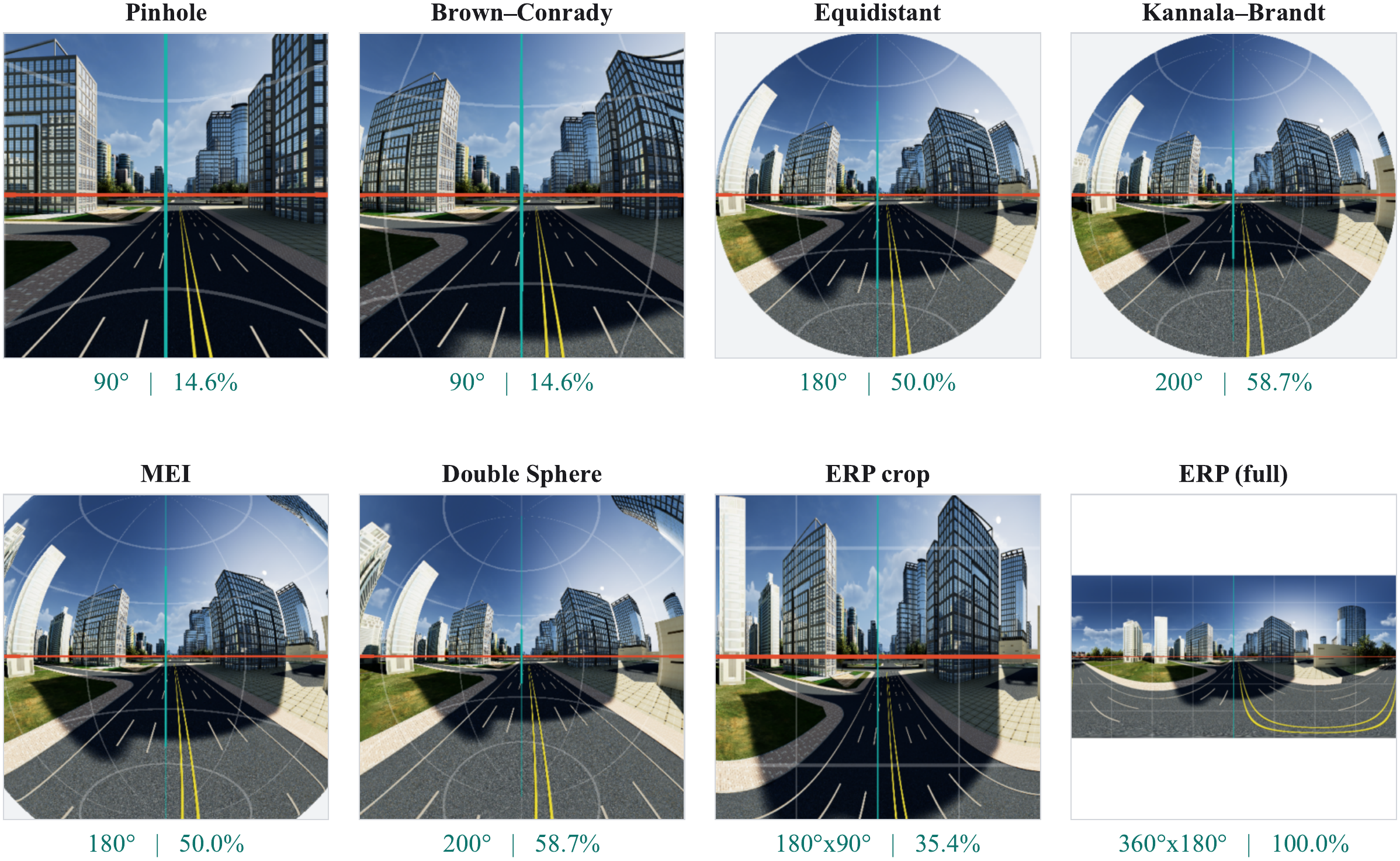}
\caption{\textbf{The eight central camera models.} The same $360^\circ$ panorama imaged through each lens, so every difference in distortion and in coverage comes from the camera model alone. Under each panel: field of view, then the fraction of the sphere that lens observes. Overlaid graticule: equator in red, prime meridian in teal, $45^\circ$ meridians and $30^\circ$ parallels in white, and the region outside the image circle in light grey.}
\label{fig::suppcams}
\end{figure*}

\begin{figure*}[!tb]\centering
\includegraphics[width=\textwidth]{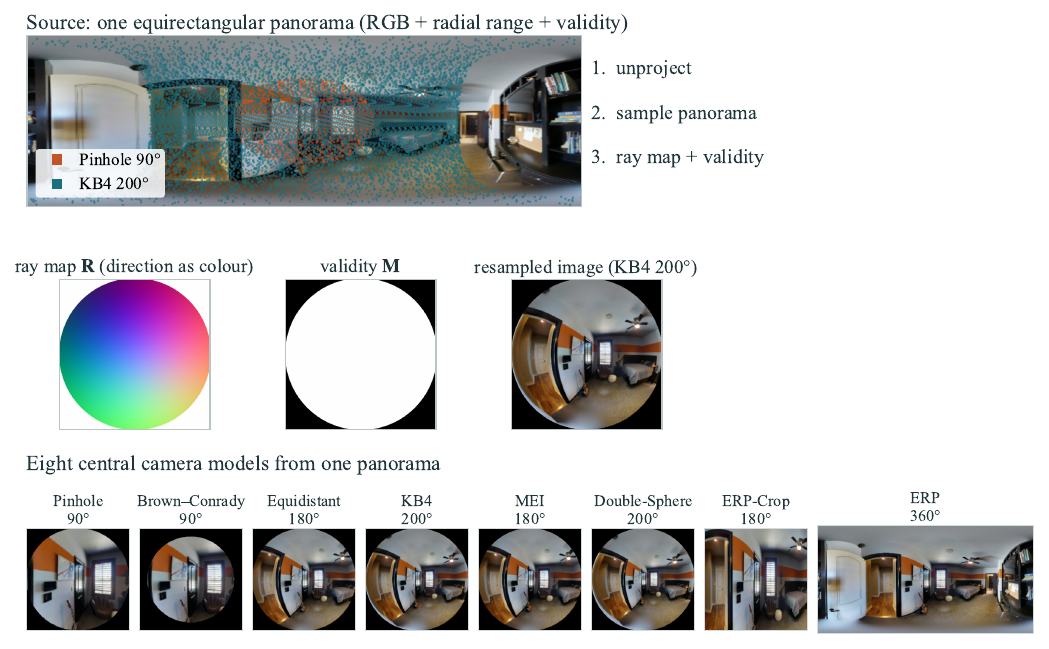}
\caption{\textbf{From one panorama to a native camera frame.} Top: the source panorama with the footprints of a $90^\circ$ pinhole and a $200^\circ$ fisheye marked. Middle: the ray map with direction shown as colour, the validity mask, and the resampled image. Bottom: the same panorama through all eight models. A fresh camera is drawn per view during training.}
\label{fig::supppipe}
\end{figure*}

\cref{fig::suppcams} images one panorama through each of the eight camera models, with a graticule drawn on the sphere before projection so the distortion each model applies is visible as a deformation of the grid.
The fraction of the sphere each one observes is given under its panel, from $14.6\%$ for a $90^\circ$ pinhole to the full sphere for equirectangular.
The parameters behind those panels are the ones the benchmark releases: pinhole $r=f\tan\theta$, Brown--Conrady with $k_1=-0.18$ and $k_2=0.02$, equidistant $r=f\theta$, Kannala--Brandt with a fourth-order polynomial truncated at $\theta_{\max}=100^\circ$, MEI with $\xi=1.6$, double sphere with $\xi=0.25$ and $\alpha=0.55$, and the two latitude--longitude grids at $\pm90^\circ\times\pm45^\circ$ and over the full sphere.
\cref{fig::supppipe} shows how one such frame is produced.
The camera model enters in exactly one place, when the ray map is built from the calibration; downstream every lens is the same tensor of the same shape.
Pixels outside a model's invertible domain are marked invalid and are never filled with a fabricated ray, which is what the validity mask in the second row carries.

\paragraph{The narrow-pinhole limit.} For an ideal pinhole with focal length $f$ in pixels and $u$, $v$ measured from the principal point, $\mathcal{R}(\bm{u})=(u,v,f)/\sqrt{u^2+v^2+f^2}$ gives $\mathrm{d}\Omega_{\bm{u}}=\cos^3\psi_{\bm{u}}/f^2$, with $\psi_{\bm{u}}$ the angle between the ray and the optical axis, so the solid angles of the centre pixel and of the pixel at the largest angle $\psi_{\max}$ differ by the factor $1/\cos^3\psi_{\max}$, already $5.2$ for the corner of a square $90^\circ$ frame.
\cref{eq::sa} therefore reduces to \cref{eq::stockw} only in the limit of a narrow pinhole.

\section{Accuracy Against Field of View}
\label{sec::suppfov} 
\begin{figure*}[!tb]\centering
\includegraphics[width=\textwidth]{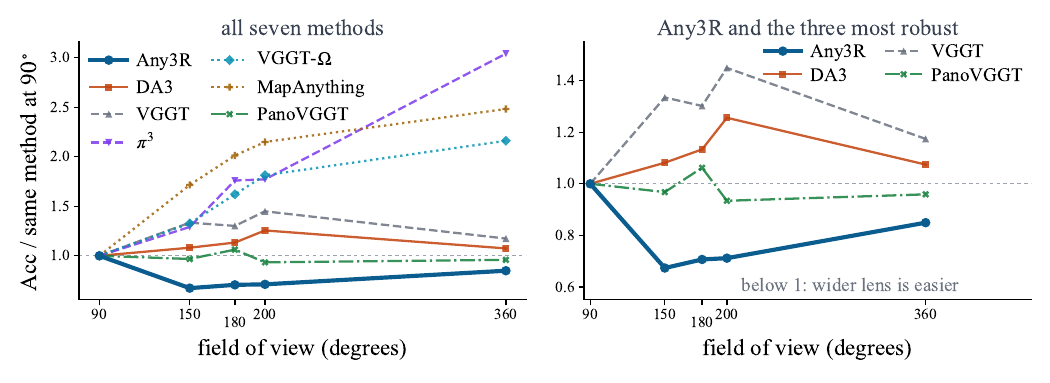}
\caption{\textbf{Acc against field of view, each method normalised by its own $90^\circ$ value and averaged over the three datasets.} The left panel shows all seven general-purpose methods and the right one Any3R and the three that degrade least, on a magnified scale. Below one means the wider lens is the easier one for that method.}
\label{fig::suppfov}
\end{figure*}

Normalising each method by its own $90^\circ$ value is what makes the comparison about the lens rather than about the method's overall quality.
A method that is uniformly weak but equally weak everywhere traces a flat line; a method that depends on the projection traces a rising one.
\cref{fig::suppfov} plots the five field-of-view bands of \cref{tab::p13fov} with each method divided by its own Acc at $90^\circ$ and averaged over the three datasets of the paired pool, so the curves start together and only the slope is compared.
A value below one means the method is more accurate on the wider lens than on the narrow one it was measured against.
Any3R stays below one at every wider field of view, because the encoding and the objective both read the measured ray field, so a wider lens is simply more of the scene per view.
PanoVGGT stays close to one and dips below it everywhere except at $180^\circ$, because its fixed equirectangular assumption is satisfied at the wide end and violated at the narrow end, which is the same failure as the perspective models with the axis reversed.
Every perspective-native method rises above one in every band.

\section{Real Pinhole Imagery}
\label{sec::dl3dv} Every camera in the benchmark is rendered from a panorama, so a reader may reasonably ask whether the model has learnt something about resampled imagery rather than about cameras.
\cref{tab::pose} therefore evaluates camera pose on DL3DV~\cite{ling2024dl3dv}, a capture set of real phone and camera video with no panoramic origin and no resampling step, over $60$ scenes at three views each.
On a pinhole the upper half of the rotary frequencies still spans the image as the stock encoding does, so the mechanism is meant to leave this regime close to the pretrained one.
Our mean over scenes of the per-scene median rotation error is the lowest of the seven methods and our AUC at $30^\circ$ the second highest, while DA3 leads the AUC at $5^\circ$.
Any3R is thus at the level of VGGT on every measure, so the pinhole prior survives.
PanoVGGT, trained on panoramas alone, makes the mirror error and reads every photograph as a panorama, so its rotation error is far larger than that of any other method.

\begin{figure*}[!tb]\centering
\includegraphics[width=\textwidth]{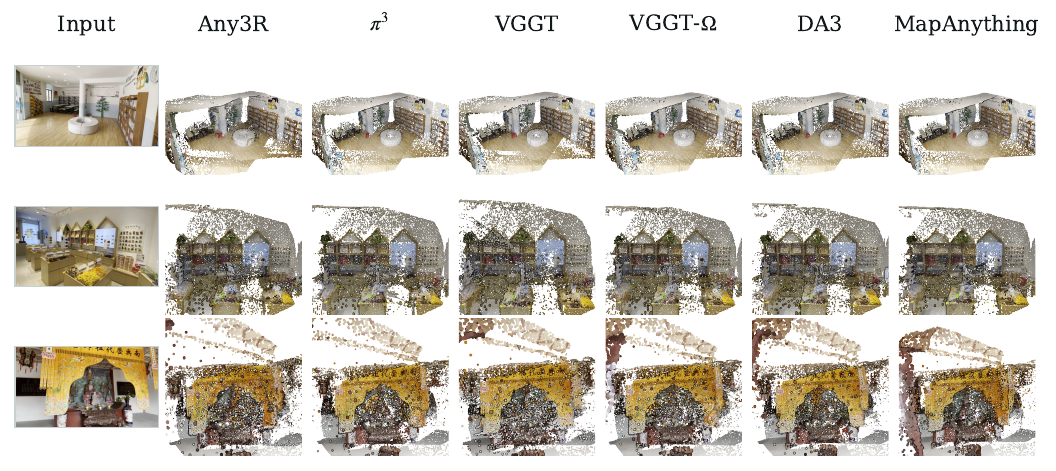}
\caption{\textbf{Real pinhole imagery} on three DL3DV scenes with three views each. Every method fuses its own per-view points with its own predicted poses, one similarity transform fitted on the tracks of the released structure-from-motion reconstruction places each cloud, and each row is rendered from one viewpoint. Sky pixels are found once per input image and removed from every method, and each cloud drops its own farthest $10\%$ of points from its own cameras.}
\label{fig::dl3dv}
\end{figure*}

\cref{fig::dl3dv} shows reconstructions on the same capture set.
DL3DV provides no dense depth, so the only ground truth in the figure is the set of sparse tracks that places each cloud, and the clouds are compared by eye.
The DL3DV frames also make visible a convention of the benchmark that the main text passes over.
Each rendered camera is limited to the cone of its declared field of view, so the $90^\circ$ pinhole cell of \cref{fig::cams} is an inscribed disc with dark corners and about $\pi/4$ of the frame carrying content, whereas a DL3DV frame is a full rectangle.
The cone is what keeps the solid angle equal across camera models at a fixed field of view, so a row of \cref{tab::main} compares sampling rather than coverage, and accuracy there is computed over observed points only.

\section{Matched Coverage}
\label{sec::suppcrop} A wide lens shows more of a room than a narrow one, so a benchmark spanning $90^\circ$ to $360^\circ$ invites the reading that a method is credited for coverage rather than for geometry.
This section removes the difference in coverage and asks the question again.
Each input is centre-cropped and resized back to its original pixel count, so the patch count and the network are unchanged and the field of view is the only variable.
Each variant is scored against the ground-truth points it can itself see, so a narrower view is not charged for the part of the room it no longer contains.
Cropping to half the angular radius leaves $53^\circ$ for the two rectilinear cameras, $90^\circ$ for the equidistant fisheye, $100^\circ$ for KB4 and $180^\circ$ for the equirectangular frame.

\begin{table*}[t]
\centering
\footnotesize
\caption{\textbf{Cropping every camera to a narrower field of view.} Median Acc$\downarrow$ (m) over 10 scenes when each input is centre-cropped to half its angular radius and resized back, scored against the ground-truth points the crop can see.}
\label{tab::crop}
\setlength{\tabcolsep}{2pt}
\begin{tabular}{llccccccc}
\toprule
Dataset & Camera (FoV after crop) & Any3R & $\pi^3$~\cite{wang2025pi3} & MapAnything~\cite{keetha2025mapanything} & DA3~\cite{depthanything3} & VGGT-$\Omega$~\cite{wang2026vggtomega} & VGGT~\cite{wang2025vggt} & PanoVGGT~\cite{panovggt2026} \\
\midrule
Matterport3D & Pinhole (53$^\circ$) & 0.0681 & \cellcolor{second}0.0619 & 0.0749 & 0.0669 & \cellcolor{best}\textbf{0.0513} & 0.0927 & 0.2134 \\
 & Brown--Conrady (53$^\circ$) & 0.0574 & \cellcolor{best}\textbf{0.0488} & 0.0778 & 0.0575 & \cellcolor{second}0.0509 & 0.0841 & 0.2412 \\
 & Equidistant (90$^\circ$) & \cellcolor{best}\textbf{0.0622} & \cellcolor{second}0.0698 & 0.1332 & 0.0865 & 0.0872 & 0.1098 & 0.2178 \\
 & KB4 (100$^\circ$) & \cellcolor{best}\textbf{0.0720} & \cellcolor{second}0.0793 & 0.1205 & 0.0890 & 0.0976 & 0.1222 & 0.2224 \\
 & ERP (180$^\circ$) & \cellcolor{best}\textbf{0.0717} & \cellcolor{second}0.2182 & 0.2476 & 0.2890 & 0.2660 & 0.2512 & 0.2567 \\
\midrule
Stanford2D3DS & Pinhole (53$^\circ$) & 0.0521 & 0.0403 & 0.0485 & \cellcolor{second}0.0391 & \cellcolor{best}\textbf{0.0364} & 0.0910 & 0.2362 \\
 & Brown--Conrady (53$^\circ$) & 0.0543 & \cellcolor{best}\textbf{0.0413} & 0.1034 & \cellcolor{second}0.0481 & 0.0482 & 0.0794 & 0.2470 \\
 & Equidistant (90$^\circ$) & \cellcolor{best}\textbf{0.0785} & \cellcolor{second}0.0826 & 0.1301 & 0.1058 & 0.1135 & 0.1092 & 0.2352 \\
 & KB4 (100$^\circ$) & \cellcolor{best}\textbf{0.0780} & \cellcolor{second}0.0864 & 0.1426 & 0.1053 & 0.1216 & 0.1262 & 0.2475 \\
 & ERP (180$^\circ$) & \cellcolor{best}\textbf{0.0593} & \cellcolor{second}0.2466 & 0.3474 & 0.2839 & 0.2982 & 0.2972 & 0.3294 \\
\bottomrule
\end{tabular}
\end{table*}

\cref{tab::crop} answers the question by the field of view left after cropping.
Any3R is best in two of the ten cells, the cropped equirectangular frame on both datasets, and ties $\pi^3$ in four, the $90^\circ$ equidistant and $100^\circ$ KB4 crops.
On the four $53^\circ$ pinhole and Brown--Conrady crops it is behind, and VGGT-$\Omega$ or $\pi^3$ has the lowest median there.
A narrow pinhole frame is what a perspective-native backbone was built for, and there it draws level or ahead, while the gap opens as the lens departs from that assumption.
The advantage at the wide end therefore survives the removal of coverage, and it does not extend to narrow pinhole crops.
Withdrawing the ray field on these cropped inputs still raises our error several times over in every cell, so the field carries weight at every field of view and not only at the wide end.

\begin{figure*}[!tb]\centering
\includegraphics[width=\textwidth]{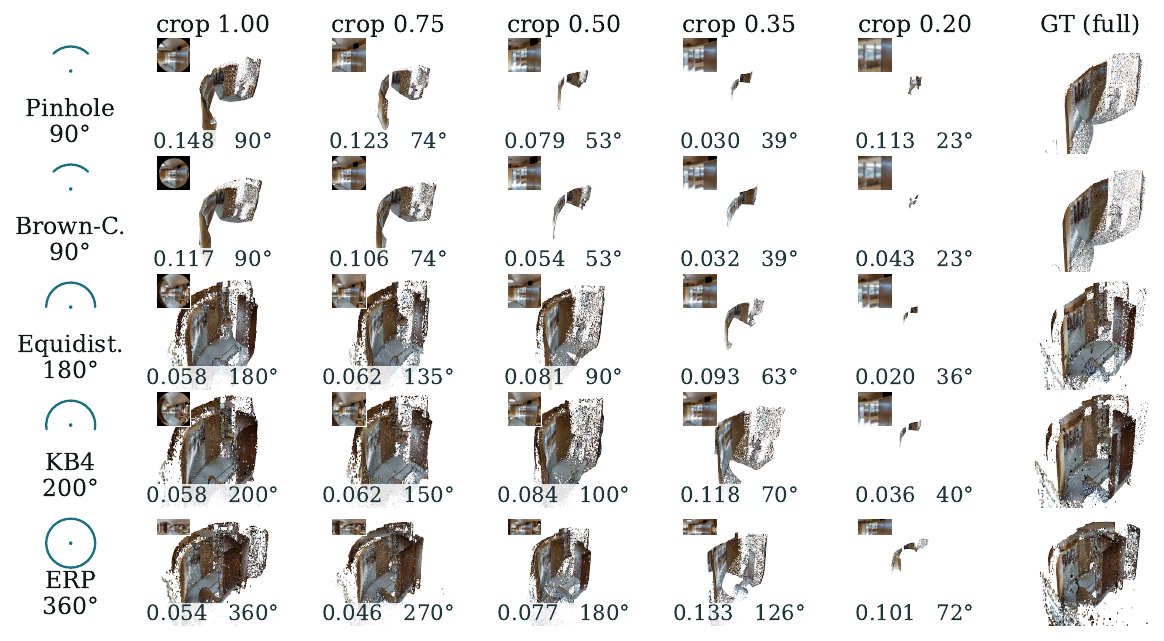}
\caption{\textbf{One camera per row, one crop per column, on Matterport3D.} The inset in each cell is the input that produced it, and the pair of numbers is its Acc and the field of view left after cropping. The viewpoint is shared along a row, so a row shows the field of view narrowing and the reconstruction shrinking together.}
\label{fig::cropqual}
\end{figure*}

\begin{figure*}[!tb]\centering
\includegraphics[width=\textwidth]{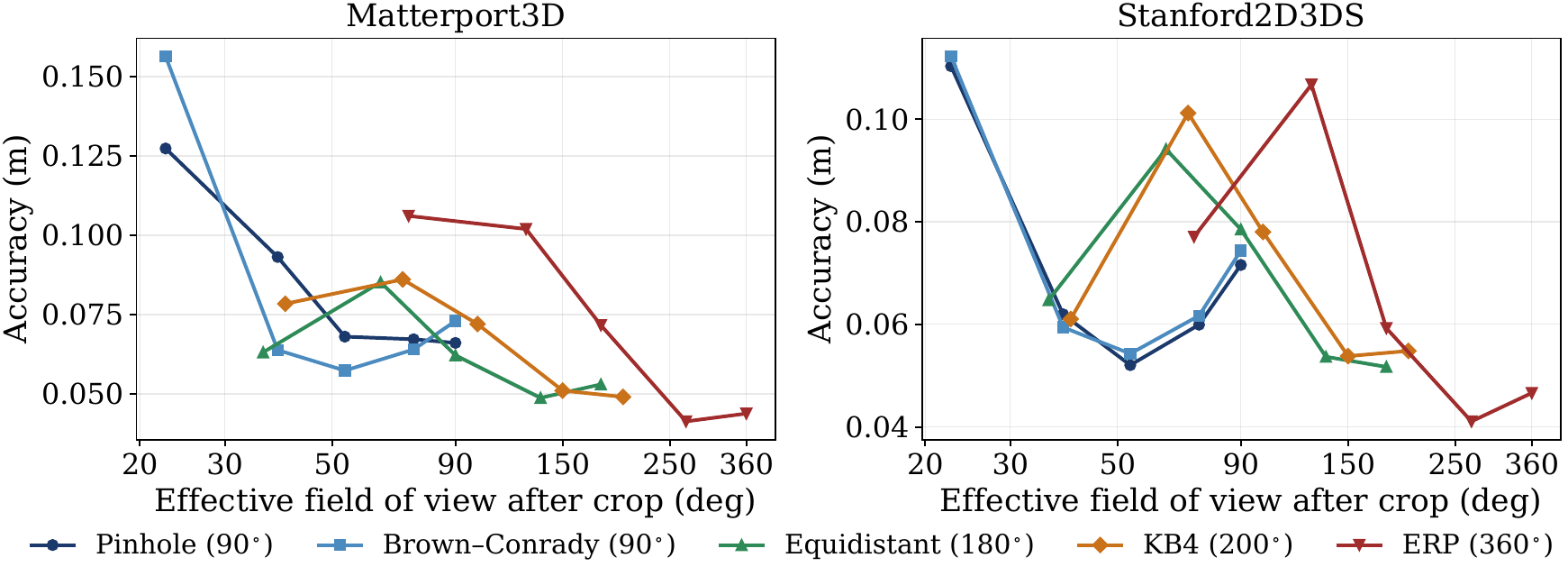}
\caption{\textbf{Accuracy against field of view with the camera held fixed.} Each input is centre-cropped to $1.00$, $0.75$, $0.50$, $0.35$ and $0.20$ of its width and resized back, so only the field of view changes. Reading a curve right to left is reading what a narrower lens costs.}
\label{fig::cropsweep}
\end{figure*}

\cref{fig::cropqual} shows one scene, camera by camera and crop by crop.
The room fills in as the crop widens, and at the narrowest crop every camera keeps only a fragment of it.

\cref{fig::cropsweep} traces the other half of the same question, with the camera held fixed and the crop varied.
Over the fifty pairs of camera and crop the rank correlation between field of view and error is $-0.57$, and the crops below $45^\circ$ have a higher mean error than those above $150^\circ$.
The trend is not monotonic, since the median across the five cameras is lowest at the $0.75$ crop on both datasets and rises slightly at the full frame, and on Stanford2D3DS the $0.35$ crop is worse than the $0.20$ one.
Narrow crops also weaken the pose, because narrow views of one room share less content.
At the full frame the three views of every scene are connected by shared content and the median relative rotation error is $2.0^\circ$, while at the $0.20$ crop $46$ of the $100$ scene and camera pairs lose that connection and the error is $13.4^\circ$.
Our Acc with predicted poses nevertheless stays within about a tenth of the same reconstruction placed with ground-truth poses at every crop, so the rise at the narrow end is not mainly a pose failure.
Because each crop is resized back to the original pixel count, a narrower crop raises angular resolution while lowering coverage, so the direction of this trend places the weight on coverage rather than on resolution.
That is the claim made in \cref{sec::intro} about a wide lens carrying more of the scene per exposure, measured here with the lens held fixed rather than exchanged for another.

\section{Training Convergence}
\label{sec::suppconv}

\begin{figure*}[!tb]\centering
\includegraphics[width=\textwidth]{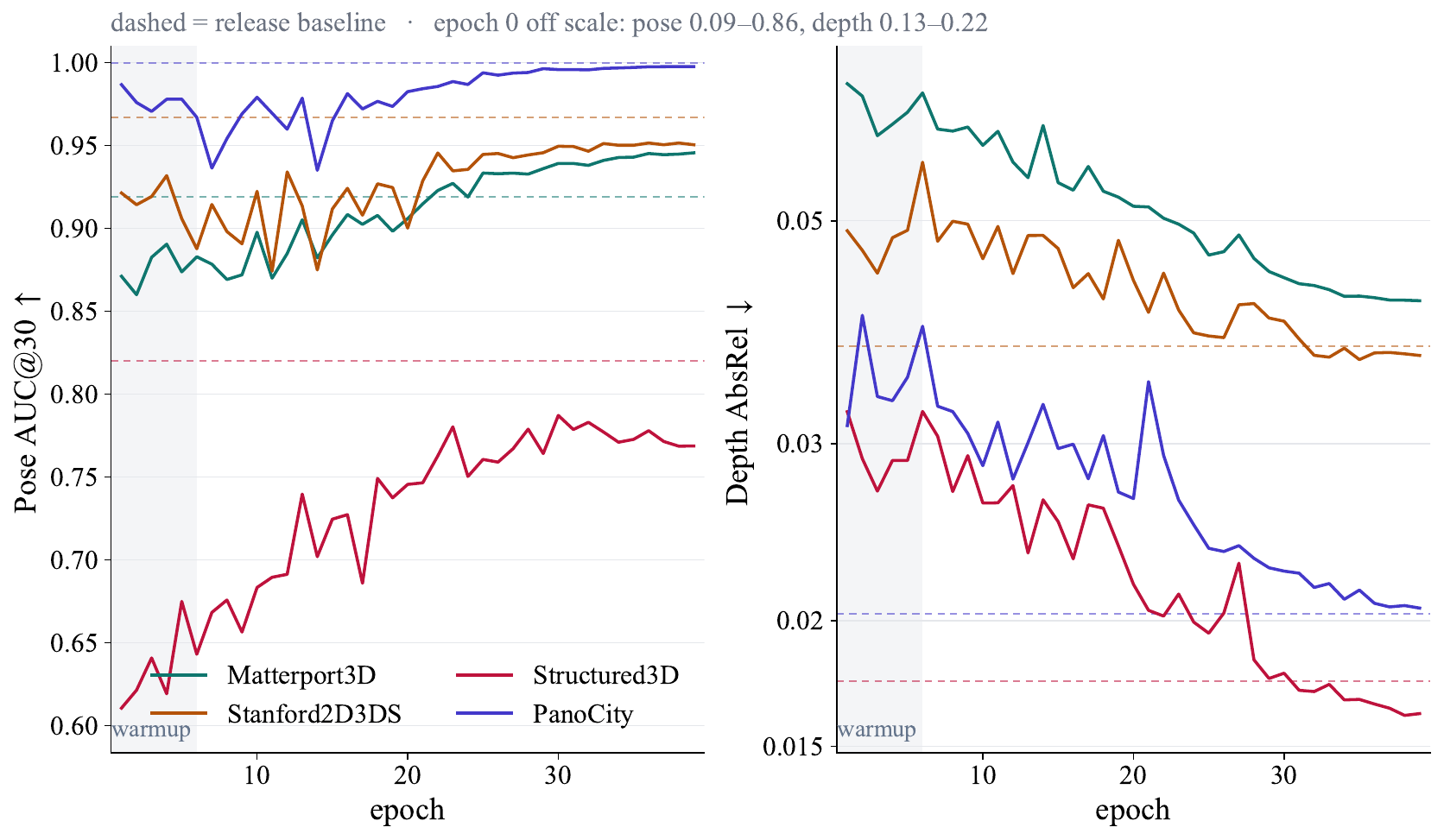}
\caption{\textbf{Convergence of an earlier full run.} Forty epochs from the original initialisation on eight A100s, evaluated under the paper protocol at $518{\times}1036$ on equirectangular input. Dashed lines give the released panoramic baseline on the same protocol; the shaded band is the warm-up.}
\label{fig::suppconv}
\end{figure*}

The released model is trained for $40$ epochs from the original initialisation rather than from a warm start and is evaluated only at its selected checkpoint, so \cref{fig::suppconv} shows the per-epoch pose accuracy and depth error of an earlier full run from the same initialisation, trained on eight A100s, against the released panoramic baseline on the same protocol.
That protocol is equirectangular input only, so it is a narrower question than the tables ask, and the outcome marks the boundary of the claim: final depth is below the baseline on three of the four datasets and level with it on the fourth, while final pose is slightly under it on three.
Any3R is the better model across camera models, which is what \cref{tab::main} and \cref{tab::p13full} measure; it is not a better equirectangular model than one trained for equirectangular input alone.

\section{The Frequency Split} The three arms differ in one switch and are otherwise the same training run, so the columns can be compared directly.
On Matterport3D the constant-angle arm has the lowest median over scenes at $180^\circ$ and $200^\circ$, and the released split lands nearer the field-of-view arm.
At $360^\circ$ the three medians are within a thousandth of each other on Matterport3D and Stanford2D3DS, which is what the split is designed to give, since the frequencies that carry the constant angle are the lower half and those are the ones an equirectangular grid barely exercises.
At $90^\circ$ the constant-angle arm also has the lowest median on both indoor datasets.
The released configuration is the split, which matches the field-of-view arm on pose while recovering part of the constant-angle arm's advantage on geometry.

\begin{table*}[!tb]\centering\footnotesize
\setlength{\tabcolsep}{1.6pt}
\caption{\textbf{Accuracy of the three rotary-frequency configurations.} Acc$\downarrow$ (m) averaged over the cameras of \cref{tab::main} at each field of view, as the mean and as the median over the scenes of each cell.}
\label{tab::bands}
\begin{tabular}{lcccccccccccc}
\toprule
 & \multicolumn{3}{c}{Matterport3D} & \multicolumn{3}{c}{Stanford2D3DS} & \multicolumn{3}{c}{Structured3D} & \multicolumn{3}{c}{PanoCity} \\
\cmidrule(lr){2-4}\cmidrule(lr){5-7}\cmidrule(lr){8-10}\cmidrule(lr){11-13}
FoV & const. & FoV & split & const. & FoV & split & const. & FoV & split & const. & FoV & split \\
\midrule
\multicolumn{13}{l}{\emph{mean over scenes}} \\
90$^\circ$ & \cellcolor{best}\textbf{0.066} & 0.084 & \cellcolor{second}0.083 & \cellcolor{second}0.084 & \cellcolor{best}\textbf{0.083} & 0.084 & 0.783 & \cellcolor{best}\textbf{0.690} & \cellcolor{second}0.732 & \cellcolor{best}\textbf{1.185} & 1.272 & \cellcolor{second}1.202 \\
180$^\circ$ & \cellcolor{best}\textbf{0.042} & 0.054 & \cellcolor{second}0.053 & \cellcolor{best}\textbf{0.049} & 0.065 & \cellcolor{second}0.059 & \cellcolor{best}\textbf{0.334} & \cellcolor{second}0.431 & 0.436 & \cellcolor{best}\textbf{0.734} & \cellcolor{second}0.839 & 0.844 \\
200$^\circ$ & \cellcolor{best}\textbf{0.044} & 0.055 & \cellcolor{second}0.053 & \cellcolor{best}\textbf{0.054} & 0.070 & \cellcolor{second}0.063 & \cellcolor{best}\textbf{0.233} & 0.354 & \cellcolor{second}0.317 & \cellcolor{best}\textbf{0.805} & 1.070 & \cellcolor{second}0.882 \\
360$^\circ$ & 0.047 & \cellcolor{best}\textbf{0.045} & \cellcolor{second}0.045 & 0.069 & \cellcolor{best}\textbf{0.053} & \cellcolor{second}0.064 & \cellcolor{best}\textbf{0.126} & 0.197 & \cellcolor{second}0.128 & \cellcolor{best}\textbf{0.645} & 0.760 & \cellcolor{second}0.663 \\
\midrule
\multicolumn{13}{l}{\emph{median over scenes}} \\
90$^\circ$ & \cellcolor{best}\textbf{0.057} & 0.061 & \cellcolor{second}0.058 & \cellcolor{best}\textbf{0.057} & 0.068 & \cellcolor{second}0.061 & \cellcolor{best}\textbf{0.406} & \cellcolor{second}0.417 & 0.427 & \cellcolor{second}1.059 & \cellcolor{best}\textbf{1.048} & 1.121 \\
180$^\circ$ & \cellcolor{best}\textbf{0.039} & 0.051 & \cellcolor{second}0.048 & \cellcolor{best}\textbf{0.038} & 0.047 & \cellcolor{second}0.047 & \cellcolor{best}\textbf{0.151} & \cellcolor{second}0.293 & 0.323 & \cellcolor{best}\textbf{0.657} & \cellcolor{second}0.803 & 0.804 \\
200$^\circ$ & \cellcolor{best}\textbf{0.043} & 0.054 & \cellcolor{second}0.050 & \cellcolor{best}\textbf{0.042} & 0.055 & \cellcolor{second}0.053 & \cellcolor{best}\textbf{0.105} & \cellcolor{second}0.190 & 0.227 & \cellcolor{best}\textbf{0.680} & 1.007 & \cellcolor{second}0.877 \\
360$^\circ$ & \cellcolor{best}\textbf{0.044} & 0.045 & \cellcolor{second}0.045 & \cellcolor{best}\textbf{0.042} & 0.043 & \cellcolor{second}0.042 & \cellcolor{best}\textbf{0.046} & 0.086 & \cellcolor{second}0.048 & \cellcolor{second}0.600 & 0.648 & \cellcolor{best}\textbf{0.587} \\
\bottomrule
\end{tabular}\end{table*}

\label{sec::bands} \cref{tab::bands} isolates the split described in \cref{sec::arope} against the two arms that apply a single scale to all $16$ frequencies.
Over the camera models used here, one coordinate unit varies by a factor of $1.000$ in elevation and azimuth on the constant-angle half and by $2.82$ and $5.74$ on the other half.
Counting each individual camera rather than one per model family gives $5.92$ and $12.01$.
The agreement at the patch centres derived in \cref{sec::arope} holds with the split in place.
The three checkpoints share every state-dict key and byte count, so they are distinguished only by their hashes, and they were trained identically apart from this switch.
The pre-registered criterion was judged on the earlier measurement with ground-truth poses.
There the split improved the median of every cell at $180^\circ$ and at $200^\circ$ against the field-of-view arm, and against the constant-angle arm it did not, with medians $14.0\%$ and $15.9\%$ worse at those two fields of view and no cell improved, while that arm's pose accuracy collapses and the split's does not.
The criterion for adopting the split as a strict improvement was thus not met, and with predicted poses in \cref{tab::bands} the constant-angle arm again has the lowest median at $180^\circ$ and $200^\circ$ on every dataset.
We release the split configuration for its pose behaviour and record the geometry shortfall here.

\end{document}